%% file: main.tex
\documentclass[10pt,journal,compsoc]{IEEEtran}
\usepackage{ifpdf}
\ifCLASSOPTIONcompsoc
  \usepackage[nocompress]{cite}
\else
  \usepackage{cite}
\fi
\input{config}

\begin{document}

\title{\textbf{\method}: Tracking and Behavior Recognition of Chimpanzees}

\author{Xiaoxuan~Ma\textsuperscript{*}\,\orcidlink{0000-0003-0571-2659},~
        Yutang~Lin\textsuperscript{*}\,\orcidlink{0009-0004-4933-1203},~
        Yuan~Xu\,\orcidlink{0009-0001-7585-5674},~
        Stephan P.~Kaufhold\,\orcidlink{0000-0001-6316-4334},~
        Jack~Terwilliger\,\orcidlink{0000-0002-0235-7455},~ \\
        Andres~Meza\,\orcidlink{0000-0002-4283-0833},~
        Yixin~Zhu~\textsuperscript{\Letter}\,\orcidlink{0000-0001-7024-1545},~
        Federico~Rossano~\textsuperscript{\Letter}\,\orcidlink{0000-0002-6544-7685},~
        and~Yizhou~Wang~\textsuperscript{\Letter}\,\orcidlink{0000-0001-9888-6409}% <-this % stops a space 
\IEEEcompsocitemizethanks{\IEEEcompsocthanksitem Xiaoxuan Ma and Yuan Xu are with Center on Frontiers of Computing Studies, School of Computer Science, Peking University.
E-mail: maxiaoxuan@pku.edu.cn, xuyuan@stu.pku.edu.cn.
\IEEEcompsocthanksitem Yutang Lin is with Yuanpei College, Peking University.
E-mail: yutang.lin@stu.pku.edu.cn.
\IEEEcompsocthanksitem Stephan P. Kaufhold, Jack Terwilliger, Andres Meza and Federico Rossano are with Department of Cognitive Science, University of California San Diego.
E-mail: \{spkaufho, jterwilliger, anmeza, frossano\}@ucsd.edu.
\IEEEcompsocthanksitem Yixin Zhu is with Institute for Artificial Intelligence, Peking University.
E-mail: yixin.zhu@pku.edu.cn.
\IEEEcompsocthanksitem Yizhou Wang is with Center on Frontiers of Computing Studies, School of Compter Science, Peking University, and with Institute for Artificial Intelligence, Peking University, and with Nat'l Eng. Research Center of Visual Technology, and also with Nat'l Key Lab of General Artificial Intelligence, Peking University.
E-mail: yizhou.wang@pku.edu.cn.}
\thanks{\textsuperscript{*} Equal contribution.
\textsuperscript{\Letter} Corresponding authors: Yixin Zhu, Federico Rossano and Yizhou Wang.
}}

\IEEEtitleabstractindextext{%
\begin{abstract}
Understanding non-human primate behavior is crucial for improving animal welfare, modeling social behavior, and gaining insights into both distinctly human and shared behaviors.
Despite recent advances in computer vision, automated analysis of primate behavior remains challenging due to the complexity of their social interactions and the lack of specialized algorithms. Existing methods often struggle with the nuanced behaviors and frequent occlusions characteristic of primate social dynamics.
This study aims to develop an effective method for automated detection, tracking, and recognition of chimpanzee behaviors in video footage.
Here we show that our proposed method, AlphaChimp, an end-to-end approach that simultaneously detects chimpanzee positions and estimates behavior categories from videos, significantly outperforms existing methods in behavior recognition.
AlphaChimp achieves approximately 10\% higher tracking accuracy and a 20\% improvement in behavior recognition compared to state-of-the-art methods, particularly excelling in the recognition of social behaviors. This superior performance stems from AlphaChimp's innovative architecture, which integrates temporal feature fusion with a Transformer-based self-attention mechanism, enabling more effective capture and interpretation of complex social interactions among chimpanzees.
Our approach bridges the gap between computer vision and primatology, enhancing technical capabilities and deepening our understanding of primate communication and sociality.
We release our code and models at \projpage and hope this will facilitate future research in animal social dynamics. This work contributes to ethology, cognitive science, and artificial intelligence, offering new perspectives on social intelligence.
\end{abstract}

% Note that keywords are not normally used for peerreview papers.
}

% make the title area
\maketitle

\IEEEdisplaynontitleabstractindextext

\IEEEpeerreviewmaketitle

\IEEEraisesectionheading{\section{Introduction}\label{sec:introduction}}
\subsubsection*{Background and challenges}

\begin{figure*}[t!]
    \centering
    \includegraphics[width=\linewidth]{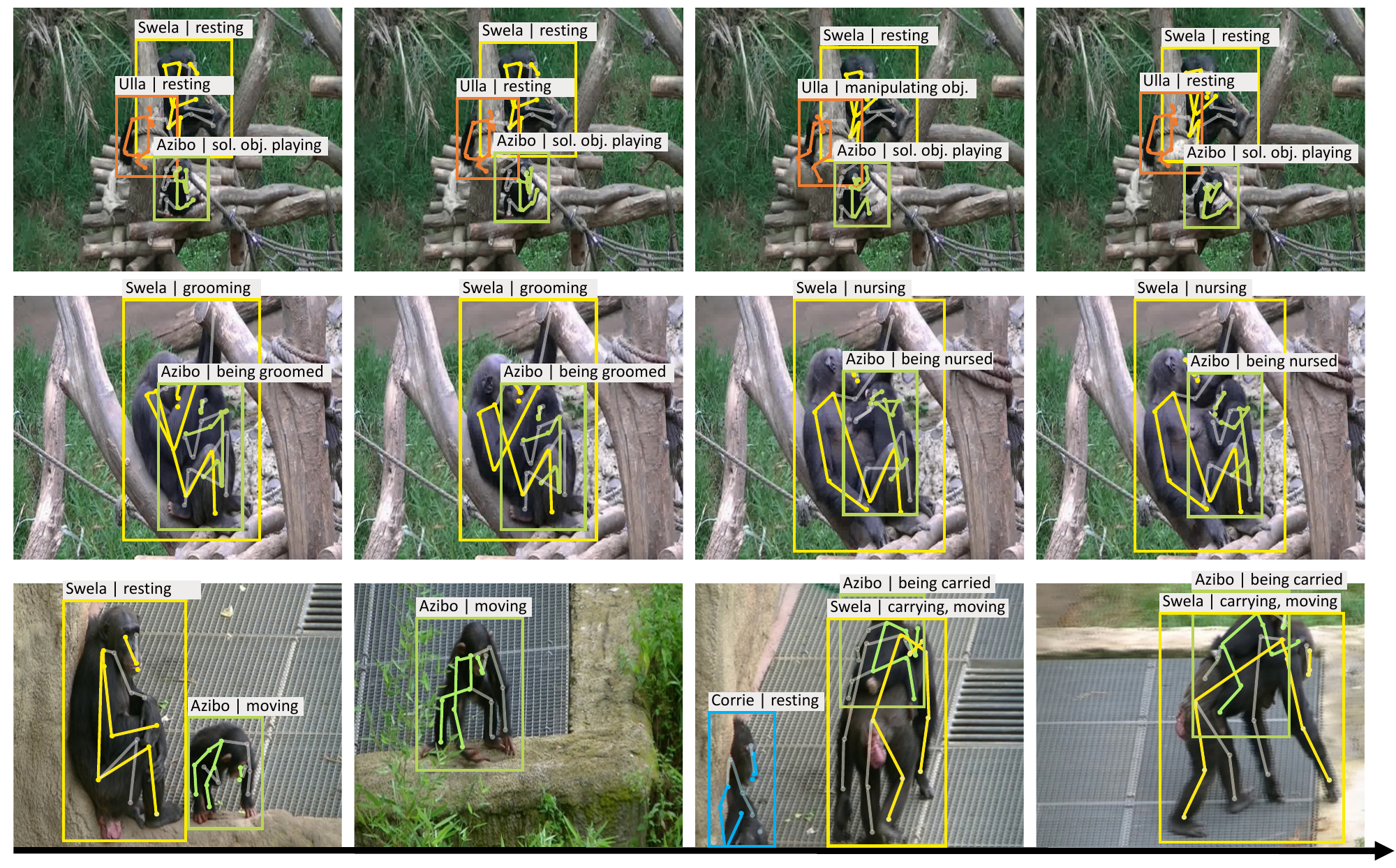} 
    \caption{\textbf{Sample frames and annotations from the \dataset dataset.} We present three video sequences where an infant chimpanzee, named Azibo, is focused. While we also annotate visibility for both the bounding box and the keypoint, these are omitted here for clarity.}%
    \label{fig:annotation}
    \vspace{-1.4em}
\end{figure*}

\IEEEPARstart{S}tudying the behavior of non-human primates is essential for gaining insights into human evolution \cite{langergraber2012generation} and improving animal welfare \cite{dawkins2003behaviour,gonyou1994study}. Given the close phylogenetic relationship between humans and non-human primates, it provides an ethically sound and effective avenue to study the origins of human sociality \cite{sequencing2005initial}. Traditional field research typically requires researchers to enter wildlife conservation areas for extended durations, sometimes spanning multiple years. This involves habituating primate groups to human presence, capturing video footage, and laboriously manually coding these videos for subsequent statistical analysis \cite{hobaiter2017variation,frohlich2020begging,surbeck2017comparison,luncz2018costly,sirianni2015choose}. 

While video coding is heralded as the gold standard for distilling rich, nuanced behavioral patterns \cite{wiltshire2023deepwild}, its practical utility hinges on the efficiency of the coding process. This process demands researchers with specialized expertise and is also prone to attentional biases. The time- and labor-intensive nature of manual coding, coupled with the potential for human error, presents significant challenges to the field. These limitations underscore the need for more efficient and objective methods of analyzing primate behavior, which could potentially accelerate research progress and enhance the reliability of findings in this crucial area of study.

Recent advances in computer vision offer promising avenues for the automated analysis of non-human primate behaviors, particularly chimpanzees. However, the paucity of high-quality longitudinal datasets remains a significant impediment to progress in this field. Assembling comprehensive chimpanzee behavioral data is an arduous task that demands substantial resources and specialized expertise. This process involves continuous video recording coupled with meticulous manual annotation, with a paramount emphasis on annotation accuracy and consistency.

Existing primate datasets present various limitations. Some, such as those developed by \cite{marks2022deep} and \cite{bala2020automated}, restrict subjects to indoor enclosures, resulting in atypical and constrained environments that may not accurately represent natural behaviors. Others, including works by \cite{labuguen2021macaquepose}, \cite{desai2022openapepose}, \cite{ng2022animal}, and \cite{yao2023openmonkeychallenge}, rely on sourcing and labeling primate images from online resources. While these approaches offer certain advantages, they often fail to capture the complex social dynamics inherent to group-living primates such as chimpanzees. This oversight significantly limits comprehensive studies of chimpanzees' social behaviors and relationships, which are crucial aspects of their natural behavior patterns.

\subsubsection*{The \textbf{\dataset} dataset}

\begin{figure*}[t!]
    \centering
    \begin{subfigure}[t!]{0.665\linewidth}
        \centering
        \includegraphics[width=\linewidth]{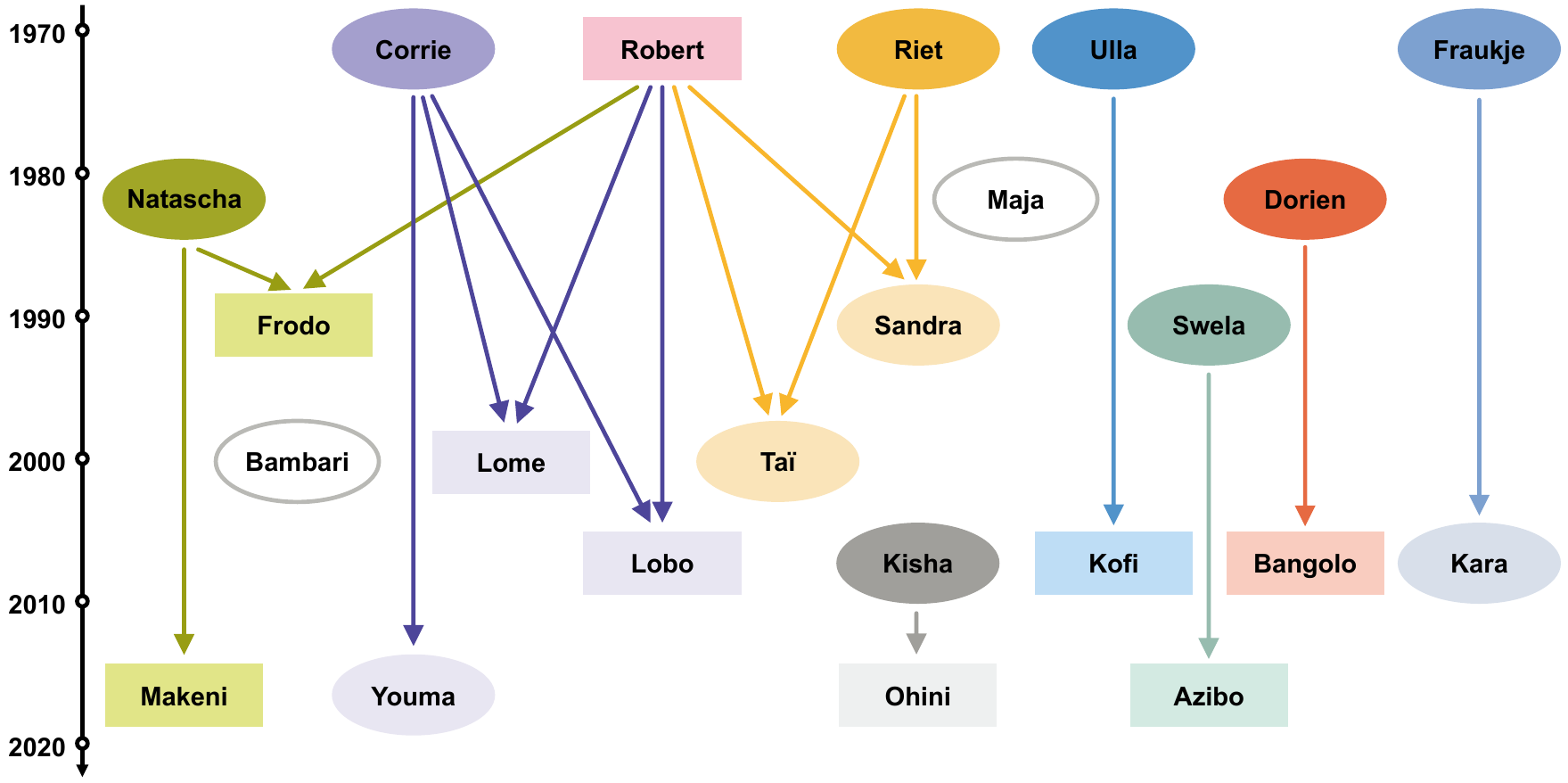} 
        \caption{}
        \label{fig:kinship}
    \end{subfigure}%
    \hfill%
    \begin{subfigure}[t!]{0.335\linewidth}
        \includegraphics[width=\linewidth]{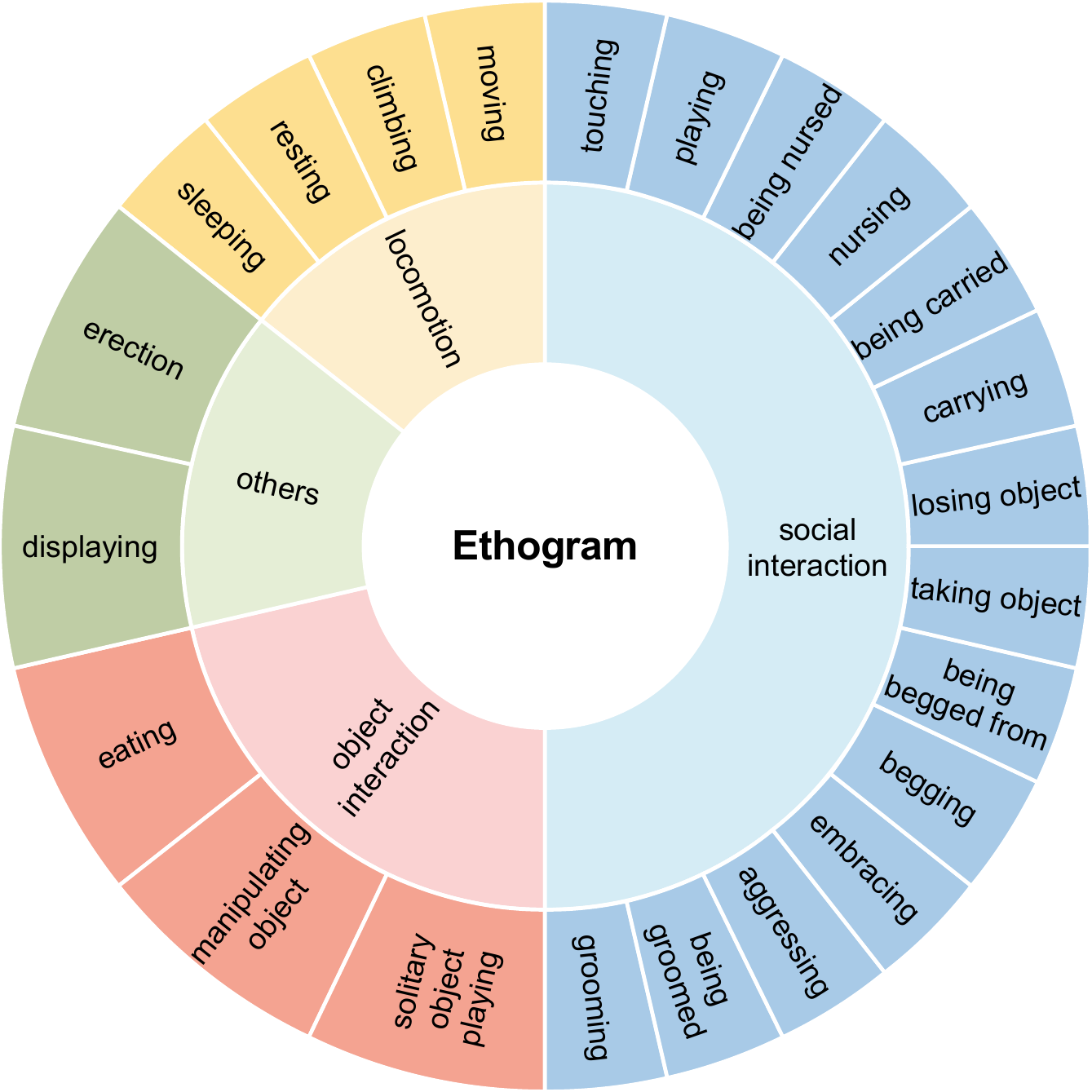}
        \caption{}
        \label{fig:ethogram}
    \end{subfigure}%
    \caption{\textbf{(a) Kinship of the observed chimpanzee group.} Rectangles and ellipses represent males and females, respectively, with arrows flowing from the parents to the child. A missing entry arrow indicates that one of the parents is unknown. Their vertical position relative to the time axis indicates the year of birth. \textbf{(b) Ethogram with annotated behaviors.}}
    \vspace{-1.4em}
\end{figure*}

To address existing dataset limitations, we introduce \dataset, a comprehensive longitudinal dataset for in-depth study of chimpanzee social behavior in a semi-naturalistic setting, with annotations including instance bounding boxes, body poses, and spatial-temporal action labels. A comparison with other datasets is provided in \cref{tab:dataset_compare}. \dataset focuses on a specific chimpanzee group at Leipzig Zoo, Germany, particularly a juvenile male named Azibo (illustrated in \cref{fig:annotation})\footnote{Details about Azibo can be found at \url{https://tinyurl.com/azibo-chimp/}.}. The data, collected from 2015 to 2018 using \textit{focal sampling} \cite{altmann1974observational}, covers Azibo's development since birth within a group characterized by well-defined kin relationships (depicted in \cref{fig:kinship}). The dataset encompasses the daily activities of over 20 chimpanzees, comprising 163 video recordings with approximately 160,500 frames and a total duration of about 2 hours. This extensive coverage allows for comprehensive analysis of chimpanzee behavior and social dynamics in a setting closely approximating their natural habitat, offering a unique perspective on individual development within a chimpanzee community.

\begin{table*}[t!]
    \centering
    \caption{\textbf{Comparison of \dataset with existing primate behavioral datasets.} Square-bracketed numbers denote label counts for the chimpanzee category. $\oslash$ denotes undocumented. For the ``Species'' row, G represents general, P for primates, M for macaques, C for chimpanzees, and C+g for chimpanzees and gorillas. In the ``Source'' row, I stands for Internet, Z for zoo, C for cage, W for wild, and CP for captive.}
    \label{tab:dataset_compare}
    \setlength{\tabcolsep}{5pt}
    \resizebox{\linewidth}{!}{%
        \begin{tabular}{c c c c c c c c c c c c c c c}
            \toprule 
            \multirow{3}{*}{Dataset} & \multirow{3}{*}{Species} & \multicolumn{4}{c}{Track 1} & & \multicolumn{4}{c}{Track 2} & & \multicolumn{2}{c}{Track 3}  & \multirow{3}{*}{Source}\\
            & & \multicolumn{4}{c}{detection, tracking, \acs{reid}} & & \multicolumn{4}{c}{pose estimation} & & \multicolumn{2}{c}{action recognition}  &\\
            \cmidrule{3-6} \cmidrule{8-11} \cmidrule{13-14}
            & & ID \# & frame \# & box \# & track & & frame \# & pose \# & track & dim. & & class \# & label \# &\\
            \toprule
            AP-10K & & & & & & & & 13,028 & & & & & &\\
            \cite{yu2021ap} & \multirow{-2}{*}{G} & \multirow{-2}{*}{\xmark} & \multirow{-2}{*}{\xmark} & \multirow{-2}{*}{\xmark} & \multirow{-2}{*}{\xmark} & & \multirow{-2}{*}{10,015} & [$<$500] & \multirow{-2}{*}{\xmark} & \multirow{-2}{*}{2D} & & \multirow{-2}{*}{\xmark} & \multirow{-2}{*}{\xmark} &\multirow{-2}{*}{I}\\
            \rowcolor{Azure} AnimalKingdom & & & & & & & & 99,297 & & & & & 30,100 &\\
            \rowcolor{Azure} \cite{ng2022animal} & \multirow{-2}{*}{G} & \multirow{-2}{*}{\xmark} & \multirow{-2}{*}{\xmark} & \multirow{-2}{*}{\xmark} & \multirow{-2}{*}{\xmark} & & \multirow{-2}{*}{33,099} & [576] & \multirow{-2}{*}{\xmark} & \multirow{-2}{*}{2D} & & \multirow{-2}{*}{140} & [$\oslash$] &\multirow{-2}{*}{I}\\
            OpenApePose & & & & & & & & 71,868 & & & & & &\\
            \cite{desai2022openapepose} & \multirow{-2}{*}{P} & \multirow{-2}{*}{\xmark} & \multirow{-2}{*}{\xmark} & \multirow{-2}{*}{\xmark} & \multirow{-2}{*}{\xmark} & & \multirow{-2}{*}{71,868} & [18,010] & \multirow{-2}{*}{\xmark} & \multirow{-2}{*}{2D} & & \multirow{-2}{*}{\xmark} & \multirow{-2}{*}{\xmark} &\multirow{-2}{*}{I}\\
            \rowcolor{Azure} OpenMonkeyChallenge & & & & & & & & 111,529 & & & & & &\\
            \rowcolor{Azure} \cite{yao2023openmonkeychallenge} & \multirow{-2}{*}{P} & \multirow{-2}{*}{\xmark} & \multirow{-2}{*}{\xmark} & \multirow{-2}{*}{\xmark} & \multirow{-2}{*}{\xmark} & & \multirow{-2}{*}{111,529} & [$<$10,000] & \multirow{-2}{*}{\xmark} & \multirow{-2}{*}{2D} & & \multirow{-2}{*}{\xmark} & \multirow{-2}{*}{\xmark} & \multirow{-2}{*}{I \& Z}\\
            OpenMonkeyStudio & & & & & & & & 33,192 & & & & & & C\\
            \cite{bala2020automated} & \multirow{-2}{*}{M} & \multirow{-2}{*}{\xmark} & \multirow{-2}{*}{\xmark} & \multirow{-2}{*}{\xmark} & \multirow{-2}{*}{\xmark} & & \multirow{-2}{*}{194,518} & [0] & \multirow{-2}{*}{\cmark} & \multirow{-2}{*}{3D} & & \multirow{-2}{*}{\xmark} & \multirow{-2}{*}{\xmark} & (6.7$m^2$)\\
            \rowcolor{Azure} MacaquePose & & & & & & & & 16,393 & & & & & &\\
            \rowcolor{Azure} \cite{labuguen2021macaquepose} & \multirow{-2}{*}{M} & \multirow{-2}{*}{\xmark} & \multirow{-2}{*}{\xmark} & \multirow{-2}{*}{\xmark} & \multirow{-2}{*}{\xmark} & & \multirow{-2}{*}{13,083} & [0] & \multirow{-2}{*}{\xmark} & \multirow{-2}{*}{2D} & & \multirow{-2}{*}{\xmark} & \multirow{-2}{*}{\xmark} & \multirow{-2}{*}{I \& Z}\\
            SIPEC & & & & 2,200 & & & & & & & & & & C\\
            \cite{marks2022deep} & \multirow{-2}{*}{M} & \multirow{-2}{*}{4} & \multirow{-2}{*}{191} & [0] & \multirow{-2}{*}{\cmark} & & \multirow{-2}{*}{\xmark} & \multirow{-2}{*}{\xmark} & \multirow{-2}{*}{\xmark} & \multirow{-2}{*}{\xmark} & & \multirow{-2}{*}{4} & \multirow{-2}{*}{$\oslash$} & (15$m^2$)\\
            \rowcolor{Azure} PanAf20K  & & & 179,956 & & & & & & & & & & 201,516 &\\
            \rowcolor{Azure} \cite{brookes2024panaf20k} & \multirow{-2}{*}{C+g} & \multirow{-2}{*}{\xmark} & [$\oslash$] & \multirow{-2}{*}{$\oslash$} & \multirow{-2}{*}{\xmark} & & \multirow{-2}{*}{\xmark} & \multirow{-2}{*}{\xmark} & \multirow{-2}{*}{\xmark} & \multirow{-2}{*}{\xmark} & & \multirow{-2}{*}{9} & [$\oslash$] & \multirow{-2}{*}{W}\\
            CCR  & & & & & & & & & & & & & &\\
            \cite{bain2019count} & \multirow{-2}{*}{C} & \multirow{-2}{*}{13} & \multirow{-2}{*}{936,914} & \multirow{-2}{*}{1,937,585} & \multirow{-2}{*}{\cmark} & & \multirow{-2}{*}{\xmark} & \multirow{-2}{*}{\xmark} & \multirow{-2}{*}{\xmark} & \multirow{-2}{*}{\xmark} & & \multirow{-2}{*}{\xmark} & \multirow{-2}{*}{\xmark} & \multirow{-2}{*}{W}\\
            \rowcolor{Azure} ChimpBehave  & & & & & & & & & & & & & &\\
            \rowcolor{Azure} \cite{fuchs2024forest} & \multirow{-2}{*}{C} & \multirow{-2}{*}{\xmark} & \multirow{-2}{*}{12,000} & \multirow{-2}{*}{$\oslash$} & \multirow{-2}{*}{\xmark} & & \multirow{-2}{*}{\xmark} & \multirow{-2}{*}{\xmark} & \multirow{-2}{*}{\xmark} & \multirow{-2}{*}{\xmark} & & \multirow{-2}{*}{7} & \multirow{-2}{*}{$\oslash$} & \multirow{-2}{*}{Z}\\
            \midrule
            \textbf{\dataset} (Ours) & \multirow{2}{*}{C} & \multirow{2}{*}{23} & \multirow{2}{*}{160,500}  & \multirow{2}{*}{56,324}  & \multirow{2}{*}{\cmark} & & \multirow{2}{*}{16,028} & \multirow{2}{*}{56,324} & \multirow{2}{*}{\cmark} & \multirow{2}{*}{2D} & & \multirow{2}{*}{23} & \multirow{2}{*}{64,289} & CP\\
            \cite{ma2023chimpact} & & & & & & & & & & & & & & (4400$m^2$)\\
            \bottomrule 
        \end{tabular}%
    }%
\end{table*}

\dataset features comprehensive annotations, encompassing each individual's detection, tracking, identification, pose estimation, and spatiotemporal action labels, as illustrated in \cref{fig:annotation}. To ensure data precision, each chimpanzee's identity is verified by an experienced behavioral researcher familiar with the Leipzig group. A key feature of \dataset is the implementation of a detailed ethogram (\cref{fig:ethogram}), developed by the same expert, for fine-grained action labeling. This custom-designed ethogram categorizes behaviors into locomotion, object interaction, social interaction, and other actions, each encompassing several detailed sub-categories annotated in our dataset. To our knowledge, \dataset is the \textbf{first} to provide ethogram annotations specifically for the machine learning and computer vision community. This ethogram-based annotation system represents a significant advancement in bridging traditional methods in primatology and modern computational approaches, providing a standardized, biologically relevant framework for behavior classification. By integrating this comprehensive ethogram, \dataset not only offers rich behavioral data but also ensures that the annotations are grounded in established primatological research methods, enhancing its utility for both behavioral and computational studies.

While computer vision has made significant strides in human-centric tasks, such as human pose estimation \cite{sun2019deep,xiao2018simple}, progress on chimpanzee-specific challenges has been limited by the scarcity of suitable datasets. Despite their close genetic relationship to humans \cite{sequencing2005initial}, recognizing chimpanzee behaviors presents unique challenges due to their distinct morphology, appearance, and keypoint articulation. We evaluated prominent human perception methods on three key tracks: (i) detection, tracking, and \ac{reid}, (ii) pose estimation, and (iii) spatiotemporal action detection. Our findings reveal that existing methods struggle with our chimpanzee-specific dataset, with particular deficiencies in social behavior detection. These results underscore the necessity for developing sophisticated, chimpanzee-specific perception models that can accurately capture the nuances of their behavior and social interactions.

\subsubsection*{\method for tracking and behavior recognition}

To address these challenges, we introduce \method, the \textbf{first} fully integrated, end-to-end model specifically designed for chimpanzee detection, tracking, and behavioral recognition in video streams (\cref{fig:pipeline}). \method leverages a DETR-based architecture \cite{carion2020end,zhu2021deformable,zhang2023dino} to simultaneously detect and classify chimpanzees while recognizing their behaviors within a unified framework. Key features of \method include multi-resolution temporal information integration, which aggregates crucial contextual cues to enhance tracking and behavioral recognition, and the attention mechanism to explore spatial relationships, significantly improving the precision of both tracking and behavioral recognition. By focusing on both contextual and temporal dynamics, \method substantially enhances the accuracy of recognizing complex \textit{social} behaviors among chimpanzees. This comprehensive approach allows \method to capture the nuances of chimpanzee behavior more effectively than existing human-centric models, particularly in the context of social interactions.

Experiments on the \dataset dataset demonstrate that \method not only streamlines the perception of primate videos but also significantly enhances accuracy. Our integrated approach yields approximately a 10\% accuracy increase in tracking evaluation protocols, validating its effectiveness in complex scenarios. Notably, \method outperforms \ac{sota} action detection models, \textbf{achieving a 20\% improvement} in the accuracy of detecting social interactions. As the first model to provide end-to-end detection and behavior recognition of chimpanzees in a unified framework, \method achieves \ac{sota} performance across multiple tasks. This new approach highlights the potential of specialized, integrated models to advance primate behavioral research, providing a powerful tool for analyzing complex chimpanzee behaviors and social dynamics. 

\subsubsection*{Summary}

A preliminary version of this work was presented at NeurIPS 2023 \cite{ma2023chimpact}. In this expanded study, our contributions are threefold:
\begin{itemize}[leftmargin=*]
    \item We introduce \method, the \textbf{first} end-to-end and unified framework designed for automated detection, tracking, and fine-grained behavioral recognition of chimpanzees in video footage.
    \item \method achieves notable improvement over all existing \ac{sota} models on the \dataset benchmark across diverse tasks, with a \textbf{10\% improvement} in tracking and a \textbf{20\% improvement} in behavior recognition, despite those models being specifically tailored for each task. This highlights the versatility of our method, with significant advancements in recognizing social behaviors.
    \item This unified framework and the \dataset dataset collectively offer a new resource and platform for the community for advanced techniques for better perception of chimpanzees, ultimately contributing to a deeper understanding of non-human primates.
\end{itemize}

These advancements represent a significant step forward in automated primate behavior recognition, offering new possibilities for comprehensive and accurate studies of chimpanzee social dynamics and behaviors from video data.

\section{Related work}

\subsection{Computer vision for animals}

In recent years, several new datasets and benchmarks leveraging computer vision techniques to advance animal research have been introduced. These efforts span a wide range of species and tasks. For example, 3D-ZeF20 \cite{pedersen20203d} introduces 3D tracking of zebrafish to the \ac{mot} benchmarks, while AnimalTrack \cite{zhang2023animaltrack} focuses on multi-animal tracking across various species. In the realm of pose estimation, AP-10K \cite{yu2021ap} and APT-36K \cite{yang2022apt} address this task for diverse species. AnimalKingdom \cite{ng2022animal} extends the scope to fine-grained multi-label action recognition. Several studies explore multi-agent behavior understanding from a social interaction perspective \cite{sun2021multi,sun2023mabe22}. KABR \cite{kholiavchenko2024kabr} contributes by collecting videos from drones flown over the Mpala Research Centre in Kenya. Recently, PanAf20K \cite{brookes2024panaf20k} curated a dataset for chimpanzee behavior recognition, but it lacks clear social bonds or fine-grained ethogram. Distinctively, \dataset stands out as a comprehensive benchmark, encompassing three varied downstream tasks and featuring detailed annotations of social interactions within the same chimpanzee group. This enables a nuanced, longitudinal analysis of chimpanzee social dynamics.

\subsection{Human video datasets}

In contrast to animal-centric video datasets, a more substantial collection exists for human subjects, addressing diverse human-centric video understanding tasks. These datasets cover a wide range of applications in computer vision. For multi-person tracking, the \ac{mot} Challenge \cite{milan2016mot16} serves as a primary benchmark. Human pose estimation is well-served by datasets such as COCO \cite{lin2014microsoft} and MPII \cite{andriluka20142d}, which provide extensive annotations for body keypoints. In the domain of action recognition, datasets like Kinetics \cite{kay2017kinetics}, ActivityNet \cite{caba2015activitynet}, and AVA \cite{gu2018ava} offer large-scale video collections with diverse human activities. While \dataset encompasses analogous tasks to these human-centric datasets, it introduces unique challenges specific to chimpanzee behavior. This approach allows for the adaptation and advancement of human-centric computer vision techniques to the study of non-human primates, bridging the gap between human and animal behavior analysis.

\subsection{Datasets on primate behavioral understanding}

Most existing primate datasets focus primarily on individual primate detection and pose estimation, with limitations that hinder behavioral analysis in context. Datasets derived from confined laboratory settings \cite{bala2020automated,marks2022deep} may induce atypical behavioral patterns and constrain the expression of species-typical behaviors, while those collected from online sources \cite{labuguen2021macaquepose,desai2022openapepose,ng2022animal,yao2023openmonkeychallenge} often lack longitudinal interactions crucial for analyzing chimpanzee social dynamics. The CCR dataset \cite{bain2019count}, chronicling 13 chimpanzees in the Bossou forest over two years, primarily focuses on individual detection and recognition, lacking behavioral annotations necessary for studying the social dynamics of wild primates.

Recent efforts like PanAf20K \cite{brookes2024panaf20k} include behavioral annotations but provide either coarse labels or only limited fine-grained annotations (in PanAf500) with minimal focus on social behaviors. It also lacks detailed information on social relationships, longitudinal observation, and keypoint annotations. The ChimpBehave dataset \cite{fuchs2024forest}, introduced to explore cross-dataset generalization, covers only a short period and lacks social behavior data, making it unsuitable for studying chimpanzee social development.

In contrast, \dataset offers a multifaceted approach, encompassing identities, kinship, detection labels, pose annotations, and fine-grained action labels based on an ethogram. The features and size of our dataset position it as a highly relevant benchmark and tool for developing advanced automated chimpanzee behavior analysis methods and enhancing the overall understanding of primate behavior. \cref{tab:dataset_compare} provides a detailed comparison, highlighting the unique features of \dataset. Despite the recent introduction of newer chimpanzee datasets, \dataset \cite{ma2023chimpact} remains the most comprehensive dataset available today, addressing the limitations of existing datasets.

\subsection{Computational methods for primate behavioral analysis}

Categorizing and analyzing behavior is instrumental in understanding primate social dynamics and cognitive abilities. Behavioral analysis often encompasses subtasks like individual detection, tracking, and identification \cite{bain2019count,marks2022deep}, pose estimation \cite{labuguen2021macaquepose,desai2022openapepose,mathis2018deeplabcut,wiltshire2023deepwild}, and behavior recognition \cite{ng2022animal,bain2021automated,brookes2024chimpvlm}. While each task has specialized techniques, many are rooted in human behavioral research. Numerous algorithms exist for human tracking \cite{bewley2016simple,pang2021quasi}, pose estimation \cite{sun2019deep,xiao2018simple}, and behavior recognition \cite{feichtenhofer2019slowfast}. However, due to the scarcity of relevant primate datasets, primate behavioral analysis often repurposes algorithms designed for humans, including:
\begin{itemize}[leftmargin=*]
    \item \textbf{Detection, tracking, and \ac{reid}} identify individual primates in videos, often leveraging established object or human detection algorithms like Mask-RCNN \cite{he2017mask}. For instance, SIPEC \cite{marks2022deep} employs Mask-RCNN with a ResNet backbone \cite{he2016deep} to track and segment macaques. The method uses a region proposal network (RPN) to generate candidate bounding boxes, followed by a classification and segmentation head to refine the detections. \cite{bain2019count} utilize CNNs to crop and identify individual chimpanzees, implementing a two-stage approach where a detector first localizes chimpanzees, followed by a separate CNN for individual identification.
    \item \textbf{Pose estimation} discerns primate poses, frequently adapting human pose estimation methods like SimpleBaseline \cite{xiao2018simple}. DeepLabCut \cite{mathis2018deeplabcut,lauer2022multi}, for instance, employs ResNet-50 with ImageNet pre-trained weights for 2D animal pose estimation. It uses a fully convolutional architecture to predict heatmaps for each keypoint, allowing for precise localization of body parts. SIPEC \cite{marks2022deep} modifies SimpleBaseline for 2D macaque poses, adapting the keypoint configuration to match macaque anatomy and fine-tuning the model on macaque-specific data.
    \item \textbf{Behavior recognition} identifies primate actions and interactions. Contemporary methods \cite{bain2021automated,bohnslav2021deepethogram} often derive from human action recognition algorithms like SlowFast \cite{schindler2021identification}. Notably, \cite{bain2021automated} integrates audio cues for classifying two simple non-interactive behaviors: nut cracking and buttress drumming. Their approach combines visual features extracted from a 3D CNN with audio features processed through a separate neural network, fusing the modalities for final classification. ChimpVLM \cite{brookes2024chimpvlm} leverages text to enhance the recognition of behaviors across 9 non-social classes, employing a vision-language model that aligns visual features with textual descriptions of behaviors to improve classification accuracy.
\end{itemize}

In contrast, we propose \method, a unified framework capable of simultaneously detecting chimpanzees and recognizing over 20 behavior classes defined in \dataset's ethogram. It demonstrates superior performance compared to task-specific methods by leveraging a single end-to-end architecture that jointly optimizes multiple tasks, allowing for more efficient feature sharing and context utilization across detection and behavior recognition.

\section{\texorpdfstring{\dataset}{}}\label{sec:dataset}

\subsection{Dataset description}

\dataset is a comprehensive collection of high-resolution video footage documenting chimpanzees at the Leipzig Zoo in Germany from 2015 to 2018. The dataset comprises approximately 2 hours of recordings, focusing primarily on Azibo, a male chimpanzee born in April 2015 to Swela. Azibo has been a member of the A-chimpanzee group since birth, providing a unique opportunity for longitudinal observation of his behavioral development and social interactions. The A-chimpanzee group, comprising more than 20 individuals, is among the most extensively studied cohorts of zoo-residing chimpanzees. Numerous behavioral and cognitive studies, both observational and experimental, have been conducted on this group \cite{baker2022wolfgang,mcewen2022primate}.

\dataset's longitudinal nature offers an unprecedented window into Azibo's growth, social interactions, and intra-group relationships within this complex social environment. The dataset captures a wide range of behaviors and social dynamics, making it an invaluable resource for researchers studying primate behavior, social cognition, and development. The high-resolution footage, combined with detailed annotations and the diverse social composition of the group, positions \dataset as a unique tool for in-depth analysis of primate social dynamics and individual development. It offers researchers the opportunity to examine subtle behavioral cues, track changes over time, and explore the intricate social fabric of chimpanzee society in a semi-naturalistic setting.  \\

\begin{figure}[t!]
    \centering
    \begin{subfigure}[t!]{\linewidth}
        \centering
        \includegraphics[width=\linewidth]{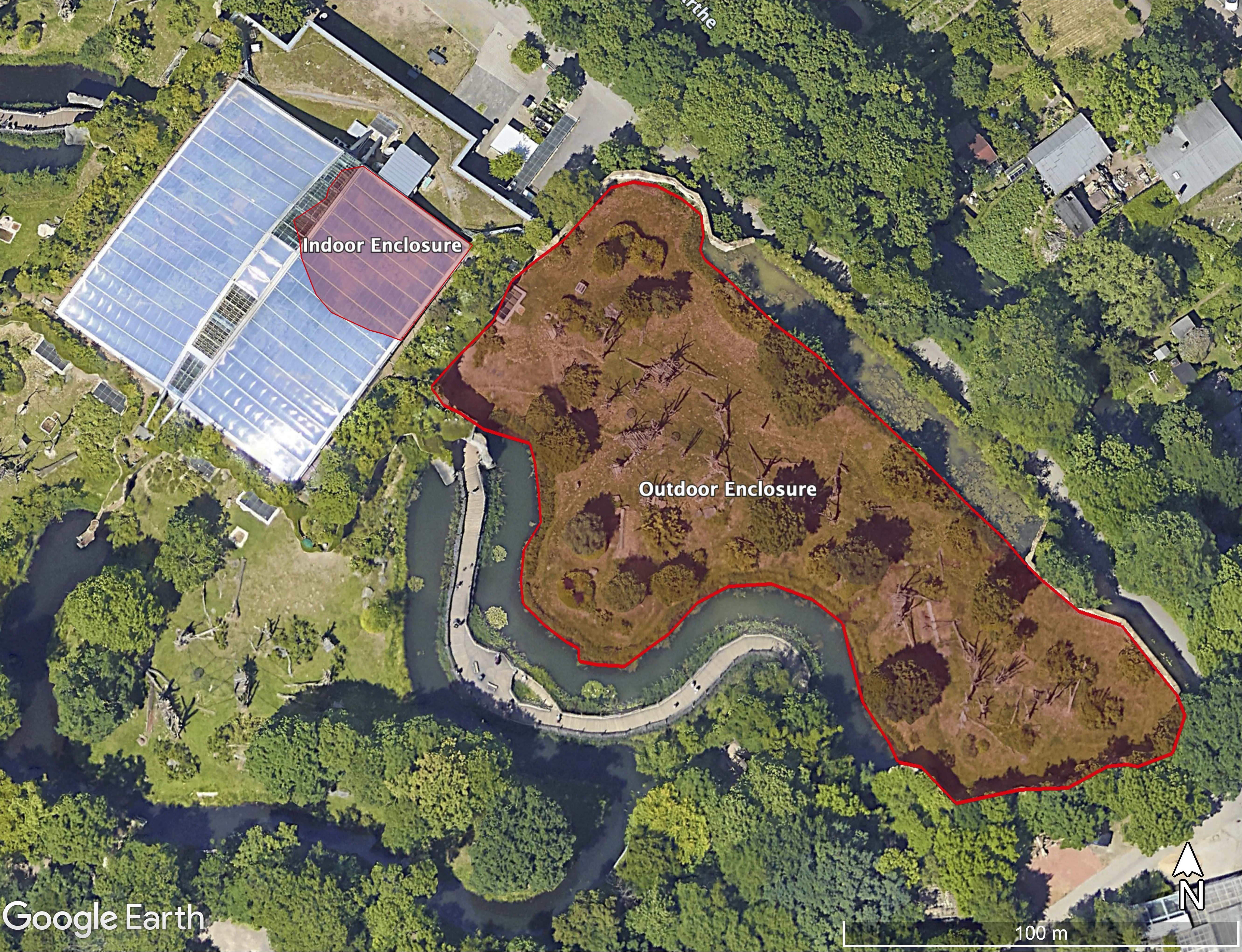}
        \caption{aerial view of Leipzig Zoo}
    \end{subfigure}%
    \\
    \begin{subfigure}[t!]{0.445\linewidth}
        \centering
        \includegraphics[width=\linewidth]{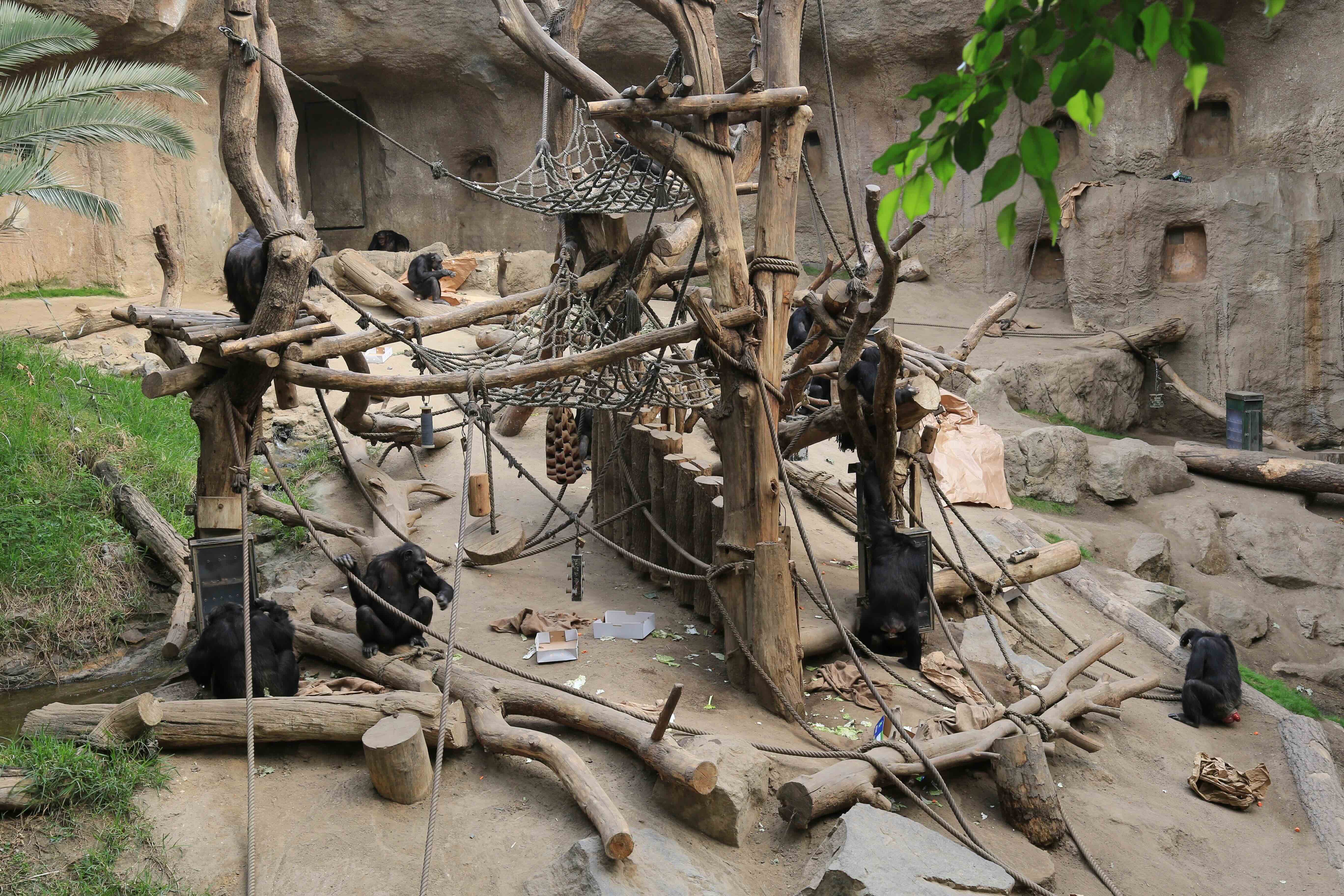}
        \caption{indoor enclosure}
        \label{fig:zoo_in}
    \end{subfigure}%
    \hfill%
    \begin{subfigure}[t!]{0.53\linewidth}
        \centering
        \includegraphics[width=\linewidth]{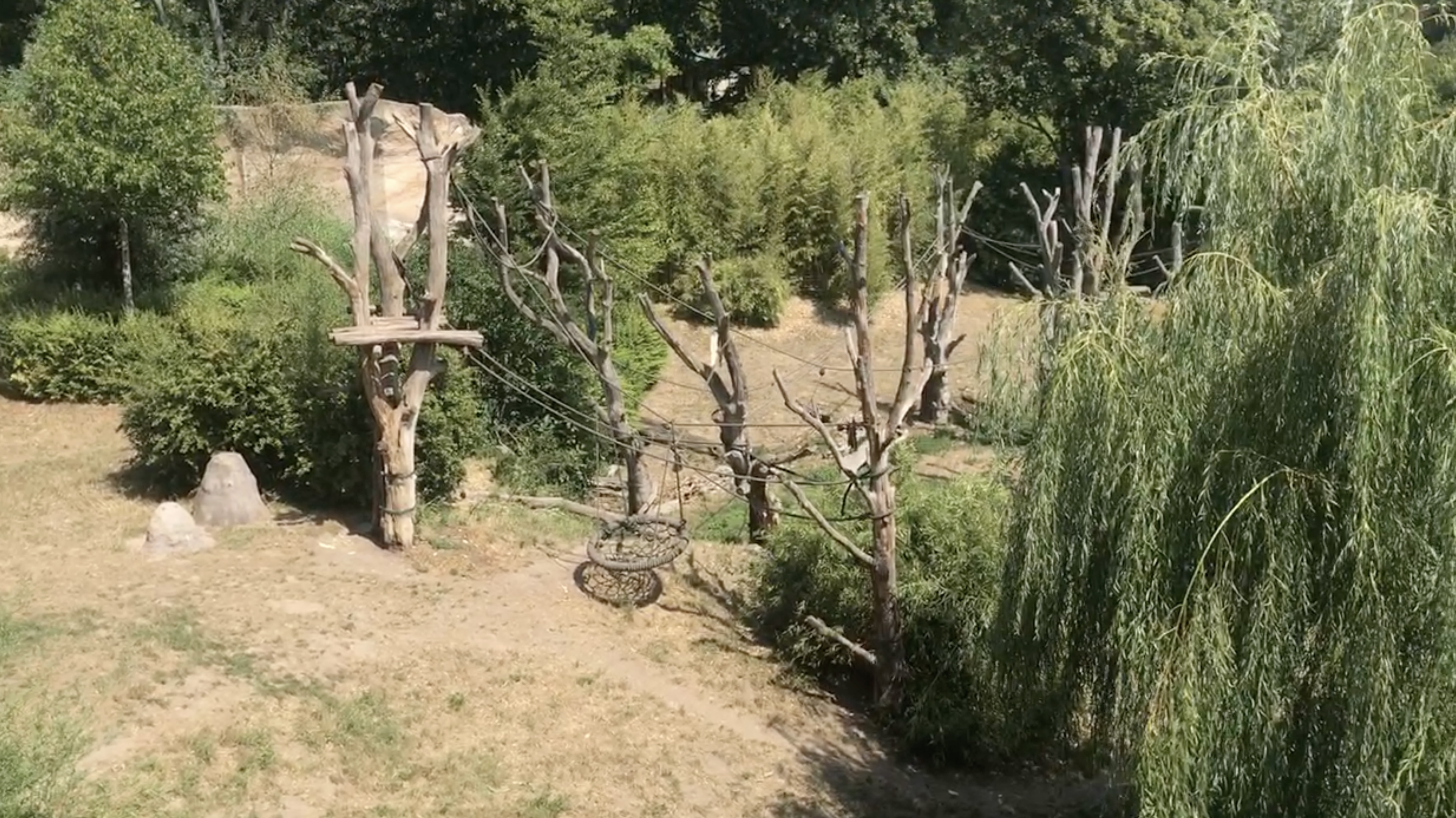}
        \caption{outdoor enclosure}
        \label{fig:zoo_out}
    \end{subfigure}%
    \caption{\textbf{Semi-naturalistic habitats at Leipzig Zoo.} (a) The aerial view of Leipzig Zoo \cite{zoo_all}, including both indoor and outdoor enclosure. (b) An example scene inside the indoor enclosure \cite{zoo_in}. Photo used under CC BY-SA 4.0. (c) An example scene of the outdoor enclosure. }
    \vspace{-1.4em}
    \label{fig:zoo}
\end{figure}

\noindent\textbf{Longitudinal data.\quad}
\dataset contains footage of a stable zoo-residing chimpanzee group over a four-year observational period, providing a unique longitudinal perspective on chimpanzee behavior development. This extended timeframe allows for comprehensive tracking of a young chimpanzee's growth and social interactions within the group context. Such longitudinal data offers valuable insights into several critical aspects of chimpanzee development.

The dataset enables the study of chimpanzee socialization processes and the evolution of social skills \cite{matsuzawa2013evolution}, the examination of social bond formation, and the gradual integration of individuals into the group's dominance hierarchy \cite{matsuzawa2006development}. Furthermore, the longitudinal nature of \dataset permits the investigation of the acquisition and transmission of group-specific cultural behaviors \cite{van2021temporal,musgrave2021ontogeny}. By capturing developmental trajectories over time, \dataset provides a unique opportunity to study the progression of social behaviors and relationships within a chimpanzee community. This longitudinal approach deepens our understanding of the interaction between individual growth and group dynamics in primate societies, making it a valuable resource for analyzing chimpanzee social behavior and cognition. \\

\begin{figure*}[t!]
    \centering
    \begin{subfigure}[t!]{.68\linewidth}
        \centering
        \includegraphics[width=\linewidth]{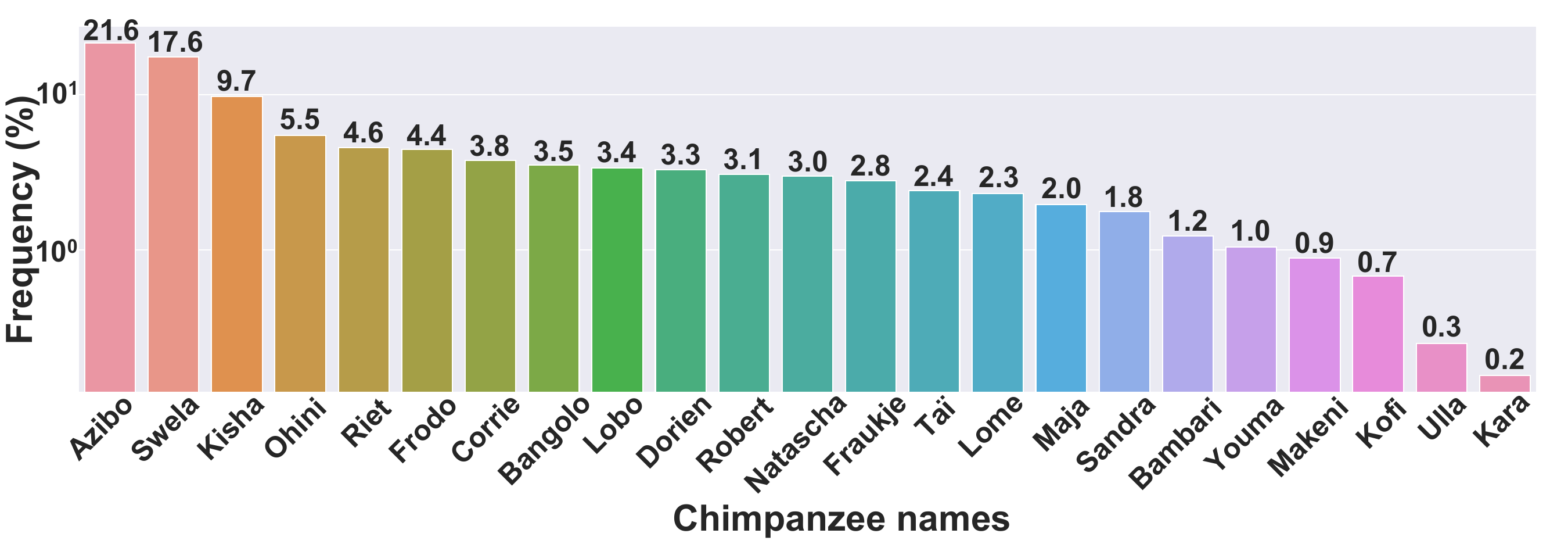}
        \caption{}
        \label{fig:identity_dist}
    \end{subfigure}%
    \\
    \begin{subfigure}[t!]{.68\linewidth}
        \centering
        \includegraphics[width=\linewidth]{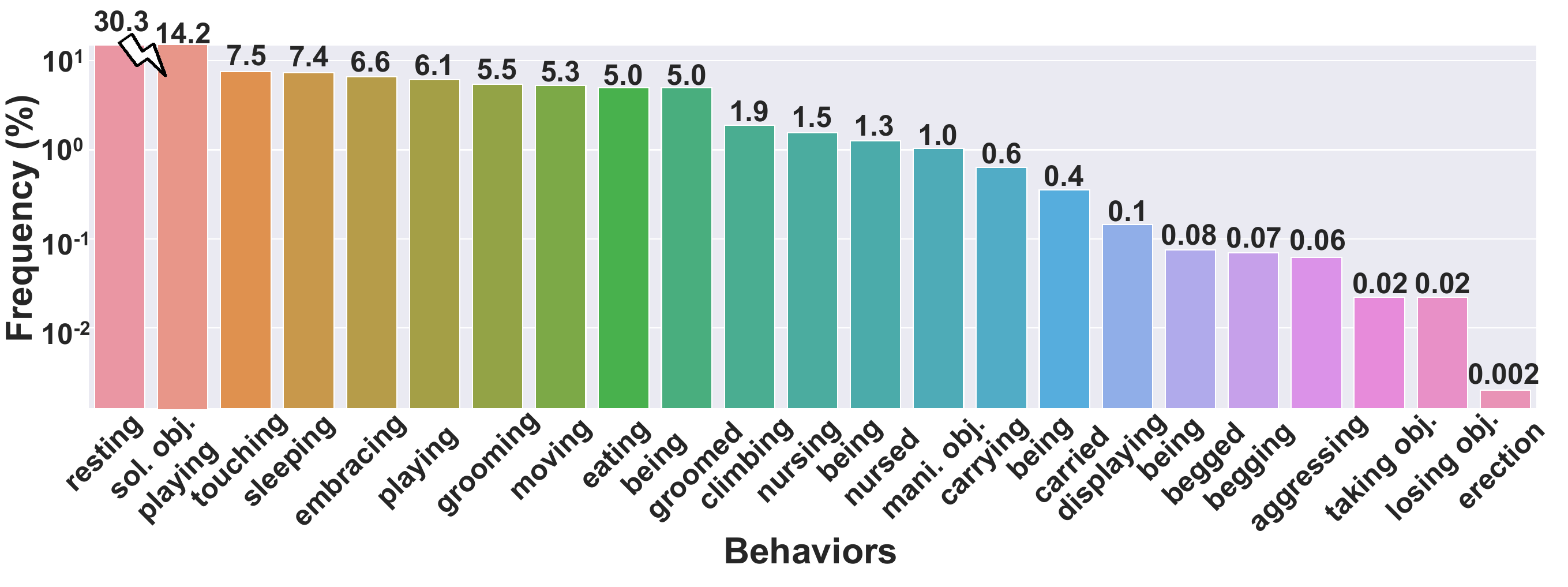}
        \caption{}
        \label{fig:behavior_dist}
    \end{subfigure}%
    \caption{\textbf{(a) Log-scale distribution of annotations per individual \textbf{(b) Log-scale distribution of annotations per behavior.}}}
    \vspace{-1.2em}
\end{figure*}

\noindent\textbf{Semi-naturalistic and social environment.\quad}
The videos in \dataset capture chimpanzees in their semi-naturalistic habitats at Leipzig Zoo, split between indoor (96 videos) and outdoor (67 videos) enclosures; see also \cref{fig:zoo}. The indoor space (\cref{fig:zoo_in}), spanning roughly 400~$m^2$, features a plethora of environmental enrichments, ranging from 15 wooden climbing structures with heights ranging from 2 to 5 meters, 8 hammocks, vegetation consisting of over 20 plant species, and 12 foraging boxes designed to simulate natural food-finding behaviors. When weather permits, the chimpanzees have access to a 4000~$m^2$ outdoor area (\cref{fig:zoo_out}), This expansive space features abundant vegetation, including 30 trees of various species, and is bordered by a 3-meter wide artificial river. The outdoor environment is further enhanced with enrichments similar to those in the indoor space. This area presents additional complexities for video analysis due to variable weather conditions and spatial arrangements. This blend of environments ensures the dataset's relevance for both naturalistic and artificial environments. The multifaceted physical and social surroundings of the chimpanzees further imbue the dataset with intricate behaviors and social dynamics.  \\

\noindent\textbf{Ethogram with solitary and social behaviors.\quad}
\dataset captures the daily life of group-living chimpanzees, offering invaluable insights into the evolution and sustenance of their social behaviors and relationships \cite{nishida2010chimpanzee}. By focusing on a juvenile chimpanzee, \dataset illuminates facets of social learning, communication, bonding, and more, all pivotal in the social and ecological life of chimpanzees \cite{bard2014gestures}. To systematically represent these behaviors, we compose an ethogram, a detailed catalog of behavioral categories, depicted in \cref{fig:ethogram}. This ethogram organizes behaviors into four primary categories, like locomotion and social interaction, each further subdivided into several fine-grained actions, meticulously annotated and validated with expert oversight. By delving into these behaviors, \dataset elucidates not only the social dynamics shaping social relationships but also the cognitive and ecological influences on juvenile chimpanzee behaviors. For more details of the ethogram and its constituent behavioral categories, readers are directed to \cref{supp:sec:ethogram}.

\subsection{Dataset collection}\label{sec:data-collection}

The focal video data were collected with the Chimpanzee-A group housed at Leipzig Zoo, Germany, using focal sampling \cite{altmann1974observational}. Videographers were instructed to focus on Azibo and his mother, Swela, but also on capturing the environmental context and his interactions with other chimpanzees. Videos from \dataset were sampled from a larger set of around 405 hours of longitudinal focal video recordings of the dyad between 2015 and 2018. These videos were recorded by several research assistants during the daytime (7am--4pm) using tripod-mounted RGB cameras. Two JVC Everio camera models were utilized across the years, filming with a framerate of 25 (Codec H.264) and with resolutions of $720 \times 578$ and $1280 \times 720$, respectively. The mother-infant dyad was filmed for about five hours each week during the observation period. The footage contains both optical zoom and camera movements.

\subsection{Dataset tasks and annotations}\label{sec:data-annot}

\dataset supports three tracks: (i) chimpanzee detection, tracking, and \ac{reid}, (ii) chimpanzee pose estimation, and (iii) spatiotemporal action detection. We provide fine-grained annotations for each track. From the extensive footage, we curated 163 video clips, each approximately 1000 frames in length. Fifteen adept annotators were then tasked with annotating bounding boxes, body keypoints, and fine-grained behavioral classes for each chimpanzee at intervals of every 10 frames. To ensure accuracy and consistency, a behavioral researcher familiar with the chimpanzee group meticulously reviewed and refined the identity and behavioral class annotations. For a deeper dive into the annotation process and its quality, please refer to \cref{supp:sec:datadetail} and our dedicated \dataprojpage, which provides additional resources and information. \\

\noindent\textbf{Detection, tracking, and \ac{reid}.\quad}
This task encompasses the detection and tracking of individual chimpanzees across video sequences, subsequently coupled with their re-identification. \dataset features over 23 distinct chimpanzee individuals, each identified by a primate expert familiar with the Leipzig A-group chimpanzees. Initially, annotators were instructed to delineate the bounding box of each chimpanzee, ensuring consistent box IDs for the same individual throughout a video clip. Subsequently, the expert matched these box IDs with the corresponding true names of the chimpanzees, resulting in the identification of 23 unique individuals. Additionally, every annotated bounding box is attached with a visibility attribute, indicating if the chimpanzee is fully visible, truncated, or occluded in a given frame. Such visibility annotations can support the reasoning of the chimpanzee behavior, potentially bolstering tracking robustness. \cref{fig:identity_dist} illustrates the occurrence frequency (on a \textit{log} scale) of each individual, revealing a long-tail distribution. This pattern aligns with the focal sampling strategy, where Azibo is the primary subject. Notably, Swela, Azibo's mother, also exhibits a high occurrence frequency, resonating with prior studies \cite{boesch1996emergence}.  \\

\begin{figure}[ht!]
    \centering
    \includegraphics[width=.45\linewidth]{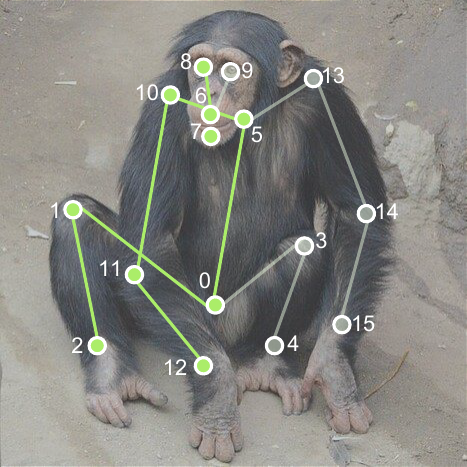}
    \caption{\textbf{Keypoint definitions for chimpanzee.}}
    \vspace{-1.4em}
    \label{fig:keypointdef}
\end{figure}

\noindent\textbf{Pose estimation.\quad}
Pose estimation aims to predict the locations of the chimpanzee joints that have semantic meaning, such as the knee and shoulder, from an input image. There are four keypoints on the chimpanzee's face (\ie, two for the eyes, and one each for the upper and lower lips), for a total of 16 chimpanzee keypoints (refer to \cref{tab:keypointdef,fig:keypointdef}). Annotators are tasked with marking the 2D joint coordinates and the visibility status of each joint. We adopt the visibility protocol from the COCO 2D human keypoint annotations \cite{lin2014microsoft}, where a value of 0 indicates a joint outside the image frame, 1 signifies an obscured joint within the image, and 2 designates a clearly visible joint. Such an annotation protocol affords reason about chimpanzee's orientation and action based on facial joint visibility. For instance, the chimpanzee might be eating something if the two lips are apart. Sample frames showing pose annotations are depicted in \cref{fig:annotation}. Notably, \dataset holds the potential for future expansion to encompass pose tracking tasks, analogous to the PoseTrack \cite{andriluka2018posetrack} for humans.  \\

\begin{table}[t!]
    \centering
    \caption{\textbf{Keypoint definitions for chimpanzee.}} 
    \label{tab:keypointdef}
    \setlength{\tabcolsep}{3pt}
    \begin{tabular}{cl c cl}
        \toprule
        No. & Definition & & No. & Definition\\
        \midrule
        0     & Root of hip & & 8     & Right eye\\
        1     & Right knee & & 9    & Left eye\\
        2     & Right ankle & & 10    & Right shoulder\\
        3     & Left knee  & & 11    & Right elbow\\
        4     & Left ankle  & & 12    & Right wrist\\
        5     & Neck     & & 13  & Left shoulder\\
        6     & Upper lip & & 14    & Left elbow\\
        7     & Lower lip & &15    & Left wrist\\
        \bottomrule
    \end{tabular}%
    \vspace{-1.4em}
\end{table}

\noindent\textbf{Spatiotemporal action detection.\quad}
Spatiotemporal action detection seeks to attribute one or multiple behavioral labels to each bounding box containing a chimpanzee, leveraging the spatiotemporal context within a video clip. Our ethogram, detailed in \cref{fig:ethogram}, delineates 23 nuanced subcategories of behaviors and guides the fine-grained annotations of chimpanzee behavior, such as ``climbing'' within the ``locomotion'' category. Notably, within the realm of social interactions, we meticulously differentiate between the action performer and receiver. For instance, the grooming behavior is bifurcated into ``grooming'' and ``being groomed.'' Every chimpanzee in a frame has its subcategory behavior annotated. It is not uncommon for an individual to simultaneously exhibit multiple behaviors, exemplified by Swela's ``carrying'' and ``moving'' actions in \cref{fig:annotation}. The distribution of these behavioral annotations, visualized in \cref{fig:behavior_dist} on a \textit{log} scale, reveals a long-tail distribution, mirroring the authentic behavioral tendencies of chimpanzees in their natural habitats. We provide more details of the ethogram and the annotation distribution statistics in \cref{supp:sec:ethogram}.

\dataset emerges as an invaluable resource for researchers spanning the domains of primatology, comparative psychology, computer vision, and machine learning. It furnishes a comprehensive and varied array of annotations, paving the way for in-depth analysis of multifaceted chimpanzee behaviors and catalyzing the development of advanced machine learning algorithms. The inherent long-tail distribution presents a formidable challenge for chimpanzee identification and behavior recognition and beckons explorations into few-shot learning in future endeavors.

\begin{figure*}[t!]
    \centering
    \includegraphics[width=\linewidth]{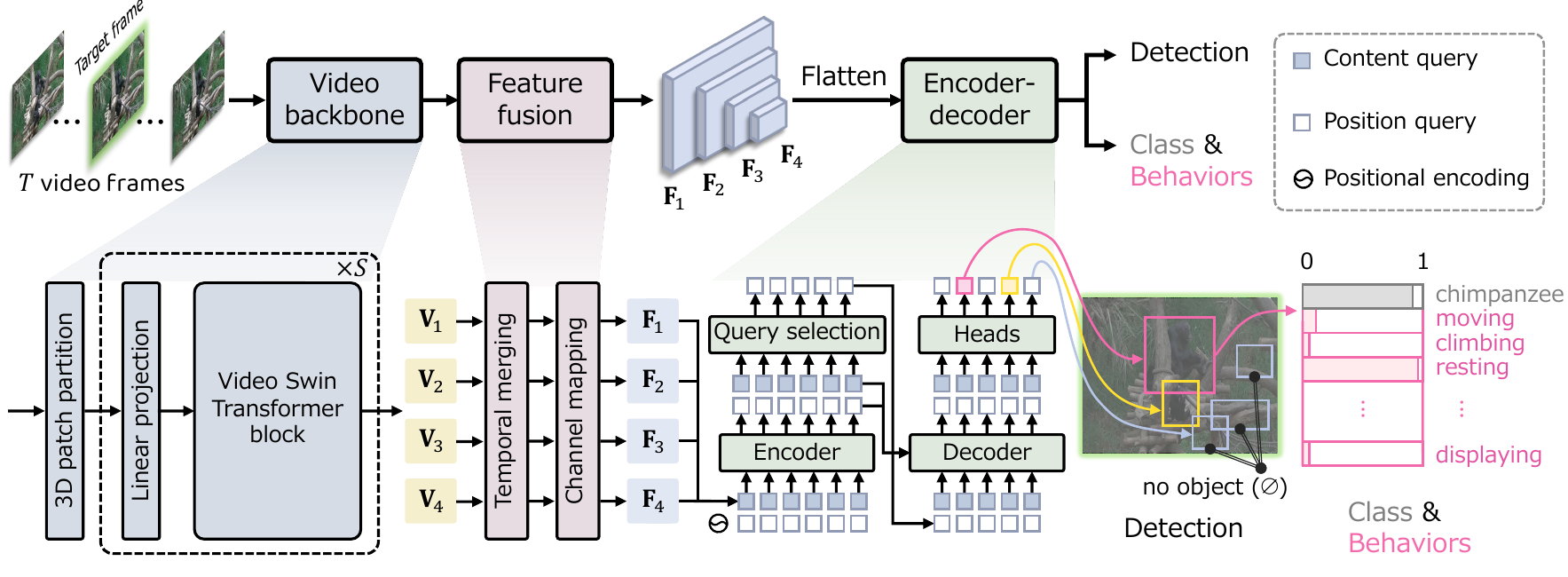}
    \caption{\textbf{Overview of \method.} Given a sequence of adjacent video frames, \method predicts a set of detection boxes for the target frame and simultaneously identifies the class label and behaviors within each bounding box. Initially, a video backbone extracts multi-scale video features $\{\videofeat_i\}_{i=1}^{S}$. These features are then processed by a fusion module that aggregates the temporal context, resulting in $\{\feat_i\}_{i=1}^{S}$. Subsequently, a transformer-based encoder-decoder converts the flattened multi-scale feature tokens and positional encodings into a set of content queries and position queries. A query selection mechanism then chooses a fixed number of encoder features to act as initial position queries for the decoder. Finally, the decoder, equipped with prediction heads, refines these queries to generate the bounding boxes as well as the class labels and behaviors within each box.}%
    \label{fig:pipeline}%
    \vspace{-1.4em}
\end{figure*}

\section{\method}

In this section, we present \method, a comprehensive and unified framework for video-based chimpanzee detection and behavior recognition. \cref{fig:pipeline} provides an overview of \method's architecture. To detect and recognize chimpanzees and their behaviors in a target frame, we leverage temporal context from adjacent video frames. Our approach extracts multi-scale features, $\{\videofeat_i\}_{i=1}^{S}$, using a video backbone network (\cref{sec:backbone}). These features are then fused to obtain $\{\feat_i\}_{i=1}^{S}$ and flattened (\cref{sec:fusion}), creating a unified representation of spatial and temporal information.

The core of \method is an encoder-decoder architecture, inspired by the DETR model \cite{carion2020end} but tailored for our specific task. This Transformer-based structure utilizes a combination of content and position queries to process the fused features (\cref{sec:dino}). The encoder-decoder simultaneously optimizes for chimpanzee detection and behavior classification, enabling \method to identify bounding boxes containing chimpanzees and categorize their behaviors in an end-to-end manner. By integrating these components, \method ensures robust detection, categorization, and behavior analysis of chimpanzees within dynamic video contexts.

\subsection{Multi-scale temporal feature extraction}\label{sec:backbone}

To extract multi-resolution temporal features, we employ Video Swin Transformer \cite{liu2022video} as our backbone. This architecture processes a video sequence of $T$ frames, where each frame is composed of $H \times W \times 3$ pixels. The backbone treats each 3D patch of size $2 \times 4 \times 4 \times 3$ as a token, enabling it to capture both spatial and temporal information effectively.

The initial 3D patch partition layer transforms the input into $\frac{T}{2} \times \frac{H}{4} \times \frac{W}{4}$ 3D tokens, with each token represented by a $\chin$-dimensional feature vector. This transformation sets the stage for subsequent processing through a series of $S$ Video Swin Transformer blocks \cite{liu2022video}, each generating temporal features at different resolutions.

Each block incorporates a linear projection layer, adhering to the standard practice outlined in \cite{liu2022video}. These projection layers perform patch merging along the spatial dimensions, progressively reducing spatial resolution while increasing feature dimensionality. It's worth noting that the linear projection in the first block is an exception, as it maintains the original spatial dimensions.

The output of this process is a set of $S$ multi-scale temporal features, denoted as $\{\videofeat_i\}_{i=1}^{S}$, where each feature $\videofeat_i$ is extracted from its corresponding block. This multi-scale approach allows our model to capture both fine-grained details and broader contextual information, crucial for accurate chimpanzee detection and behavior recognition.

For a more detailed description of the architecture, including specific layer configurations and feature dimensions, we refer readers to \cref{supp:sec:methoddetail}.

\subsection{Temporal feature fusion}\label{sec:fusion}

The multi-scale temporal features extracted by the backbone network provide essential contextual information for predicting the target frame. This section describes the process of fusing these features to create a unified representation that integrates temporal context effectively.

Initially, each feature $\videofeat_i$ in the set $\{\videofeat_i\}_{i=1}^{S}$ has a temporal dimension of size $\frac{T}{2}$, reflecting the temporal extent of the input video sequence. To condense this temporal information, we employ a temporal merging layer. This layer utilizes convolution operations to reduce the temporal dimension of each feature to 1, effectively aggregating information across the time axis.

Following temporal merging, a channel mapping layer is applied. This layer, also implemented using convolution, serves to standardize the feature channels across all scales to a common dimension. The result of this process is a set of $S$ multi-scale features, denoted as $\{\feat_i\}_{i=1}^{S}$, where each feature $\feat_i$ has $\chout$ feature dimensions. For example, $\feat_1 \in \R^{\frac{H}{4} \times \frac{W}{4} \times \chout}$ represents the feature at the first scale, maintaining the spatial dimensions of $\frac{H}{4} \times \frac{W}{4}$ with $\chout$ channels.

These fused features effectively integrate temporal context from the input video sequence, providing a robust foundation for the subsequent tasks of chimpanzee detection and behavior recognition in the target frame. By preserving information across multiple scales while condensing temporal information, this fusion process enables our model to capture both fine-grained details and broader contextual cues necessary for accurate analysis of chimpanzee behavior.

\subsection{Detection, categorization, and behavioral classification}\label{sec:dino}

Building upon advanced DETR-series models \cite{carion2020end,zhu2021deformable,zhang2023dino}, particularly DINO \cite{zhang2023dino}, we develop a comprehensive model for chimpanzee detection and behavior classification. Our approach simultaneously determines the category, location, and behaviors of chimpanzees in the target frame, leveraging a combination of position and content queries to enhance accuracy.  \\

\noindent\textbf{Overview.\quad}
The process begins with cross-scale feature fusion using the Transformer encoder's self-attention mechanism. This integrates information from the multi-level features $\{\feat_i\}_{i=1}^{S}$, capturing both fine-grained details and high-level spatial context. We flatten and concatenate features of different scales to form initial content queries for the encoder, with corresponding positional encodings serving as position queries.

A query selection mechanism then identifies $\querynum$ encoder features as initial position queries for the decoder. The decoder, augmented with auxiliary heads, transforms these position queries into bounding boxes while optimizing content queries for class and behavior determination. This dual-query approach enables precise detection and simultaneous classification of chimpanzees and their behaviors, including fine-grained multi-label prediction for complex actions.

For training, we adopt loss functions common to the DETR series \cite{carion2020end,zhang2023dino} for box regression and classification, employing one-to-one bipartite matching. To address multi-label behavior classification, we implement focal loss \cite{lin2017focal}, which effectively handles class imbalance in such scenarios.  \\

\noindent\textbf{Query selection.\quad}
To optimize position queries efficiently, we adopt a query selection scheme inspired by \cite{zhang2023dino,zhu2021deformable}. This process involves appending a classification head network after the encoder to compute confidence scores for each query, representing the likelihood of containing a chimpanzee. The top $\querynum$ queries with the highest confidence are selected as initial position queries for the decoder.

This selection mechanism serves to focus the model's attention on the most relevant spatial locations, potentially improving both detection accuracy and computational efficiency. By prioritizing high-confidence regions, the decoder can more effectively refine its predictions. The architecture of the classification head network used here is shared with other head networks in our model, details of which are elaborated in the subsequent section.  \\

\noindent\textbf{Prediction heads.\quad}
Our model employs two specialized prediction heads: one for bounding box regression (detection) and another for behavior classification. Both are implemented as \acp{mlp} and utilize refined query features from the last decoder layer's output.

The box regression head transforms position queries into $\allbbox = \{\bbox_q\}^{\querynum}_{q=1} \in \R^{\querynum \times 4}$, where each $\bbox_q \in [0, 1]^4$ represents the predicted box position for the $q^{th}$ query. Box positions are defined using center coordinates, height, and width, all relative to the image size.

The behavior head network converts content queries into two outputs: $\allcls = \{\cls_q\}^{\querynum}_{q=1} \in \R^{\querynum \times 1}$, representing class probabilities, and $\allbehave = \{\action_q\}^{\querynum}_{q=1} \in \R^{\querynum \times \actionnum}$, representing behavior probabilities, where $\actionnum = 23$ is the total number of behavior classes. Here, $\cls_q \in [0, 1]$ provides the confidence score for the $q^{th}$ query, indicating the probability of a chimpanzee in the bounding box. $\action_q \in [0,1]^K$ is a multi-label probability vector obtained using the Sigmoid function, where each dimension represents the probability of a corresponding behavior class.

For query selection, we integrate an additional behavior head following the encoder. This head generates class probabilities used as confidence scores for selecting queries, as discussed in the previous section.   \\

\noindent\textbf{Training losses.\quad}
Our training process follows established practices in object detection \cite{carion2020end,zhang2023dino}. We begin with bipartite matching based on bounding box positions and class labels to establish a one-to-one correspondence between predicted and \ac{gt} sets. This matching ensures that each prediction is uniquely associated with a \ac{gt} object, facilitating more effective learning.

\begin{table*}[t!]
    \center
    \caption{\textbf{Results of the detection, tracking, and \acs{reid} track on the \dataset test set.} The row highlighted in light blue is the performance reference on the human tracking dataset \acs{mot}-17 \cite{milan2016mot16}. $-$ denotes unreported. Results for the input video at two different resolutions are reported.}
    \label{tab:tracking}
    \setlength{\tabcolsep}{11pt}
    \resizebox{\linewidth}{!}{%
        \begin{tabular}{l c c c c c c c c  c }
            \toprule 
            Method & HOTA $\uparrow$ & \ac{mot}A $\uparrow$ & \ac{mot}P $\uparrow$ & IDF1 $\uparrow$ & mAP $\uparrow$ & nFP $\downarrow$ & nFN $\downarrow$ & nIDs $\downarrow$\\
            \toprule 
            \rowcolor{Azure} OC-SORT \cite{cao2023observation} & 63.2 & 78.0 & $-$ & 77.5 & $-$ & 2.7 & 19.0 & 0.3 \\
            \midrule
            \hline
            \rowcolor{ggrey} \textit{Resolution: 1440 $\times$ 800}  & & & & & & & &\\ 
            SORT \cite{bewley2016simple} & 39.8 & 43.2 & 20.3 & 37.7 & 71.4 & 16.1 & 37.8 & 2.8\\
            \midrule
            DeepSORT \cite{wojke2017simple} & 40.2 & 43.2 & 20.3 & 38.4 & 71.4 & 16.1 & 37.8 & 2.9\\
            \midrule
            Tracktor \cite{bergmann2019tracking} & 49.5 & 50.5 & 22.6 & 55.6 & 70.7 & 13.8 & 35.2 & 0.5\\
            \midrule
            QDTrack \cite{pang2021quasi} & 50.3 & 54.2 & 22.2 & 55.8 & 77.8 & 19.7 & 24.6 & 1.4\\
            \midrule
            ByteTrack \cite{zhang2022bytetrack} & 49.2 & 43.9 & 20.3 & 55.2 & 70.3 & 18.0 & 37.4 & 0.7\\
            \midrule
            OC-SORT \cite{cao2023observation} & 47.9 & 42.1 & 20.5 & 53.3 & 70.5 & 20.3 & 36.6 & 1.1\\
            \midrule
            \hline
            \rowcolor{ggrey} \textit{Resolution: 576 $\times$ 576} & & & & & & & &\\ 
            SORT \cite{bewley2016simple} & 27.2 & 34.4 & 23.1 & 24.0 & 63.7 & 12.4 & 48.8 & 4.1\\
            \midrule
            DeepSORT \cite{wojke2017simple} & 30.8 & 31.8 & 23.0 & 32.4 & 63.7 & 12.1 & 48.6 & 7.1\\
            \midrule
            QDTrack \cite{pang2021quasi} & 47.9 & 46.5 & 24.1 & 52.6 & 73.9 & 24.8 & 27.2 & 1.4 \\
            \midrule
            ByteTrack \cite{zhang2022bytetrack} & 38.0 & 35.4 & \textbf{25.7} & 45.2 & 58.5 & \textbf{9.3} & 49.4 & 0.7\\
            \midrule
            OC-SORT \cite{cao2023observation} & 41.5 & 39.1 & 25.1 & 47.2 & 64.4 & 14.2 & 45.3 & 1.2\\
            \midrule
            \textbf{\method} (Ours)  & \textbf{56.3} & \textbf{60.0} & 21.6 & \textbf{65.6} & \textbf{75.2} & 14.2 & \textbf{25.1} & \textbf{0.5} \\
            \bottomrule 
        \end{tabular}%
    }%
\end{table*}

Following the matching, we apply set prediction losses for both box regression and classification. For box regression, we employ a combination of $L1$ loss and \ac{giou} loss \cite{rezatofighi2019generalized}. The $L1$ loss addresses direct positional errors, while \ac{giou} loss captures the overall geometric similarity between predicted and \ac{gt} boxes. For classification, including both object class and behavior classes, we utilize focal loss \cite{lin2017focal}. This choice is particularly effective for handling class imbalance, which is common in multi-label classification scenarios like behavior recognition.

This comprehensive loss formulation ensures that our model learns to accurately localize chimpanzees while simultaneously classifying their behaviors, addressing the multi-faceted nature of our task.

\section{Benchmarking \textbf{\dataset}}

To rigorously assess \dataset, we benchmark a suite of representative methods across the aforementioned three tracks: (i) detection, tracking, and ReID, (ii) pose estimation, and (iii) spatiotemporal action detection. Our computational framework leverages four NVIDIA GeForce RTX 3090 GPUs (24GB) for both training and evaluation across all tracks. In the subsequent sections, we delve into the implementation details, baseline methods, and evaluation metrics for each track. Through this benchmarking, we aim to establish baseline performance levels, identify key challenges, and provide insights that will guide future research efforts in automating chimpanzee behavior using \dataset.

\subsection{Detection, tracking, and \texorpdfstring{\ac{reid}}{}}\label{sec:exp-tracking}

\noindent\textbf{Setting.\quad}
We evaluate several prominent \ac{mot} algorithms on \dataset, including both classical methods such as SORT \cite{bewley2016simple}, DeepSORT \cite{wojke2017simple}, and Tracktor \cite{bergmann2019tracking}, as well as the \ac{sota} methods such as ByteTrack \cite{zhang2022bytetrack}, and OC-SORT \cite{cao2023observation}. All implementations are based on the MMTracking \cite{mmtrack2020} codebase. The detection backbone is YOLOX \cite{ge2021yolox}. Each method undergoes training for 10 epochs, adhering to the official configurations, which encompass optimizer settings, batch size, data augmentation techniques, and pre-trained models. Given that the three classical methods \cite{bewley2016simple,wojke2017simple,bergmann2019tracking} lack inherent \ac{reid} modules, we supplement with a dedicated \ac{reid} network built on ResNet-50 \cite{he2016deep}. The training curves of these methods provided in the \cref{supp:fig:supp_val_tracking} affirm convergence within the training epochs.

We split the video clips in \dataset into 80\% train, 10\% validation, and 10\% test. Both the train set and test set cover all the individuals. Models are trained on the training set, with performance metrics reported on the test set. We employ widely-accepted evaluation metrics, drawing from convention in human/object detection, tracking, and \ac{reid} \cite{bewley2016simple,pang2021quasi,zhang2022bytetrack}. Specifically, we utilize (i) \ac{map} \cite{lin2014microsoft} to gauge the detection accuracy, and (ii) the CLEAR metrics \cite{bernardin2008evaluating} (\ac{mota}, \ac{motp}, \ac{fp}, \ac{fn}, \ac{id}s), \ac{idf1} \cite{ristani2016performance}, and \ac{hota} \cite{luiten2021hota} to evaluate various facets of the tracking performance. It is worth noting that for \ac{fp}, \ac{fn}, and \ac{id}s, we report normalized values and denote these metrics as n\ac{fp}, n\ac{fn}, and n\ac{id}s, respectively.  \\

\begin{table*}[t!]
    \center
    \caption{\textbf{Results of the pose estimation track on \dataset test set.} The row highlighted in light blue is the performance reference on the human pose estimation dataset COCO \cite{lin2014microsoft}. $-$ denotes unreported.}
    \label{tab:pose}
    \setlength{\tabcolsep}{9pt}
    \resizebox{\linewidth}{!}{%
        \begin{tabular}{l l l c c c c c c c c}
            \toprule 
            & Method & Backbone & PCK@0.05 & PCK@0.1 & AP & AP$^{50}$ & AP$^{75}$ & AP$^M$ & AP$^L$ & AR\\
            \toprule 
            \rowcolor{Azure}
            & \parbox{2.2cm}{HRNet \cite{sun2019deep}} & HRNet-W32 & $-$ & $-$ & 74.4 & 90.5 & 81.9 & 70.8 & 81.0 & 79.8\\
            \midrule
            & \multirow{3}{*}{\parbox{2.2cm}{SimpleBaseline\\\cite{xiao2018simple}}} & ResNet-50 & 25.3 & 46.2 & 8.6 & 27.4 & 3.9 & 0.3 & 12.5 & 17.3\\ 
            & & ResNet-101 & 26.2 & 46.4 & 8.7 & 27.5 & 4.2 & 0.3 & 12.9 & 17.7\\ 
            & & ResNet-152 & 26.3 & 47.3 & 9.3 & 29.2 & 4.7 & 0.5 & 13.4 & 18.6\\ 
            \cmidrule{2-11}
            & \multirow{4}{*}{\parbox{2.2cm}{RLE \cite{li2021human}}} & MobileNetV2 & 27.5 & 48.1 & 16.7 & 43.1 & 11.1 & 2.0 & 17.7 & 19.5\\ 
            & & ResNet-50 & 28.2 & 47.1 & 16.3 & 41.2 & 11.4 & 1.3 & 17.4 & 20.0\\
            & & ResNet-101 & 28.2 & 46.5 & 16.2 & 41.1 & 10.8 & 2.1 & 17.3 & 20.1\\
            \multirow{-7}{*}{\rotatebox{90}{\textit{Regression}}} & & ResNet-152 & 30.0 & 48.4 & 18.1 & 43.0 & 13.5 & 1.4 & 19.2 & 22.3\\
            \midrule 
            & \parbox{2.2cm}{CPM \cite{wei2016convolutional}} & CPM & 40.7 & 60.4 & 21.6 & 51.0 & 17.1 & 9.5 & 22.4 & 25.4\\
            \cmidrule{2-11}
            & \parbox{2.2cm}{Hourglass \cite{newell2016stacked}} & Hourglass-4 & 44.6 & 60.8 & 20.6 & 48.9 & 16.0 & 4.6 & 23.7 & 28.2\\
            \cmidrule{2-11}
            & \parbox{2.3cm}{MobileNetV2 \cite{sandler2018mobilenetv2}} & MobileNetV2 & 39.8 & 59.4 & 19.4 & 48.5 & 14.3 & 2.3 & 20.6 & 23.2\\ 
            \cmidrule{2-11}
            & \multirow{3}{*}{\parbox{2.2cm}{SimpleBaseline\\\cite{xiao2018simple}}} & ResNet-50 & 43.3 & 61.7 & 22.1 & 51.5 & 17.7 & 3.7 & 23.4 & 26.3\\ 
            & & ResNet-101 & 42.8 & 60.7 & 21.7 & 52.5 & 16.7 & 4.3 & 23.0 & 26.2\\ 
            & & ResNet-152 & 43.9 & 61.6 & 22.7 & 53.4 & 18.3 & 5.3 & 23.9 & 27.1\\
            \cmidrule{2-11}
            & \multirow{2}{*}{\parbox{2.2cm}{HRNet \cite{sun2019deep}}} & HRNet-W32 & 48.6 & 65.6 & 25.9 & 58.2 & 22.1 & 6.1 & 27.0 & 30.3\\
            & & HRNet-W48 & 47.3 & 64.5 & 25.1 & 57.2 & 21.0 & 6.9 & 26.2 & 29.6\\ 
            \cmidrule{2-11}
            & \multirow{5}{*}{\parbox{2.2cm}{DarkPose \cite{zhang2020distribution}}} & ResNet-50 & 43.7 & 62.1 & 22.8 & 53.8 & 18.8 & 3.4 & 24.1 & 27.1\\  
            & & ResNet-101 & 43.1 & 61.2 & 22.1 & 52.6 & 17.6 & 4.0 & 23.4 & 26.5\\  
            & & ResNet-152 & 43.5 & 61.2 & 22.4 & 53.2 & 17.4 & 4.6 & 23.7 & 26.7\\
            & & HRNet-W32 & 48.7 & 65.6 & 25.7 & 58.4 & 21.3 & 5.6 & 26.9 & 30.1\\ 
            & & HRNet-W48 & 47.6 & 64.5 & 25.8 & 58.0 & 21.5 & 6.6 & 27.0 & 30.2\\ 
            \cmidrule{2-11}
            & \multirow{2}{*}{\parbox{2.2cm}{HRFormer \cite{yuan2021hrformer}}} & HRFormer-S & 45.1 & 61.4 & 23.0 & 53.1 & 19.7 & 5.5 & 24.1 & 27.1\\ 
            \multirow{-15}{*}{\rotatebox{90}{\textit{Heatmap-based}}} & & HRFormer-B & 46.4 & 63.0 & 24.1 & 55.3 & 20.1 & 5.2 & 25.4 & 28.2\\ 
            \bottomrule 
        \end{tabular}%
    }%
\end{table*}

\noindent\textbf{Results.\quad}
\cref{tab:tracking} presents the performance of various pose estimation methods on the \dataset test set. Each method was evaluated over three independent runs, with the reported values representing the average results. For a comprehensive analysis of result stability, including variance measures, we refer readers to our conference paper \cite{ma2023chimpact}. A holistic view of the results reveals that QDTrack \cite{pang2021quasi} emerges as the top performer. However, it does suffer from a higher count of identity switches compared to other methods. In terms of detection performance, the YOLOX algorithm \cite{ge2021yolox} stands toe-to-toe with Faster R-CNN \cite{ren2015faster}. A discernible trend is evident among contemporary tracking methods, which seem to excel in identity association capabilities over their classical counterparts. This is corroborated by marked improvements in tracking metrics like IDF1 and IDs. Such a trend intimates that the latest tracking methods might be adept at maintaining consistent object identities, a pivotal aspect when tracking and analyzing individual trajectories within chimpanzee cohorts.

While the results garnered by the array of tracking algorithms are commendable, they still lag behind the benchmarks set on human-centric datasets \cite{zhang2022bytetrack,pang2021quasi,cao2023observation}. This disparity can be attributed to challenges like the low contrast and low color variation of the body fur of chimpanzees, compounded by intricate self-occlusions. Nonetheless, this very observation accentuates the significance of \dataset. It not only offers a challenging arena for tracking algorithms but also stands as an ideal platform for pioneering and refining tracking methods tailored for chimpanzees and other non-human primates.

\subsection{Pose estimation}

\noindent\textbf{Setting.\quad}
We benchmark several \ac{sota} human pose estimation methods on \dataset, such as \ac{hrnet} \cite{sun2019deep} and DarkPose \cite{zhang2020distribution}. These methods represent both heatmap-based and regression-based paradigms, providing a comprehensive evaluation across different pose estimation approaches. We implement these methods using the MMPose \cite{mmpose2020} framework to ensure consistency in our experimental setup. All models undergo training for 210 epochs, adhering to their respective official configurations for optimizers, batch sizes, and learning rates. This extended training period allows for thorough model convergence. To assess model performance and verify the absence of overfitting, we present validation curves on the AP metric in \cref{supp:fig:supp_val_pose}. Detailed implementation specifics are available in \cref{supp:sec:experiment}.

For evaluation, we maintain the same train/test partition as in the tracking task. We employ \ac{map} with various thresholds, following conventions in human pose estimation \cite{lin2014microsoft}. Additionally, we incorporate the \ac{pck} metric \cite{andriluka20142d,ng2022animal}, which quantifies the fraction of accurately predicted keypoints within a distance threshold. Specifically, PCK@$\delta$ uses a threshold defined as $\delta \times max(height, width)$ of the chimpanzee's bounding box. This metric provides insights into body joint localization accuracy, allowing us to assess the models' performance on chimpanzee anatomy.  \\

\noindent\textbf{Results.\quad}
\cref{tab:pose} consolidates these pose estimators' performances on the \dataset test set. Notably, the heatmap-based DarkPose \cite{zhang2020distribution} with an HRNet \cite{sun2019deep} backbone emerges as the top-performing model. This trend aligns with observations in human pose estimation, where heatmap-centric methods \cite{wei2016convolutional,xiao2018simple,newell2016stacked,sun2019deep} predominantly lead the pack, attributed to their robustness against pose and appearance variations. However, the heatmap representation may be less accurate in scenarios where multiple joints are occluded or closely spaced, and it demands heftier computational and memory resources.
Conversely, the newer regression-based methods \cite{li2021human} are computationally leaner but tend to be more susceptible to overfitting and generally lag in performance.

These results underscore that the task of chimpanzee pose estimation is distinct and nuanced, and cannot be seamlessly addressed by merely repurposing human-centric pose estimation methods. We believe there are two primary reasons for this: (i) chimpanzees exhibit unique joint flexibility and a broader range of motion, and (ii) the visual texture and appearance of chimpanzee fur diverge significantly from human skin. These insights emphasize the need for chimpanzee specific pose estimation strategies.

\begin{table*}[t!]
    \center
    \caption{\textbf{Results of spatiotemporal action detection track on \dataset test set.} The row highlighted in light blue is the performance reference on the human action dataset AVA \cite{gu2018ava}. ``\textit{with \ac{gt} box}'' and ``\textit{with Det. box}'' mean using \ac{gt} bounding boxes or detected boxes, respectively. ``\textit{w.} NL/Max/Avg LFB'' denotes using non-local, max, or average LFB module. ``\textit{w.} Ctx'' indicates using both the RoI feature and the global pooled feature for classification. ``mAP,'' ``mAP$_L$,'' ``mAP$_O$,'' and ``mAP$_S$ represent the overall mAP and mAP for $\underline{L}$ocomotion, $\underline{O}$bject interaction, and $\underline{S}$ocial interaction. $\oslash$ denotes not applicable. $-$ denotes unreported. }
    \label{tab:action}
    \setlength{\tabcolsep}{23pt}
    \resizebox{\linewidth}{!}{%
        \begin{tabular}{l l c c c c}
            \toprule
            Method & Module & mAP & mAP$_L$ & mAP$_O$ & mAP$_S$\\
            \toprule
            \rowcolor{Azure}
            SlowFast \cite{feichtenhofer2019slowfast} & & 25.8 & $\oslash$ & $\oslash$ & $\oslash$\\
            \rowcolor{Azure}
            VideoPrism-B \cite{zhao2024videoprism} & & 30.6 & $\oslash$ & $\oslash$ & $\oslash$\\
            \rowcolor{Azure}
            VideoPrism-g \cite{zhao2024videoprism} & & 36.2 & $\oslash$ & $\oslash$ & $\oslash$\\
            \midrule
            \hline
            \rowcolor{ggrey} \textit{with \ac{gt} box} & & & & &\\ 
            %\midrule
            ACRN \cite{sun2018actor} & & 24.4 & 58.7 & 33.8 & 14.7 \\
            \midrule
            & \textit{w.} NL LFB & 22.0 & 50.1 & 32.3 & 13.5 \\
            & \textit{w.} Max LFB & 23.2 & 45.0 & 31.2 & 17.7 \\
            \multirow{-3}{*}{LFB \cite{wu2019long}} & \textit{w.} Avg LFB & 21.3 & 45.0 & 29.8 & 14.7 \\
            \midrule
            & & 20.9 & 48.1 & 36.2 & 11.5 \\
            \multirow{-2}{*}{SlowOnly \cite{feichtenhofer2019slowfast}} & \textit{w.} Ctx & 22.3 & 52.3 & 31.2 & 13.8 \\
            \midrule
            & & 21.9 & 53.0 & 30.6 & 12.9 \\
            \multirow{-2}{*}{SlowFast \cite{feichtenhofer2019slowfast}} & \textit{w.} Ctx & 24.3 & 56.8 & 31.5 & 15.6 \\
            \midrule
            CoCa-B \cite{yu2022coca} & & 12.6 & $-$ & $-$ & $-$ \\
            \midrule
            InternVideo-B \cite{wang2022internvideo} & & 24.0 & $-$ & $-$ & $-$ \\
            InternVideo-L \cite{wang2022internvideo} & & 25.7 & $-$ & $-$ & $-$\\
            \midrule
            UMT-B \cite{li2023unmasked} & & 25.0 & $-$ & $-$ & $-$ \\
            UMT-L \cite{li2023unmasked} & & 24.7 & $-$ & $-$ & $-$ \\
            \midrule
            VideoPrism-B \cite{zhao2024videoprism} & & 28.8 & $-$ & $-$ & $-$ \\
            VideoPrism-g \cite{zhao2024videoprism} & & 31.5 & $-$ & $-$ & $-$ \\
            \midrule
            \hline
            \rowcolor{ggrey} \textit{with Det. box} & & & & & \\ 
            ACRN \cite{sun2018actor} &  & 13.4 & 26.8 & 14.4 & 7.1 \\
            \midrule
            &  & 11.8 & 25.8 & 13.1 & 5.2 \\
            \multirow{-2}{*}{SlowOnly \cite{feichtenhofer2019slowfast}} & \textit{w.} Ctx  & 13.9 & 27.4 & 14.4 & 7.7 \\
            \midrule
            &  & 13.5 & 27.2 & 13.7 & 7.3 \\
            \multirow{-2}{*}{SlowFast \cite{feichtenhofer2019slowfast}} & \textit{w.} Ctx   & 16.2 & 27.5 & 14.3 & 11.9 \\ 
            \midrule
            \textbf{\method} & & \textbf{34.3} & \textbf{50.3} & \textbf{31.3} & \textbf{29.3}   \\
            \bottomrule 
        \end{tabular}%
    }%
\end{table*}

\subsection{Spatiotemporal action detection}

\noindent\textbf{Setting.\quad}
We benchmark several representative human action detection baselines on \dataset using the MMAction2 \cite{2020mmaction2} codebase. The evaluated methods include ARCN \cite{sun2018actor}, LFB \cite{wu2019long}, and SlowFast with its variant SlowOnly \cite{feichtenhofer2019slowfast}. These methods represent a range of approaches in spatiotemporal action detection, from recurrent networks to long-term feature banks and multi-pathway architectures. All models undergo training for 20 epochs with a batch size of 32, maintaining consistent optimizers and learning rates as in their official implementations. To verify convergence and assess training stability, we provide convergence curves in \cref{supp:fig:supp_val_action}. In addition to these baselines, we report results from recent \ac{sota} methods \cite{yu2022coca,wang2022internvideo,li2023unmasked,zhao2024videoprism}, some of which utilize foundation models. For instance, VideoPrism \cite{zhao2024videoprism} is a video understanding foundation model. The reported numbers for these methods are sourced from VideoPrism \cite{zhao2024videoprism}, which adheres to the same training and evaluation protocols as our benchmark.

To ensure fair comparison, we provide \ac{gt} bounding boxes for each chimpanzee during both training and testing, following the protocol established in \cite{tang2020asynchronous}. This approach allows us to focus on the action detection performance without the confounding factor of object detection accuracy. For researchers interested in exploring ablative modules or implementation details, we direct them to \cref{supp:sec:experiment}. We maintain consistency with previous tracks by adopting the same train-test split. Performance evaluation is conducted using mAP across all 23 action classes, adhering to standard practices in the field \cite{feichtenhofer2019slowfast,tang2020asynchronous}. To provide a more nuanced understanding of model performance, we also evaluate the mAP within each of the four behavioral types separately. This granular analysis allows us to identify strengths or weaknesses of different methods across various categories of chimpanzee behavior.  \\

\noindent\textbf{Results.\quad}
\cref{tab:action} (middle block ``\textit{with \ac{gt} box}'') summarizes the action detection algorithms' performances on the \dataset test set. The overall mAP aligns with results on human action datasets, demonstrating the feasibility of automated action detection for video coding and further analyses of chimpanzee behavior. Notably, locomotion behaviors achieve a higher \ac{map}, likely due to their solitary nature and distinct patterns, while object manipulation tasks show lower performance. For detailed \ac{map} of the ``others'' category, which registers the lowest performance (almost 0\% accuracy) due to limited data, we refer readers to our previous work \cite{ma2023chimpact}. This category comprises just 0.14\% of action instances across two fine-grained classes. This imbalance suggests potential benefits from applying few-shot learning methods in future research.

The latest method, VideoPrism \cite{zhao2024videoprism}, which utilizes large models, shows significant improvement over previous approaches. This advancement underscores \dataset's effectiveness in evaluating and driving model performance, while also highlighting its complexity and diversity that challenge current technologies, positioning it as a valuable platform for advancing spatiotemporal action detection algorithms. The dataset's challenges in capturing diverse chimpanzee behaviors provide opportunities for developing more robust and adaptable models. We anticipate that \dataset will further promote studies into the social dynamics of non-human primates in semi-naturalistic environments, bridging the gap between computer vision advancements and primate behavior research.

\section{\textbf{\method} is the new \texorpdfstring{\ac{sota}}{}}

\subsection{Implementation details}

Our \method framework processes video sequences of $T=8$ frames. To ensure fair comparison across different tasks, we train two model variants with distinct input resolutions. For tracking, we use a resolution of $576 \times 576$, while for spatiotemporal action detection, we employ a $256 \times 256$ resolution. These resolutions align with existing methods in their respective tracks, facilitating fair comparisons. We employ Swin-L \cite{liu2022video} as the video backbone, comprising $S=4$ Swin Transformer blocks. The Transformer architecture consists of 12 encoder and 12 decoder layers, incorporating 4 reference points \cite{zhu2021deformable} in the deformable attention module. Detailed architectural specifications are available in \cref{supp:sec:methoddetail}. Based on our dataset analysis revealing an average of 3 chimpanzees per image, with a maximum of 9, we set the decoder's fixed query number to $\querynum=10$. We use the \dataset training set with a batch size of 64. The model is optimized using Adam with a learning rate of $1e-4$ for 20K iterations. 8 HGX-A800-SXM GPUs are used in the training. 

\subsection{Improvements over prior methods}

\noindent\textbf{Detection, tracking, and \ac{reid}.\quad}
As shown in \cref{tab:tracking}, \method demonstrates significant improvements over \ac{sota} methods in tracking performance. At equivalent resolutions, our approach achieves approximately 10\% higher \ac{hota} scores, indicating superior overall tracking capability. The impressive \ac{map} scores highlight \method's excellent detection performance, while the nID metric underscores its robust tracking stability.

Notably, \method remains highly competitive and often outperforms higher-resolution methods, despite using lower-resolution input videos. This suggests that our approach effectively compensates for reduced spatial information by leveraging temporal cues and multi-task learning. Such performance indicates \method's potential for real-world applications where computational resources or video quality may be limited.  \\

\noindent\textbf{Spatiotemporal action detection.\quad}
\cref{tab:action} (``\textit{with Det. box}'') compares \method's action detection performance with existing \ac{sota}s. Due to \method's end-to-end design for simultaneous bounding box and behavior category prediction, we cannot directly use \ac{gt} bounding boxes as input. To ensure fair comparison, we reproduce open-source methods using detected boxes from our trained \method. The results reveal \method's substantial improvement over existing algorithms, achieving a 20\% increase in overall \ac{map}. This advancement is particularly evident in the challenging ``social'' category, where \method shows a 20\% enhancement. Our method's success stems from its innovative architecture, which integrates temporal feature fusion with the Transformer's self-attention mechanism.

Moreover, our \method outperforms even the latest methods like VideoPrism \cite{zhao2024videoprism}, which utilizes foundation models and extensive pretraining along with \ac{gt} boxes as input. This remarkable performance underscores our tailored design for socially interactive animals like chimpanzees. The ability to accurately detect and classify social behaviors of our \method is crucial for understanding chimpanzee social dynamics and could have significant implications for primatology research and conservation efforts.

\subsection{Ablation study}

\begin{table}[t!]
    \centering
    \caption{\textbf{Results of different frame lengths $T$ of the video input on the \dataset test set for behavior classification.}}
    \label{tab:videolength}
    \setlength{\tabcolsep}{12pt}
    \begin{tabular}{l c c c c}
        \toprule
        $T$ & mAP & mAP$_L$ & mAP$_O$ & mAP$_S$ \\
        \midrule
        4 & 32.3 & 44.5 & 31.8 & 28.0\\
        8 & \textbf{34.3} & \textbf{50.3} & 31.3 & \textbf{29.3}\\
        16 & 34.0 & 49.1 & \textbf{36.5} & 27.8\\
        \bottomrule 
    \end{tabular}%
\end{table}

\begin{figure}[t!]
    \centering
    \includegraphics[width=.85\linewidth]{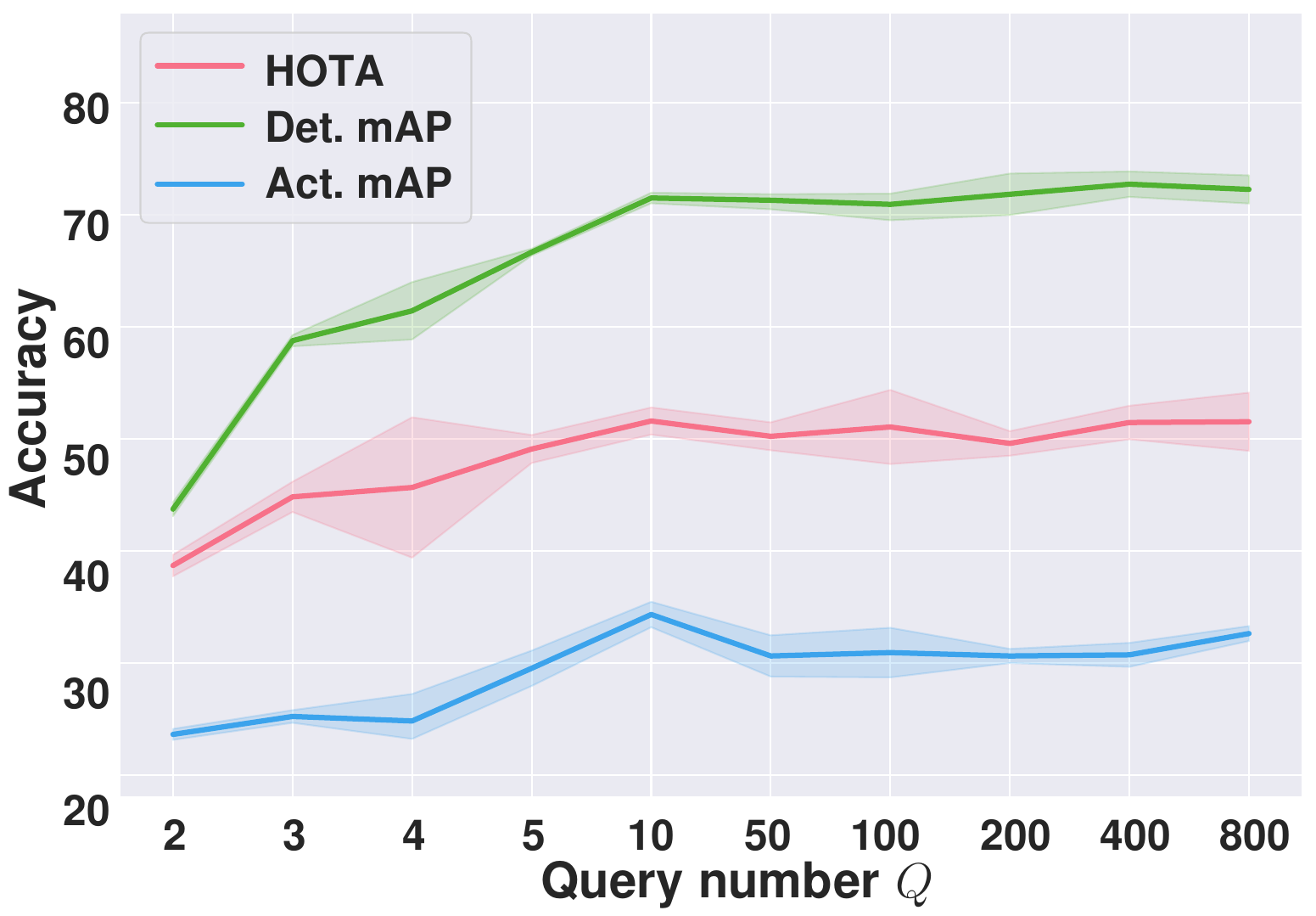}
    \caption{\textbf{Results of different query number $\querynum$ on tracking and spatiotemporal action detection.} We plot HOTA, detection mAP (Det. mAP), and the overall action detection mAP (Act. mAP). Performance consistently improves up to $\querynum=10$, where it stabilizes, balancing efficiency and stability.}%
    \label{fig:query_num_meanstd}%
    \vspace{-0.8em}
\end{figure}

\noindent\textbf{Video input frames $T$.\quad}
\cref{tab:videolength} presents the ablation results for varying frame length $T$ of the input video. Using only 4 video frames yields lower performance compared to 8 frames, as a larger temporal window provides more temporal information. However, increasing the number of input frames beyond 8 leads to a slight decline in performance. This may be due to additional frames introducing more complex behavioral changes, making accurate estimation more challenging. The results suggest that 8 frames strike an optimal balance between temporal context and performance.   \\

\begin{figure*}[t!]
    \centering
    \includegraphics[width=.85\linewidth]{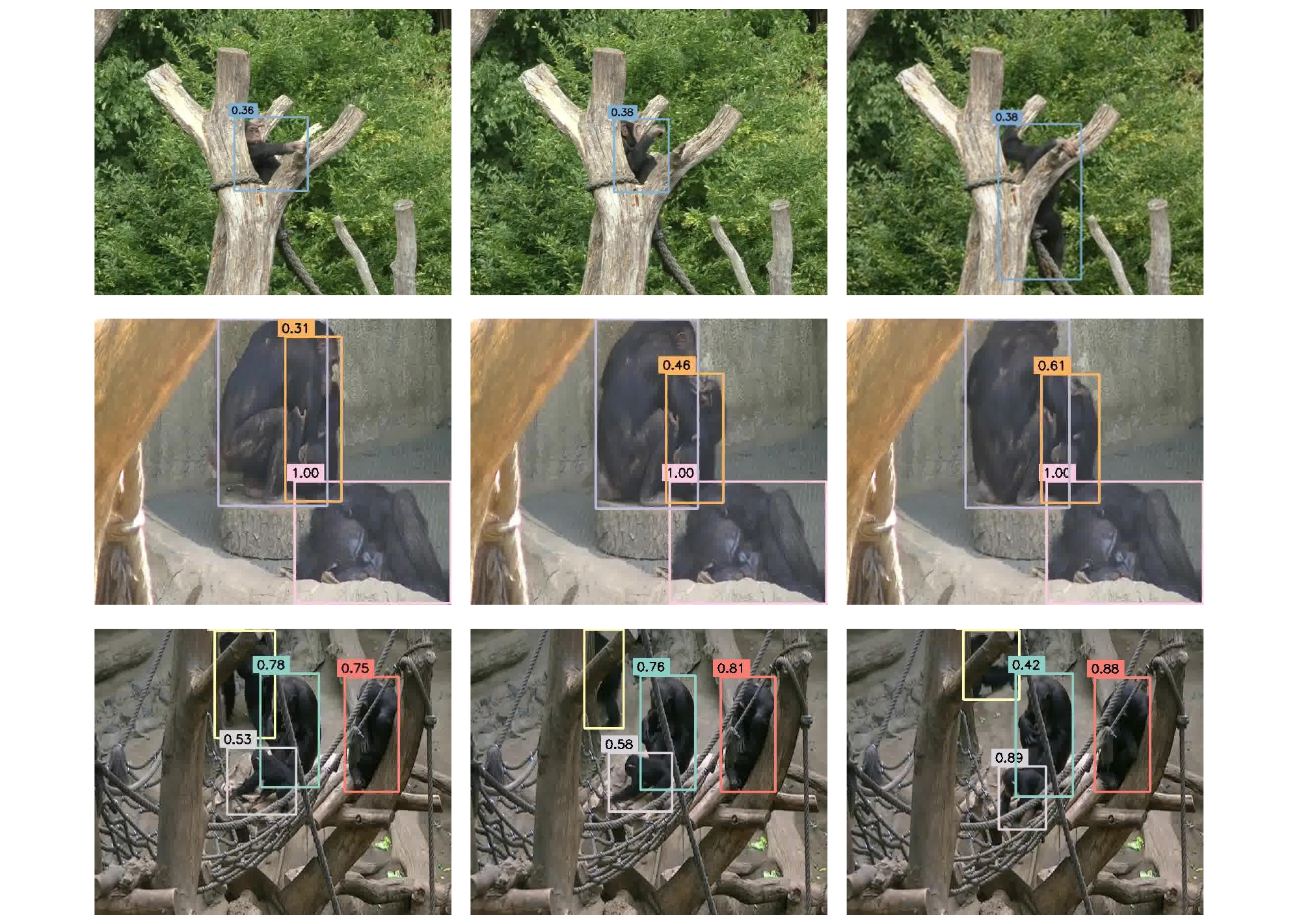}
    \caption{\textbf{Visualization of \method's detection and tracking results on \dataset test set.} Consistent colored boxes denote successful tracking of the same chimpanzee, while numbers indicate chimpanzee classification confidence scores}%
    \vspace{-1.0em}
    \label{fig:track_success}%
\end{figure*}

\begin{figure}[t!]
    \centering
    \includegraphics[width=\linewidth]{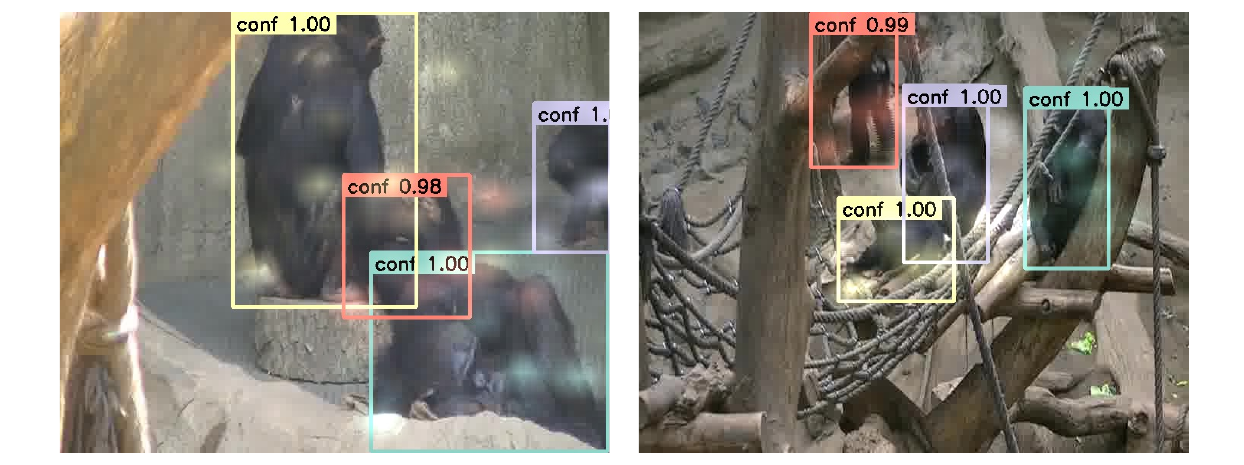}
    \caption{\textbf{Visualization of reference points in the deformable attention module.} Each query's reference points are colored to match its box, with blurring applied according to attention weights, where brighter points indicate higher significance. }%
    \vspace{-1.0em}
    \label{fig:vis_attention}%
\end{figure}

\noindent\textbf{Query number $\querynum$.\quad}
\cref{fig:query_num_meanstd} illustrates the impact of different query numbers $\querynum$ in query selection. We plot the mean and variance for \ac{hota}, detection \ac{map} (Det. \ac{map}), and action detection \ac{map} (Act. \ac{map}). Based on \dataset statistics, there is an average of 3 chimpanzees per frame, with a maximum of 9. As expected, we observe rapid improvement in model performance as $\querynum$ increases up to 10, after which performance stabilizes. To balance computation and model stability, we choose $\querynum=10$ for our experiments. This value results in low variance with stable and superior performance across all metrics. It effectively accommodates the typical range of chimpanzees in scenes while providing headroom for more crowded scenarios.

\subsection{Qualitative results}

\cref{fig:track_success} shows our model's detection and tracking results. Each row represents a sequence from a \dataset test clip. The numbers above each box indicate categorization confidence, while consistently colored boxes denote successful tracking of individuals. Our model shows robustness in handling occlusions: in the second row, an infant chimpanzee (orange box) is slightly obscured by an adult, and in the third row, a chimpanzee (yellow box) is mostly hidden by a tree. Despite these challenges, our method maintains accurate detection and consistent tracking. We further visualize the reference points within the deformable attention module in \cref{fig:vis_attention}, with each point blurred according to its attention weights for enhanced visualization. Notably, these reference points predominantly focus on keypoint areas of each chimpanzee. This suggests that joint regions may exhibit distinct and unique patterns crucial for chimpanzee categorization and differentiation from other objects.

\begin{figure*}[t!]
    \centering
    \includegraphics[width=.85\linewidth]{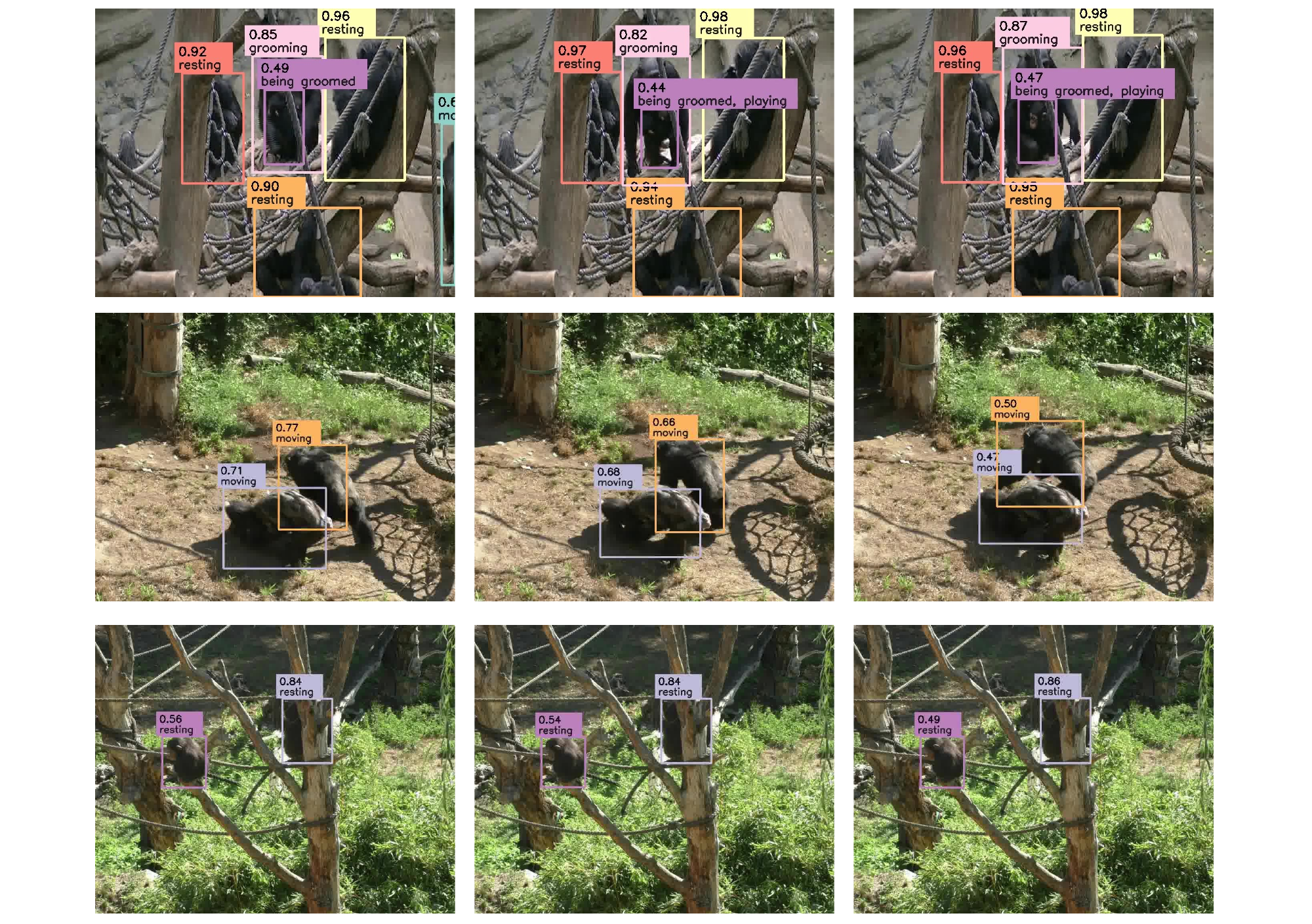}
    \caption{\textbf{Visualization of \method's tracking and behavior detection results on \dataset test set.} Consistent colored boxes denote successful tracking of the same chimpanzee, while numbers indicate chimpanzee classification confidence scores.}%
    \vspace{-1.4em}
    \label{fig:quality_result}%
\end{figure*}

\begin{figure}[t!]
    \centering
    \includegraphics[width=\linewidth]{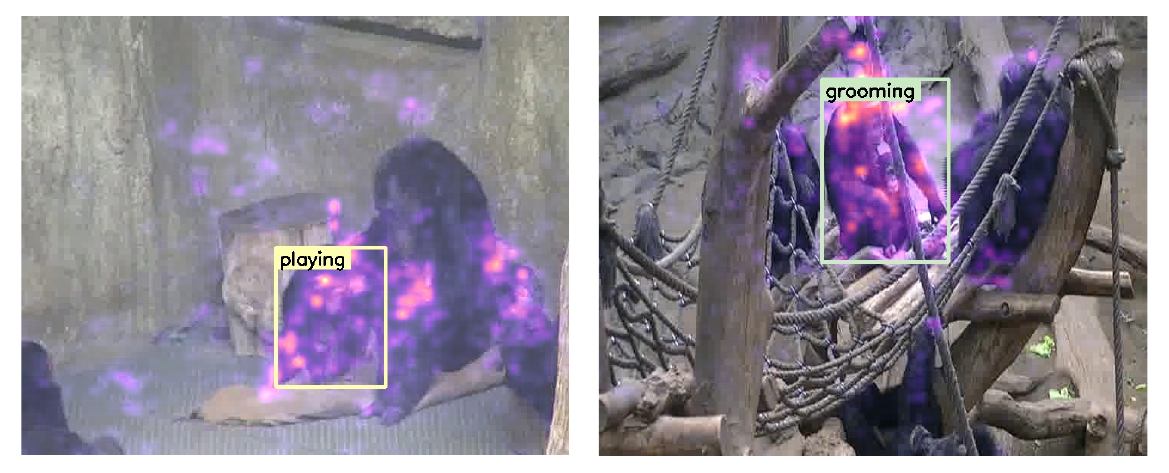}
    \caption{\textbf{Visualization of the behavior recognition probability gradient.} We visualize $\frac{\partial\action}{\partial \image}$, where $\action$ represents the behavior recognition probability for the chimpanzee boxed in image $\image$.}%
    \vspace{-1.4em}
    \label{fig:vis_gradient}%
\end{figure}

In \cref{fig:quality_result}, we present three examples of \method's tracking and behavior recognition results on the \dataset test set. The method simultaneously predicts detection boxes around chimpanzees, maintains consistent tracking (indicated by boxes of the same color across frames), and recognizes behaviors. The number above each box represents the categorization probability score. These examples highlight \method's ability to handle the significant challenges presented by \dataset, including accurate detection of occluded chimpanzees and precise behavior determination. A notable instance is in the first row, where the method successfully identifies a young chimpanzee heavily obscured by an adult. We further visualize the gradient $\frac{\partial\action}{\partial \image}$ of our behavior prediction probabilities in \cref{fig:vis_gradient}. This visualization highlights that, in predicting behaviors, our model focuses not only on the individual chimpanzee but also on other interacting entities within the scene. This broader attention is crucial for accurately recognizing social behaviors that involve multiple participants. For instance, in the first case, where a young chimpanzee is seen `playing' with an adult, our \method effectively highlights the relevant body parts of the adult chimpanzee. This indicates that our model comprehensively considers the dynamics between the individuals, allowing for a nuanced understanding of social interactions within the group. Please refer to \cref{supp:sec:methodresult} and our \projpage for more experimental results and video demos.

These results highlight \method's effectiveness in handling complex scenarios with multiple chimpanzees, occlusions, and varied behaviors. Its consistent tracking and accurate behavior recognition in challenging conditions suggest its potential for advancing automated chimpanzee behavior analysis in real-world settings.

\section{Conclusion and future work}

This work introduces \dataset, the first longitudinal video dataset capturing detailed behaviors of group-living chimpanzees, focusing on the juvenile Azibo. It offers an unprecedented view into our closest evolutionary relatives. Through extensive experiments, we reveal the challenges of using human-centric computer vision algorithms for chimpanzee behavior and introduce \method, the first end-to-end approach that detects, tracks, and recognizes chimpanzee behaviors in a unified framework. \method enhances tracking and behavior recognition by integrating temporal context and spatial relationships. Our evaluations show that \method outperforms existing methods across tasks, highlighting its capabilities and \dataset's potential to advance interdisciplinary research. By bridging primatology and computer vision, we aim to inspire specialized techniques for non-human primates, deepening our understanding of their social dynamics and contributing to animal welfare and behavioral science.

However, our current work has limitations. \dataset is based on captive chimpanzees in a semi-natural environment, limiting observable behaviors like natural foraging, predator responses, and intergroup encounters. Focusing on Azibo results in some individuals being overrepresented, restricting the assessment of the full social network. Future work could explore alternative recording methods, such as camera traps, or include a broader range of focal subjects to enhance dataset comprehensiveness. \method does not yet incorporate pose estimation, which may potentially enhance behavior recognition. It still struggles with precise behavior detection and tracking, especially under heavy occlusion or ambiguity, such as distinguishing individuals in close proximity (see failure cases in \cref{supp:fig:failure}). Future research could focus on pose estimation tailored for chimpanzees and leveraging prior knowledge to improve detection and tracking accuracy. Please check \cref{supp:sec:disall} for a more in-depth discussion.

Additionally, integrating long-term temporal analysis could capture complex social dynamics and individual development, offering insights into chimpanzee social structures and cognitive growth. Addressing these limitations and exploring these directions could significantly advance automated chimpanzee behavior analysis, enhancing our understanding of our closest evolutionary relatives and contributing to discussions on the evolution of social behavior and cognition. \\

\noindent\textbf{Acknowledgements.\quad}
\small
The authors would like to thank the Wolfgang Köhler Primate Research Center for assisting in data collection, BasicFinder CO., Ltd. and Keyue Zhang for annotations and quality check, Jiajun Su, Wentao Zhu, and Zihao Yin for discussions and preliminary experiments, Guangyuan Jiang and Yuyang Li for their technical support on the GPU cluster, and NVIDIA for their generous support of GPUs and hardware. X. Ma, Y. Lin, Y. Xu, Y. Zhu, and Y. Wang are supported in part by the National Science and Technology Major Project (2022ZD0114900). Y. Zhu is also supported in part by the National Natural Science Foundation of China (62376031), the Beijing Nova Program, the State Key Lab of General AI at Peking University, the PKU-BingJi Joint Laboratory for Artificial Intelligence, and the National Comprehensive Experimental Base for Governance of Intelligent Society, Wuhan East Lake High-Tech Development Zone.

{
\small
\bibliographystyle{IEEEtran}
\bibliography{IEEEabrv,reference_header_short,reference}
}

% \clearpage

\cleardoublepage
\renewcommand\thesection{A\arabic{section}}
\setcounter{section}{0}
\renewcommand\thefigure{A\arabic{figure}}
\setcounter{figure}{0}
\renewcommand\thetable{A\arabic{table}}
\setcounter{table}{0}
\renewcommand\theequation{A\arabic{equation}}
\setcounter{equation}{0}
\pagenumbering{arabic}% resets `page` counter to 1
\renewcommand*{\thepage}{A\arabic{page}}
\setcounter{footnote}{0}

\section{Additional details on \texorpdfstring{\dataset}{}}\label{supp:sec:dataset}

\begin{table*}[htbp]
    \center
    \small
    \caption{\textbf{The ethogram used for the \dataset dataset.}}
    \label{supp:tab:ethogram}
    \setlength{\tabcolsep}{2pt}
    \renewcommand\arraystretch{1.3}
    \resizebox{1\linewidth}{!}{%
        \begin{tabular}{l p{0.3\textwidth} l p{0.55\textwidth}}
            \toprule 
            \textbf{category} & \textbf{definition} & \textbf{subcategory} & \textbf{subcategory definition}\\
            \toprule 
            & & 0. moving & moving horizontally, \eg, walking, running\\
            \cmidrule{3-4}
            & & 1. climbing & moving vertically, \eg, climbing up or down a structure\\
            \cmidrule{3-4}
            & & 2. resting & remaining stationary, \eg, standing, sitting, or lying\\
            \cmidrule{3-4}
            \multirow{-5}{*}{locomotion} & \multirow{-5}{\hsize}{patterns of self-initiated movement of an individual} & 3. sleeping & resting and keeping eyes closed\\
            \midrule
            & & 4. solitary object playing & non-social and non-goal-directed object interaction and exploration\\
            \cmidrule{3-4}
            & & 5. eating & consuming and processing food\\
            \cmidrule{3-4}
            \multirow{-4}{*}{object interaction} & \multirow{-4}{\hsize}{direct physical interactions with inanimate stationary or movable objects by hands, feet or mouth} & 6. manipulating object & manipulation of any kind of inanimate object excluding eating\\
            \midrule
            & & 7. grooming & a chimpanzee, the groomer, is cleaning the fur, head, hand, feet, or genitals of another chimpanzee, usually using their hands and/or mouth\\
            \cmidrule{3-4}
            & & 8. being groomed & one chimpanzee, the groomee, is getting their skin or fur cleaned by another chimpanzee\\
            \cmidrule{3-4}
            & & 9. aggressing & a chimpanzee is showing agonistic behavior towards another chimpanzee. This can range from charging and chasing another chimpanzee to direct physical contact such as slapping, hitting, and biting\\
            \cmidrule{3-4}
            & & 10. embracing & a chimpanzee is embracing another chimpanzee with their arms, not to be confused with carrying\\
            \cmidrule{3-4}
            & & 11. begging & a chimpanzee is requesting food or another object from another chimpanzee, oftentimes by extending their arm, reaching, or using an open palm begging gesture\\
            \cmidrule{3-4}
            & & 12. being begged from & a chimpanzee is requested food or another object by another chimpanzee\\
            \cmidrule{3-4}
            & & 13. taking object & taking an object from the possession of another chimpanzee, the transfer might be resisted or not\\
            \cmidrule{3-4}
            & & 14. losing object & the possession is taken by another chimpanzee\\
            \cmidrule{3-4}
            & & 15. carrying & a chimpanzee (usually an adult) carries another chimpanzee (usually an infant or juvenile) on the back, front, side, arm, or leg for more than 2 steps\\
            \cmidrule{3-4}
            & & 16. being carried & a chimpanzee (usually an infant or juvenile) is carried by another chimpanzee (usually an adult) on the back, front, side, arm, or leg for more than 2 steps.\\
            \cmidrule{3-4}
            & & 17. nursing & a female chimpanzee is nursing (breastfed, \ie, making physical contact with the nipple) an infant/juvenile\\
            \cmidrule{3-4}
            & & 18. being nursed & an infant/juvenile is being nursed (breastfed, \ie, making physical contact with the nipple) by a female chimpanzee\\
            \cmidrule{3-4}
            & & 19. playing & a chimpanzee is physically interacting with another individual in a friendly, teasing, or mock fighting way (\eg, play fighting and other behaviors)\\
            \cmidrule{3-4}
            \multirow{-35}{*}{social interaction} & \multirow{-35}{\hsize}{at least two chimpanzees are interacting in differentiated roles: with one individual initiating the social behavior (initiator) and one individual receiving the social behavior (recipient)} & 20. touching & a chimpanzee makes body contact with another chimpanzee (\eg, holding hands) and it does not fit with any of the other social interaction categories described above\\
            \midrule
            & & 21. erection & a male chimpanzee has an erect penis\\
            \cmidrule{3-4}
            \multirow{-1}{*}{others} & \multirow{-1}{\hsize}{other behaviors} & 22. displaying & a male chimpanzee, usually with puffed up hair (piloerection) and an erection, performs a dominance display, which includes walking with a swagger, swinging their arms to the sides, and making calls with increasing amplitude, commonly ending by stomping against or slapping objects. Displays can be directed at another chimpanzee or be undirected\\
            \bottomrule 
        \end{tabular}%
    }%
\end{table*}

\subsection{Ethogram}\label{supp:sec:ethogram}

\begin{figure}[t!]
    \centering
    \includegraphics[width=\linewidth]{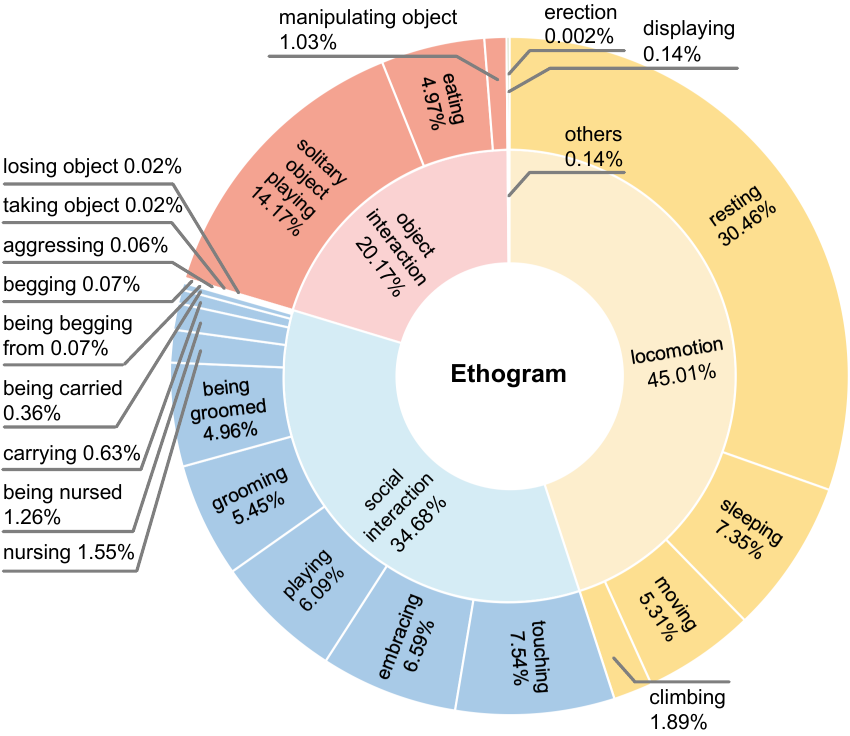} 
    \caption{\textbf{Distribution of the annotated behavior categories.} Vector graphics; zoom for details.}
    \label{supp:fig:ethogram_4_subcate}
\end{figure}

\begin{figure}[t!]
    \centering
    \includegraphics[width=\linewidth]{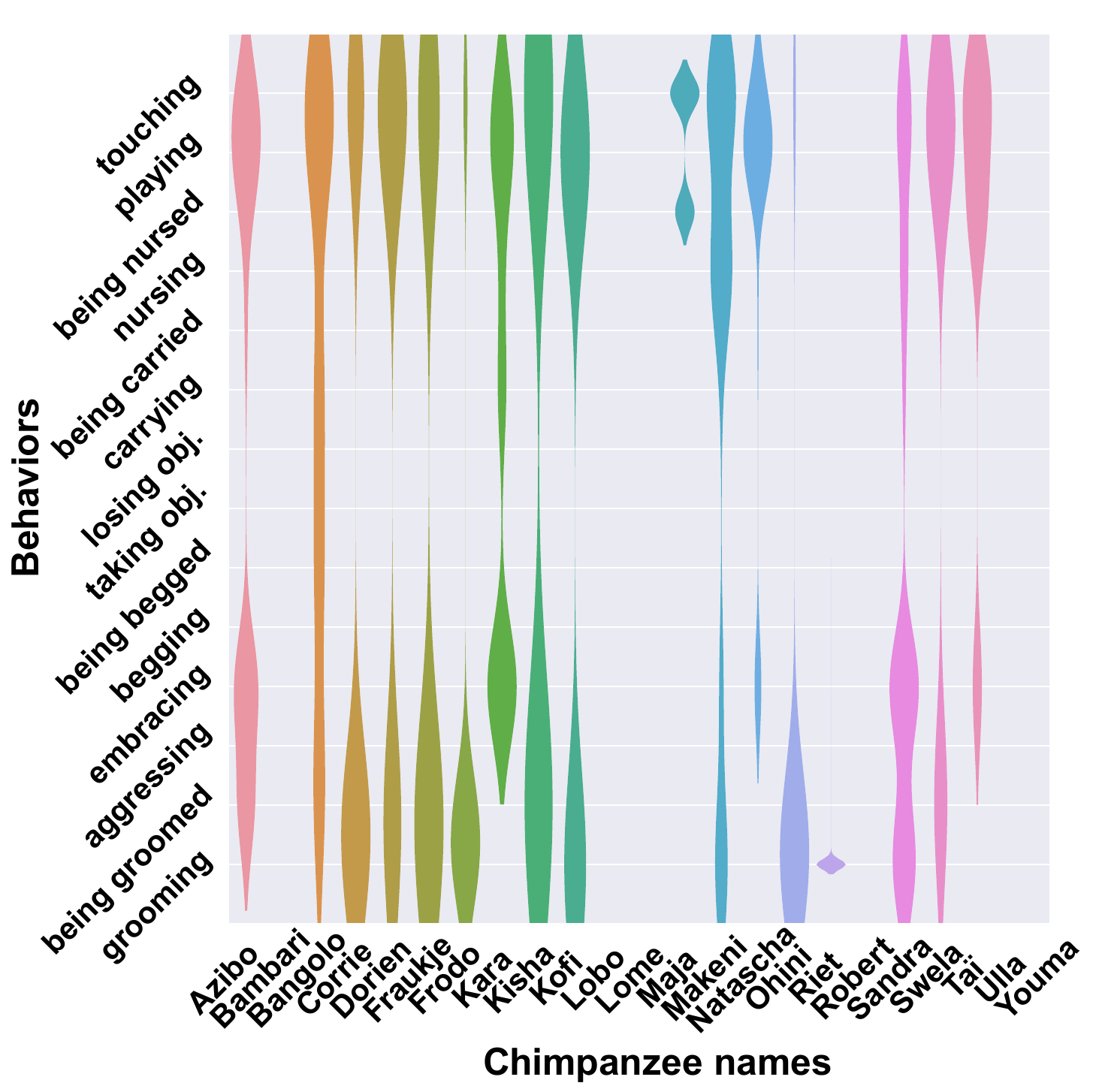} 
    \caption{\textbf{Distribution showcasing individuals alongside their respective social behaviors.} Vector graphics; zoom for details.}
    \label{supp:fig:behavior_onlysocial_identity_dist}
\end{figure}

We detail the ethogram definition in \cref{supp:tab:ethogram}, which systematically describes the daily behaviors of chimpanzees. \cref{supp:fig:ethogram_4_subcate} presents the overall distribution of annotated behaviors, with social interactions constituting approximately 35\% of total annotations. \cref{supp:fig:behavior_onlysocial_identity_dist} shows the distribution of social behaviors across individuals, highlighting grooming, playing, and touching as predominant activities within the group's social dynamics.

\begin{figure*}[htbp]
    \centering
    \includegraphics[width=0.75\linewidth]{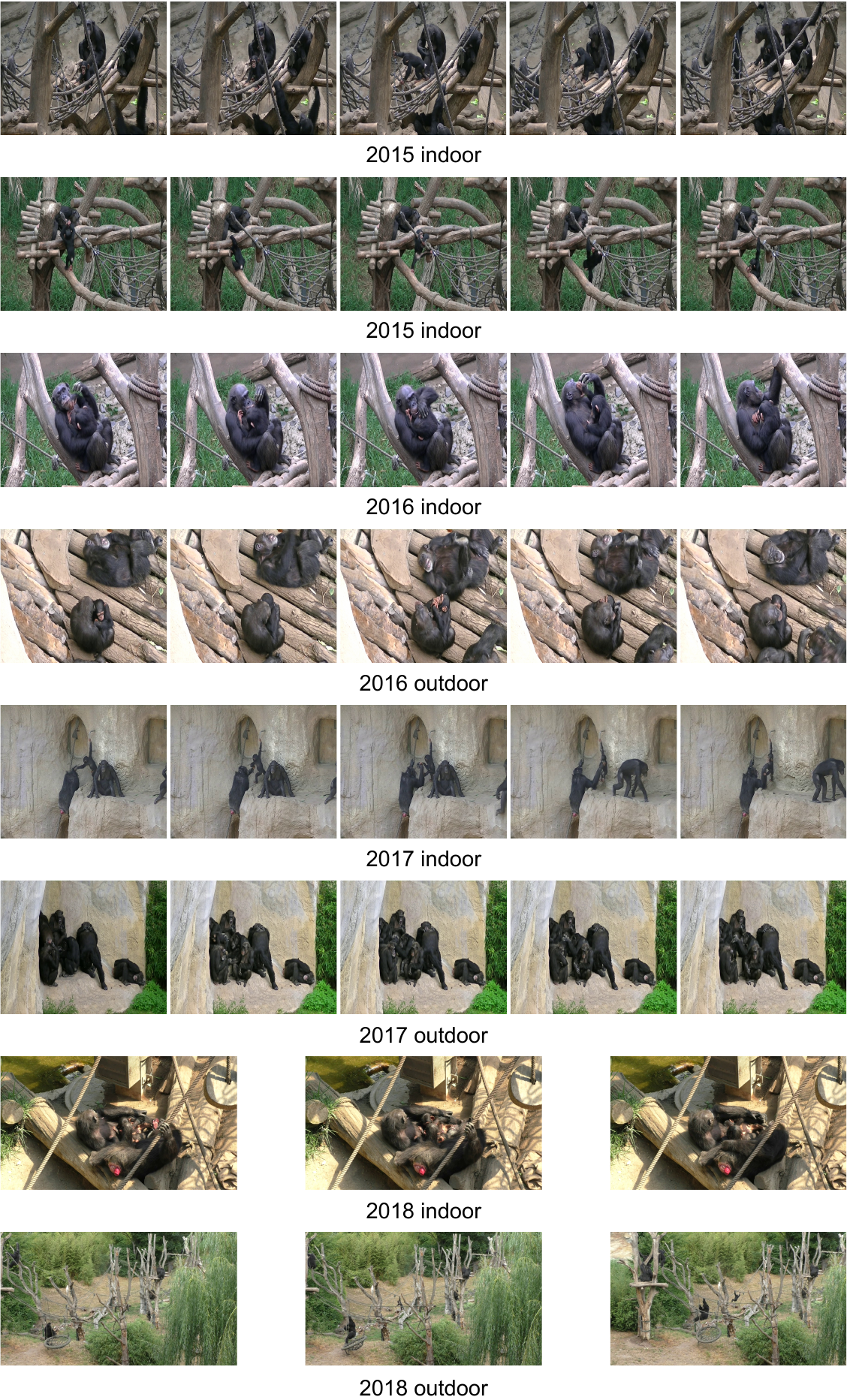} 
    \caption{\textbf{Example frames from the \dataset dataset.} \dataset possesses rich social interactions of the complex everyday life of group-living chimpanzees and contains several environmental enrichment.}
    \label{supp:fig:supp_dataset}
\end{figure*}

\subsection{Dataset details}\label{supp:sec:datadetail}

\noindent\textbf{Collection and organization.\quad}
405 hours of video footage of the Leipzig A-group chimpanzees were collected between 2015 and 2018. To create a representative sample of the footage, 163 video clips were selected, with 15, 35, 86, and 27 clips taken from each year. These video clips cover the four seasons. Each clip is 1000 frames long, with only 3 clips being shorter than 1000 frames. Visual examples from six clips, featuring both indoor and outdoor enclosures, are shown in \cref{supp:fig:supp_dataset}. The dataset covers a diverse range of physical scenarios, camera views, and social behaviors, as demonstrated in these examples. For instance, in the third row of the figure, an adult chimpanzee is shown grooming an infant chimpanzee in her arms, while later on, the same infant is nursed.  \\

\noindent\textbf{Annotation process and quality.\quad}
The annotation process was conducted using BasicFinder CO., Ltd.'s private labeling platform, which involved a team of 15 annotators and 2 managers. Prior to commencing the annotation work, our team developed comprehensive guidelines that explicitly outlined the requirements for labeling. These guidelines covered several aspects, including:

(i) Assigning a bounding box for each chimpanzee in the image.
(ii) Specifying the visibility of the bounding boxes.
(iii) Assigning tracking IDs to each bounding box for tracking purposes.
(iv) Localizing 2D keypoints within each bounding box.
(v) Indicating the visibility of each 2D keypoint.
(vi) Assigning behavior labels for each bounding box.

To ensure that the annotators followed these guidelines accurately, the project managers provided training based on the guidelines. Following the training, the annotators performed a trial annotation on a small dataset. We actively sought feedback from the annotators during this phase, which allowed us to address any issues and make necessary improvements. We conducted a thorough review of the trial annotations to verify that the quality met our standards.

During the trial labeling phase, we reached out to three labeling companies and ultimately selected BasicFinder CO., Ltd. based on their exceptional labeling quality. It is worth noting that BasicFinder CO., Ltd. has previously led the annotation efforts for the BDD100K \cite{yu2020bdd100k} dataset, which is a substantial dataset used for autonomous driving purposes. This experience demonstrates their ability to maintain high annotation standards for complex and extensive datasets. Consequently, their involvement improves the reliability of our \dataset dataset annotations as well.

Once we were confident in the quality of the trial annotations, we proceeded with the large-scale annotation process. To manage the annotations efficiently, each video clip was designated as an annotation task, and our managers assigned these tasks to individual annotators using BasicFinder CO., Ltd.'s platform, ensuring that there was no overlap in assignments. BasicFinder CO., Ltd. has implemented rigorous quality management practices throughout the annotation process. These practices include a customized workflow, complete job traceability, precise performance tracking, multiple levels of auditing, and scientific personnel management. By adhering to these practices, we were able to maintain high standards of quality and accuracy while ensuring efficient processing speed. The annotation process followed a sequential workflow of execution, review, and quality control. Experienced annotators were responsible for executing the annotations, while the manager, as well as our team, conducted thorough reviews and quality control checks. Any annotations that did not meet the required standards were sent back to the annotators for corrections. The quality control phase involved a comprehensive review and verification of all data by both the managers and our own team, ensuring the integrity and accuracy of the annotations. Once all the data had been confirmed to meet our standards of quality, we concluded the annotation process.

More specifically, to label chimpanzee identities, annotators only needed to assign a tracking ID to each chimpanzee, which was then reviewed by the primatologist in our team, who assigned the apes' names based on his knowledge of the observed Leipzig A-group chimpanzees. The process of localizing 2D keypoints within each bounding box and assigning behavior labels for each chimpanzee presented bigger challenges than other tasks. To overcome these challenges, we implemented several measures to ensure accuracy and consistency. For the labeling of 2D keypoints, we provided detailed instructions accompanied by visual illustrations, aiming to provide clear guidelines for annotators to precisely identify and mark the keypoints. 
For labeling of behaviors, we supplied example videos showcasing different chimpanzee behaviors, created by our team's experienced primatologists. These videos served as valuable references, enabling annotators to accurately assign behavior labels based on observed actions. Throughout the annotation process, the primatologists actively participated, offering their expertise and providing valuable feedback to ensure the annotations aligned with scientific standards. Finally, the behavioral primatologists in our team manually reviewed all labeled frames to ensure data reliability. These measures and the involvement of the primatologists were instrumental in enhancing the overall quality and reliability of the annotations. 

For more information on the dataset, including pre-processing scripts, and visualized annotations, please refer to our \dataprojpage.

\section{Discussion on \texorpdfstring{\dataset}{}}\label{supp:sec:dis}

\begin{figure*}[t!]
    \centering
    \hfill%
    \begin{subfigure}[]{0.333\linewidth}
        \centering
        \includegraphics[width=\linewidth]{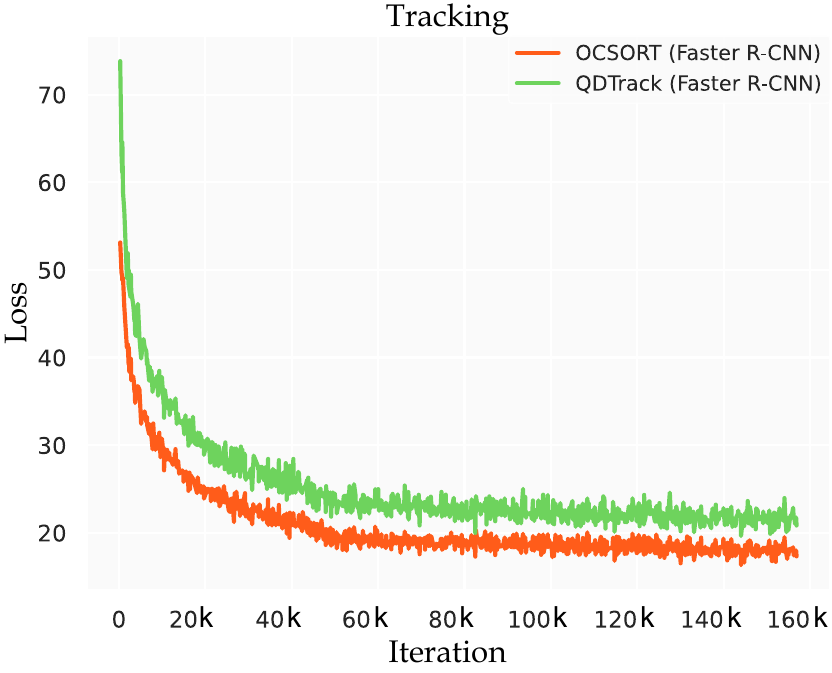}
        \caption{}
        \label{supp:fig:supp_val_tracking}
    \end{subfigure}%
    \hfill%
    \begin{subfigure}[]{0.333\linewidth}
        \centering
        \includegraphics[width=\linewidth]{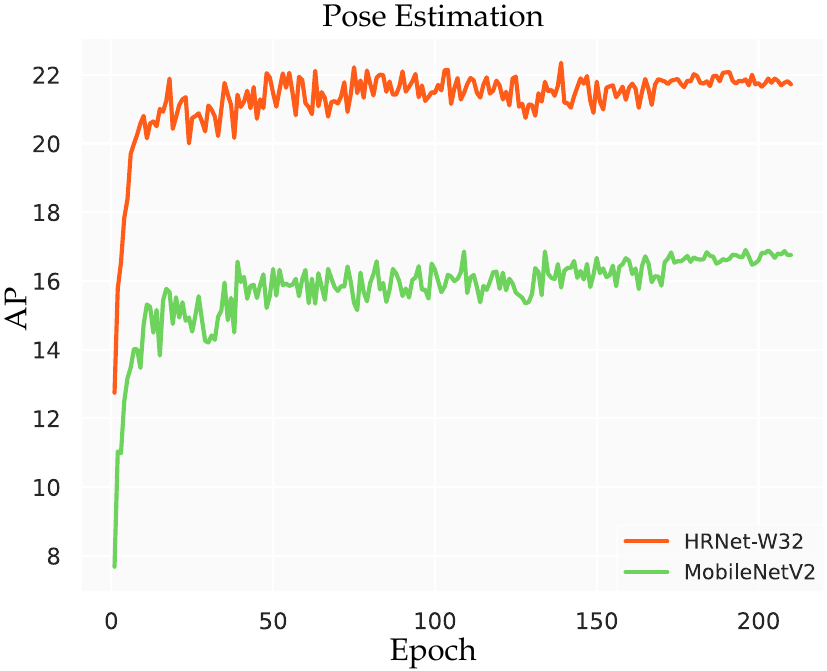} 
        \caption{}
        \label{supp:fig:supp_val_pose}
    \end{subfigure}%
    \hfill%
    \begin{subfigure}[]{0.325\linewidth}
        \centering
        \includegraphics[width=\linewidth]{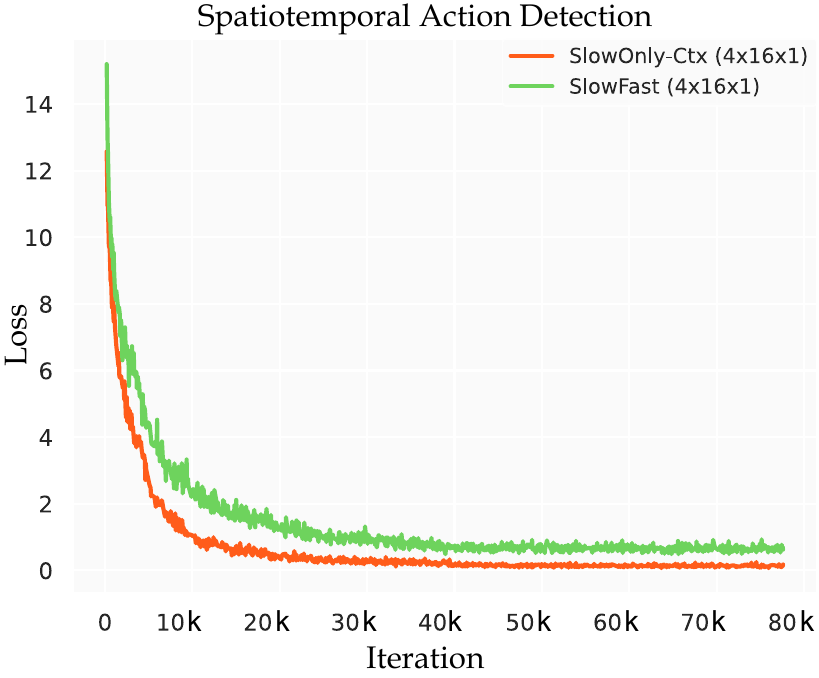} 
        \caption{}
        \label{supp:fig:supp_val_action}
    \end{subfigure}%
    \hfill%
    \caption{\textbf{Training or validation curves on three tracks of example methods.} (a) Training loss curve of example tracking methods. The training iterations correspond to 10 epochs. (b) Validation curve on the AP metric of example pose estimation methods. (c) Training loss curve of example spatiotemporal action detection methods. The training iterations correspond to 20 epochs. }
\end{figure*}

\noindent\textbf{Intended uses.\quad}
The \dataset dataset is a versatile resource that can be used for studying algorithms for chimpanzee detection, tracking, identification, pose estimation, and spatiotemporal action detection. Therefore, the dataset is both relevant for questions in computer vision and primate behavior. In the context of computer vision, it lends itself to other research topics, including but not limited to pose tracking, few-shot learning, weakly-supervised learning, and transfer learning. Considering primate behavior, the dataset shares numerous features with other video data commonly collected with captive and wild chimpanzee populations. This makes it an ideal resource for fine-grained investigations of social (\eg, grooming, nursing, aggression) and nonsocial (\eg, locomotion, object interactions) chimpanzee behaviors. We strongly encourage researchers to utilize our dataset solely for research purposes that promote animal welfare and conservation. We firmly discourage any use of the dataset for harmful activities such as poaching, hunting or any other exploitation of primates. It is crucial for researchers to approach the data with a focus on positive societal impacts and to refrain from any potential negative consequences. \\

\noindent\textbf{Ethics.\quad}
The \dataset dataset raises no ethical concerns regarding the privacy information of human subjects, as it solely focuses on chimpanzees. Studying the social behavior of chimpanzees provides an ethical and efficient means to explore aspects of human sociality due to our phylogenetic proximity. By analyzing their behaviors, we can gain insights into the evolution of human social behavior and potentially contribute to both the scientific and ethical understanding of the human condition. The ethics committee of the Wolfgang Köhler Primate Research Center approved the observational data collection for this project. \\

\noindent\textbf{Maintenance, distribution, and license.\quad}
The \dataset dataset will be maintained by the authors and made publicly available with a total of 160,500 frames (around 2 hours) on our \dataprojpage. The \dataset dataset will be distributed under the CC BY-NC 4.0 license.  \\
% More labeled videos will be released in the future.

\noindent\textbf{Wage paid to annotators.\quad}
We collaborated with BasicFinder CO., Ltd. for the annotation process. The labeling was carried out by 15 annotators, and they were offered a fair wage as per the prearranged contract. The total expenditure for the labeling process was approximately 70,000 RMB.

\section{Benchmarking \textbf{\dataset}}\label{supp:sec:experiment}

We trained all the models with officially-used training configurations for each of the three tracks. Please refer to the code implementation on our \href{https://github.com/ShirleyMaxx/ChimpACT}{Github} for details. 
Although we trained the models for different epochs in experiments conducted on different tracks, these choices were made based on conventional practices. Based on the training loss curves provided in \cref{supp:fig:supp_val_tracking,supp:fig:supp_val_action}, it can be observed that all tracking and spatiotemporal action detection methods have reached convergence within the chosen training epochs. To assess the potential overfitting of the pose estimation models, we have included the validation curve on the AP metric in \cref{supp:fig:supp_val_pose}. The validation curve demonstrates the performance of the pose models on the validation set, which indicates that the pose estimation models are not exhibiting signs of overfitting. Therefore, based on the training loss curves and the validation curve, it can be concluded that the chosen training epochs are appropriate for both tracking and pose estimation methods.

\subsection{Detection, tracking, and \texorpdfstring{\ac{reid}}{}}\label{supp:sec:tracking}

We partitioned the dataset of 163 videos into three sets: 127 videos for training, 17 for validation, and 19 for testing. Of note, all individual chimpanzees are present in both the training and testing sets. In the test set, there are 12 and 7 videos for indoor and outdoor scenes, respectively. 

For the evaluation metrics, MOTA (Multiple Object Tracking Accuracy) takes into account FP (False Positives), FN (False Negatives), and IDs (IDentity switches). Usually, FP and FN are larger than IDs; therefore, MOTA mainly assesses the detection performance. IDF1 evaluates the ability to preserve subject identities to assess identification association performance. HOTA (Higher Order Tracking Accuracy) is a recently proposed metric that considers accurate detection, association, and localization equally important, and balances their effects explicitly.  \\

\begin{figure*}[htbp]
    \centering
    \includegraphics[width=0.8\linewidth]{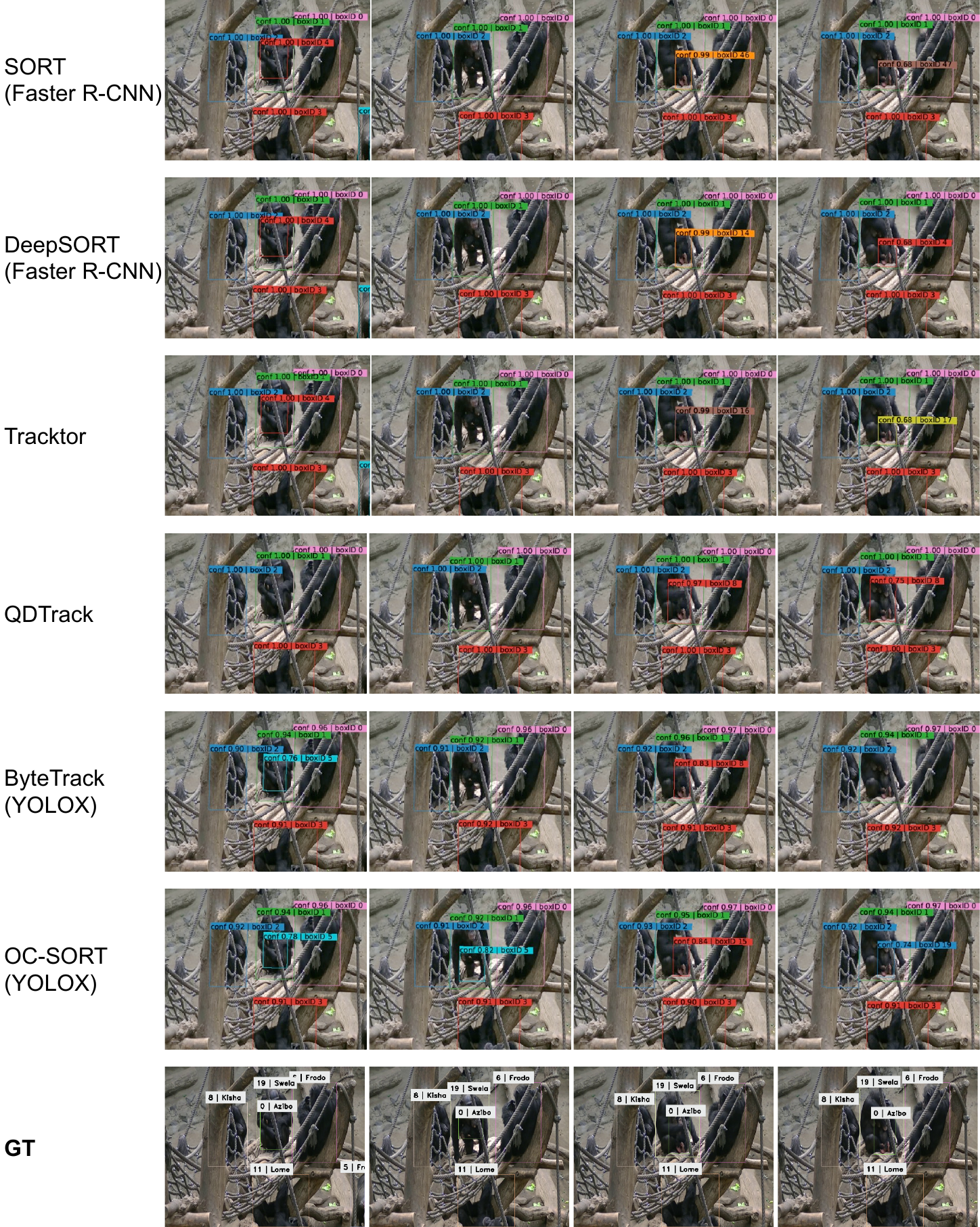} 
    \caption{\textbf{Qualitative results of representative methods on the \dataset test set on the tracking task.} For each method, we visualize the estimated confidence score (``conf'') and the associated IDs (``boxID'') of each bounding box in each frame. The ground-truth bounding boxes and chimpanzee names are shown in the last row, and we add a number left to the name to make it easier to track. Please zoom in for details.}
    \label{supp:fig:supp_vis_pred_box1}
\end{figure*}

\begin{table*}[t!]
    \center
    \small
    \caption{\textbf{Results of the pose estimation track for each keypoint on \dataset test set.} We report PCK@0.1 metric. We abbreviate the keypoint names. 
    % Please refer to \cref{tab:keypointdef} for the keypoint definition.
    Please refer to the main paper for the keypoint definition.
    }
    \label{supp:tab:supp-posepck01}
    \setlength{\tabcolsep}{2pt}
    \resizebox{\linewidth}{!}{%
        \begin{tabular}{l l l c c c c c c c c c c c c c c c c}
            \toprule 
            & Method & Backbone & 0.hip & 1.rknee & 2.rankle & 3.lknee & 4.lankle & 5.neck & 6.ulip & 7.llip & 8.reye & 9.leye & 10.rshoul & 11.relbow & 12.rwrist & 13.lshoul & 14.lelbow & 15.lwrist\\
            \toprule 
            & \multirow{3}{*}{SimpleBaseline \cite{xiao2018simple}} & ResNet-50 & 51.1 & 45.8 & 52.3 & 44.7 & 48.8 & 56.4 & 76.2 & 77.9 & 85.7 & 85.2 & 54.7 & 46.1 & 29.2 & 60.5 & 48.5 & 31.5  \\

            & & ResNet-101  & 51.3 & 49.0 & 53.3 & 47.3 & 50.2 & 58.2 & 77.1 & 78.7 & 86.4 & 86.4 & 57.9 & 46.8 & 32.5 & 60.2 & 51.8 & 35.4  \\

            & & ResNet-152  & 50.6 & 50.5 & 56.8 & 47.4 & 45.3 & 58.3 & 76.4 & 77.4 & 86.8 & 86.0 & 55.8 & 45.1 & 35.2 & 58.2 & 51.3 & 35.8  \\

            \cmidrule{2-19}
            & \multirow{4}{*}{RLE \cite{li2021human}} & MobileNetV2  & 53.1 & 46.9 & 53.8 & 49.0 & 48.7 & 61.4 & 77.1 & 78.7 & 86.3 & 85.1 & 59.2 & 41.6 & 33.5 & 59.0 & 48.2 & 31.9  \\

            & & ResNet-50  & 47.7 & 42.6 & 46.6 & 42.7 & 46.2 & 57.7 & 75.9 & 77.4 & 81.4 & 79.3 & 59.0 & 44.0 & 30.3 & 58.9 & 48.5 & 30.5  \\

            & & ResNet-101  & 51.9 & 49.4 & 52.8 & 55.4 & 49.1 & 61.0 & 79.5 & 80.4 & 87.2 & 86.6 & 60.3 & 46.4 & 40.2 & 62.0 & 53.6 & 39.0  \\

            \multirow{-7}{*}{\rotatebox{90}{\textit{Regression}}} & & ResNet-152  & 54.1 & 50.2 & 52.8 & 53.1 & 49.3 & 60.6 & 79.8 & 80.7 & 88.0 & 85.4 & 63.1 & 50.7 & 42.5 & 61.1 & 53.5 & 38.6  \\

            \midrule 
            & CPM \cite{wei2016convolutional} & CPM  & 61.1 & 65.9 & 71.7 & 59.7 & 68.7 & 67.3 & 85.5 & 87.2 & 91.1 & 90.5 & 67.0 & 60.1 & 59.6 & 67.8 & 66.3 & 53.4  \\

            & Hourglass \cite{newell2016stacked} & Hourglass-4  & 62.4 & 65.3 & 70.8 & 65.2 & 67.8 & 66.4 & 84.0 & 85.9 & 86.9 & 87.3 & 68.5 & 61.7 & 60.4 & 67.7 & 66.0 & 56.0  \\

            & MobileNetV2 \cite{sandler2018mobilenetv2} & MobileNetV2  & 58.8 & 64.8 & 71.2 & 61.3 & 64.8 & 67.0 & 83.8 & 85.4 & 91.0 & 89.1 & 69.3 & 58.6 & 56.9 & 67.4 & 64.8 & 52.1  \\

            \cmidrule{2-19}
            & \multirow{3}{*}{SimpleBaseline \cite{xiao2018simple}} & ResNet-50  & 63.2 & 67.9 & 70.7 & 64.4 & 67.5 & 66.8 & 85.1 & 86.3 & 92.8 & 90.5 & 70.6 & 59.1 & 57.7 & 67.6 & 65.0 & 54.4  \\

            & & ResNet-101 & 62.0 & 64.6 & 69.6 & 61.4 & 68.4 & 67.4 & 85.1 & 87.4 & 91.9 & 89.6 & 70.1 & 61.2 & 56.3 & 66.7 & 63.5 & 54.1  \\

            & & ResNet-152  & 64.5 & 64.6 & 69.2 & 62.5 & 69.9 & 67.3 & 86.5 & 88.5 & 91.1 & 89.7 & 72.4 & 62.4 & 58.5 & 69.9 & 66.0 & 55.3  \\

            \cmidrule{2-19}
            & \multirow{2}{*}{HRNet \cite{sun2019deep}} & HRNet-W32  & 65.8 & 69.5 & 74.5 & 66.1 & 69.2 & 70.5 & 88.2 & 90.4 & 92.6 & 92.1 & 76.1 & 67.7 & 64.4 & 72.4 & 69.5 & 62.8  \\

            & & HRNet-W48  & 61.5 & 69.1 & 74.6 & 65.1 & 70.4 & 70.7 & 87.5 & 88.9 & 93.8 & 92.2 & 75.3 & 64.7 & 61.1 & 72.1 & 70.6 & 58.9  \\

            \cmidrule{2-19}
            & \multirow{5}{*}{DarkPose \cite{zhang2020distribution}} & ResNet-50  & 62.6 & 64.4 & 68.6 & 63.2 & 66.6 & 69.9 & 86.3 & 87.7 & 91.7 & 90.5 & 73.8 & 61.5 & 59.1 & 69.6 & 66.9 & 58.0  \\

            & & ResNet-101 & 61.7 & 62.9 & 70.5 & 62.6 & 65.7 & 67.0 & 86.3 & 87.7 & 92.0 & 89.7 & 70.1 & 59.7 & 55.4 & 68.6 & 62.8 & 54.4  \\

            & & ResNet-152 & 63.3 & 68.6 & 69.1 & 62.5 & 66.0 & 67.7 & 86.5 & 88.0 & 92.6 & 89.5 & 71.8 & 61.9 & 56.1 & 69.5 & 63.3 & 53.4  \\

            & & HRNet-W32 & 63.5 & 67.3 & 74.0 & 67.2 & 71.6 & 70.0 & 88.3 & 89.5 & 93.4 & 92.1 & 75.6 & 65.3 & 64.3 & 73.1 & 69.2 & 62.6  \\

            & & HRNet-W48 & 65.9 & 69.7 & 73.5 & 67.1 & 72.8 & 72.0 & 89.6 & 91.3 & 94.5 & 91.8 & 73.3 & 62.6 & 61.2 & 71.6 & 70.8 & 62.8  \\

            \cmidrule{2-19}
            & \multirow{2}{*}{HRFormer \cite{yuan2021hrformer}} & HRFormer-S & 63.0 & 66.5 & 70.7 & 64.2 & 68.5 & 67.5 & 84.5 & 85.6 & 91.0 & 89.1 & 71.3 & 61.0 & 59.1 & 68.3 & 64.9 & 56.0  \\

            \multirow{-15}{*}{\rotatebox{90}{\textit{Heatmap-based}}} & & HRFormer-B & 61.4 & 67.2 & 71.9 & 66.3 & 70.9 & 67.7 & 84.9 & 86.2 & 93.6 & 90.6 & 71.9 & 66.3 & 62.3 & 70.8 & 67.2 & 58.0  \\

            \bottomrule 
        \end{tabular}%
    }%
\end{table*}

\begin{table*}[ht!]
    \center
    \small
    \caption{\textbf{Results of the pose estimation track for each action category on \dataset test set.} We report PCK@0.1 metric. The action category number is consistent with \cref{supp:tab:ethogram}.}
    \label{supp:tab:supp-posepck01-action}
    \setlength{\tabcolsep}{2pt}
    \resizebox{\linewidth}{!}{%
        \begin{tabular}{l l l c c c c c c c c c c c c c c c c c c c c c}
            \toprule 
            & Method & Backbone & 0 & 1 & 2 & 3 & 4 & 5 & 6 & 7 & 8 & 9 & 10 & 11 & 12 & 15 & 16 & 17 & 18 & 19 & 20 & 21 & 22\\
            \toprule 
            & \multirow{3}{*}{SimpleBaseline \cite{xiao2018simple}} & ResNet-50 & 39.8 & 47.7 & 48.0 & 56.1 & 44.4 & 64.2 & 56.7 & 35.7 & 26.1 & 81.3 & 67.0 & 51.3 & 45.0 & 26.5 & 34.5 & 3.7 & 5.5 & 29.9 & 26.9 & 87.5 & 56.6\\

            & & ResNet-101  & 39.3 & 49.0 & 48.1 & 59.5 & 46.3 & 60.9 & 57.5 & 38.9 & 20.7 & 75.0 & 69.5 & 57.5 & 45.0 & 32.1 & 35.5 & 2.9 & 6.6 & 28.8 & 26.6 & 62.5 & 52.5\\

            & & ResNet-152  & 40.8 & 45.6 & 49.9 & 56.2 & 46.5 & 63.9 & 54.8 & 37.7 & 23.6 & 68.8 & 67.4 & 62.5 & 51.3 & 30.6 & 34.3 & 6.6 & 5.3 & 29.5 & 26.1 & 75.0 & 56.6\\

            \cmidrule{2-24}
            & \multirow{4}{*}{RLE \cite{li2021human}} & MobileNetV2  & 40.8 & 48.0 & 50.0 & 52.9 & 47.1 & 63.5 & 57.7 & 38.4 & 18.1 & 62.5 & 67.6 & 53.8 & 48.8 & 28.8 & 36.5 & 5.7 & 8.5 & 29.4 & 29.8 & 62.5 & 62.5\\

            & & ResNet-50  & 42.0 & 52.1 & 51.5 & 57.6 & 50.0 & 65.2 & 57.8 & 41.0 & 20.2 & 75.0 & 69.0 & 50.0 & 55.0 & 31.9 & 35.4 & 4.6 & 3.2 & 29.0 & 31.5 & 81.3 & 61.3\\

            & & ResNet-101  & 43.3 & 46.4 & 51.8 & 58.3 & 46.6 & 68.0 & 55.8 & 34.1 & 18.7 & 75.0 & 68.5 & 48.8 & 43.8 & 31.9 & 35.2 & 6.0 & 5.6 & 29.7 & 31.6 & 75.0 & 62.6\\

            \multirow{-7}{*}{\rotatebox{90}{\textit{Regression}}} & & ResNet-152  & 41.4 & 52.9 & 50.7 & 57.2 & 49.3 & 64.2 & 56.3 & 34.1 & 17.7 & 75.0 & 72.5 & 48.8 & 46.3 & 35.6 & 31.8 & 5.4 & 6.1 & 30.2 & 30.1 & 75.0 & 59.8\\

            \midrule 
            & CPM \cite{wei2016convolutional} & CPM  & 49.4 & 59.4 & 59.0 & 60.9 & 53.9 & 73.1 & 65.2 & 46.6 & 28.4 & 81.3 & 66.6 & 52.5 & 60.0 & 41.6 & 34.8 & 10.6 & 3.8 & 36.1 & 41.8 & 68.8 & 66.2\\

            & Hourglass \cite{newell2016stacked} & Hourglass-4  & 48.1 & 66.5 & 55.3 & 63.2 & 58.5 & 71.9 & 67.4 & 50.3 & 27.8 & 81.3 & 72.8 & 47.5 & 65.0 & 44.0 & 38.2 & 14.1 & 1.8 & 40.6 & 35.3 & 81.3 & 63.6\\

            & MobileNetV2 \cite{sandler2018mobilenetv2} & MobileNetV2  & 49.8 & 58.4 & 56.1 & 59.3 & 54.8 & 71.2 & 65.1 & 52.8 & 25.8 & 75.0 & 72.8 & 60.0 & 48.8 & 39.8 & 35.8 & 11.7 & 1.5 & 35.0 & 36.2 & 81.3 & 61.4\\

            \cmidrule{2-24}
            & \multirow{3}{*}{SimpleBaseline \cite{xiao2018simple}} & ResNet-50  & 52.3 & 60.0 & 57.2 & 60.9 & 56.3 & 73.9 & 66.0 & 53.3 & 25.2 & 81.3 & 72.0 & 62.5 & 65.0 & 45.0 & 35.4 & 8.4 & 2.4 & 39.1 & 37.6 & 75.0 & 65.7\\

            & & ResNet-101 & 52.0 & 60.9 & 57.5 & 60.8 & 56.6 & 71.9 & 66.4 & 52.6 & 28.2 & 93.8 & 72.2 & 71.3 & 61.3 & 42.0 & 34.2 & 6.4 & 1.9 & 39.6 & 36.9 & 68.8 & 67.0\\

            & & ResNet-152  & 51.4 & 60.0 & 57.8 & 60.0 & 57.4 & 71.6 & 66.3 & 55.4 & 27.8 & 81.3 & 79.1 & 58.8 & 52.5 & 45.1 & 32.3 & 5.3 & 0.5 & 38.1 & 38.0 & 87.5 & 67.6\\

            \cmidrule{2-24}
            & \multirow{2}{*}{HRNet \cite{sun2019deep}} & HRNet-W32  & 56.7 & 66.1 & 60.8 & 60.9 & 60.2 & 76.3 & 69.3 & 54.8 & 27.2 & 87.5 & 74.8 & 61.3 & 63.8 & 50.6 & 38.7 & 9.1 & 2.4 & 40.2 & 41.4 & 81.3 & 65.6\\

            & & HRNet-W48  & 56.9 & 65.9 & 59.3 & 60.9 & 60.3 & 75.7 & 70.3 & 53.2 & 30.2 & 87.5 & 74.2 & 66.3 & 67.5 & 52.9 & 37.6 & 13.9 & 2.9 & 41.3 & 39.7 & 87.5 & 65.9\\

            \cmidrule{2-24}
            & \multirow{5}{*}{DarkPose \cite{zhang2020distribution}} & ResNet-50  & 52.1 & 60.9 & 57.5 & 62.1 & 57.4 & 72.6 & 66.1 & 56.0 & 26.8 & 75.0 & 72.9 & 56.3 & 61.3 & 42.1 & 31.7 & 9.6 & 0.4 & 35.3 & 39.0 & 75.0 & 66.4\\

            & & ResNet-101 & 52.6 & 62.6 & 57.6 & 61.4 & 56.1 & 71.8 & 67.7 & 51.7 & 26.3 & 81.3 & 73.4 & 65.0 & 62.5 & 44.0 & 38.2 & 6.3 & 2.6 & 36.7 & 35.7 & 87.5 & 61.1\\

            & & ResNet-152 & 52.6 & 63.3 & 57.8 & 59.4 & 57.9 & 73.2 & 67.9 & 53.0 & 25.7 & 81.3 & 76.3 & 57.5 & 65.0 & 45.0 & 35.0 & 8.7 & 1.7 & 35.9 & 37.1 & 87.5 & 68.3\\

            & & HRNet-W32 & 56.9 & 68.9 & 62.6 & 62.5 & 61.5 & 74.0 & 69.7 & 56.5 & 26.0 & 81.3 & 81.2 & 58.8 & 72.5 & 52.2 & 42.3 & 9.8 & 2.1 & 41.6 & 44.5 & 81.3 & 67.3\\

            & & HRNet-W48 & 57.6 & 67.7 & 60.3 & 59.2 & 60.3 & 73.7 & 69.6 & 56.3 & 28.5 & 93.8 & 77.2 & 53.8 & 67.5 & 52.8 & 36.7 & 4.2 & 1.4 & 39.7 & 40.4 & 75.0 & 63.9\\

            \cmidrule{2-24}
            & \multirow{2}{*}{HRFormer \cite{yuan2021hrformer}} & HRFormer-S & 52.9 & 62.3 & 55.7 & 59.6 & 56.8 & 72.3 & 68.2 & 54.2 & 23.7 & 93.8 & 75.1 & 68.8 & 52.5 & 45.1 & 33.5 & 2.5 & 1.1 & 40.6 & 34.5 & 62.5 & 66.9\\

            \multirow{-15}{*}{\rotatebox{90}{\textit{Heatmap-based}}} & & HRFormer-B & 54.2 & 63.4 & 58.0 & 61.3 & 58.8 & 72.4 & 68.2 & 52.4 & 25.4 & 81.3 & 77.5 & 55.0 & 67.5 & 46.8 & 37.5 & 12.6 & 0.7 & 40.8 & 37.7 & 75.0 & 63.7\\

            \bottomrule 
        \end{tabular}%
    }%
\end{table*}

\begin{figure*}[t!] 
    \begin{minipage}{0.6\linewidth}
        \centering 
        \small
        \captionof{table}{\textbf{Results of the pose estimation by HRNet-W32 model.} We report PCK@0.05 and PCK@0.1 metrics.} 
        \label{supp:tab:supp_pose_action_type}
        \begin{tabular}{cc c c}
            \toprule
            No. & Action & PCK@0.05 & PCK@0.1\\
            \midrule
            (a)     & resting & 43.8 & 62.5\\
            (b)     & climbing & 81.2 & 93.8\\
            (c)     & resting & 68.8 & 93.8\\
            (d)     & climbing & 75.0 & 100.0\\
            \bottomrule
        \end{tabular}%
    \end{minipage}% 
    \begin{minipage}{0.4\linewidth} 
        \centering
        \includegraphics[width=0.7\linewidth]{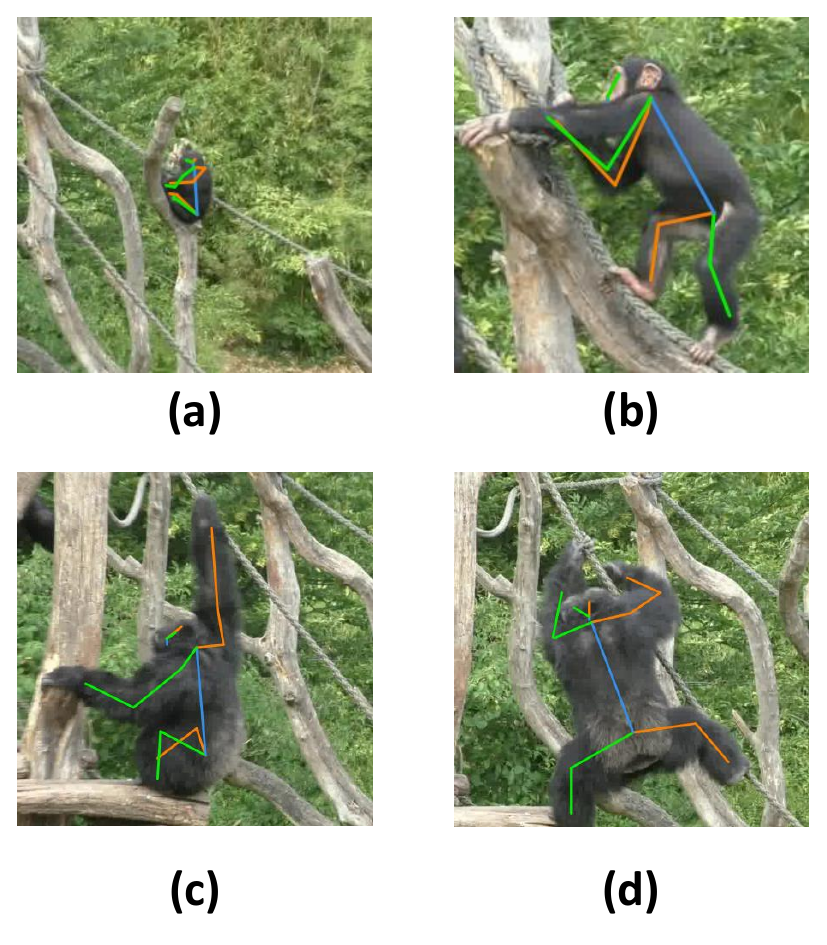}
        \captionof{figure}{\textbf{Visualization of predicted pose by HRNet-W32 \cite{sun2019deep}}.}
        \label{supp:fig:supp_pose_action_type} 
    \end{minipage}
\end{figure*}

\begin{table}[ht!]
    \centering
    \small
    \caption{\textbf{Results of the pose estimation track for non-occluded poses on \dataset test set.} We report the PCK metrics. The non-occluded poses denote those with all keypoints visible.}
    \label{supp:tab:supp-posepck01-visible}
    \setlength{\tabcolsep}{0.2pt}
        \begin{tabular}{l l l c c }
            \toprule 
            & Method & Backbone & PCK@0.05 & PCK@0.1\\
            \toprule 
            & \multirow{3}{*}{SimpleBaseline \cite{xiao2018simple}} & ResNet-50 & 47.6 & 80.6\\

            & & ResNet-101  & 47.2 & 77.9\\

            & & ResNet-152  & 54.5 & 83.0\\
            
            \cmidrule{2-5}
            & \multirow{4}{*}{RLE \cite{li2021human}} & MobileNetV2  & 47.7 & 82.4\\

            & & ResNet-50  & 52.9 & 82.4\\

            & & ResNet-101  & 28.4 & 55.1\\

            \multirow{-7}{*}{\rotatebox{90}{\textit{Regression}}} & & ResNet-152  & 60.0 & 85.5\\

            \midrule 
            & CPM \cite{wei2016convolutional} & CPM  & 74.0 & 89.4\\

            & Hourglass \cite{newell2016stacked} & Hourglass-4  & 77.6 & 88.5\\

            & MobileNetV2 \cite{sandler2018mobilenetv2} & MobileNetV2  & 67.4 & 89.0\\

            \cmidrule{2-5}
            & \multirow{3}{*}{SimpleBaseline \cite{xiao2018simple}} & ResNet-50  & 75.2 & 89.5\\

            & & ResNet-101 & 68.7 & 84.0\\

            & & ResNet-152  & 71.4 & 87.1\\

            \cmidrule{2-5}
            & \multirow{2}{*}{HRNet \cite{sun2019deep}} & HRNet-W32  & 77.6 & 92.1\\

            & & HRNet-W48  & 79.4 & 90.2\\

            \cmidrule{2-5}
            & \multirow{5}{*}{DarkPose \cite{zhang2020distribution}} & ResNet-50  & 74.6 & 87.1\\

            & & ResNet-101 & 74.6 & 88.7\\

            & & ResNet-152 & 73.0 & 89.1\\

            & & HRNet-W32 & 80.7 & 93.0\\

            & & HRNet-W48 & 78.4 & 87.1\\

            \cmidrule{2-5}
            & \multirow{2}{*}{HRFormer \cite{yuan2021hrformer}} & HRFormer-S & 70.9 & 88.5\\

            \multirow{-15}{*}{\rotatebox{90}{\textit{Heatmap-based}}} & & HRFormer-B & 75.6 & 88.4\\

            \bottomrule 
        \end{tabular}%
\end{table}

\noindent\textbf{Results.\quad}
We visualize the tracking results in \cref{supp:fig:supp_vis_pred_box1}, with the ground-truth bounding boxes and chimpanzee identities shown in the last row. We visualized the confidence scores of the estimated bounding boxes and their associated IDs in each frame obtained by the evaluated methods. It is worth noting that we do not require individual identification of each chimpanzee, but rather assign the same ID to the same animal across frames, following the common practice in multi-human tracking \cite{milan2016mot16}. The estimated box ID is therefore used solely for evaluating the tracking performance. We observed that the evaluated methods performed well in scenarios with minimal occlusion, but struggled to detect and associate the same individual chimpanzee when heavy occlusion occurred. For instance, in \cref{supp:fig:supp_vis_pred_box1}, the infant chimpanzee's bounding box is lost in some frames, and its identity is erroneously switched later due to heavy occlusion. This is a challenging task in chimpanzee detection and tracking, as occlusions frequently occur in group-living habitats. Please refer to the supplementary video for more experimental results. In conclusion, the experimental results reveal the limitations of existing methods for chimpanzee detection and tracking, underscoring the need for more robust algorithms to be developed. We believe that our dataset can make a valuable contribution to the advancement of this field, by providing a challenging benchmark for evaluating and comparing different methods.

\subsection{Pose estimation}\label{supp:sec:pose}

We followed the partition of the dataset as the first track to train and evaluate the methods. \\

\noindent\textbf{Results.\quad}
We report the PCK@0.1 for the 16 keypoints in \cref{supp:tab:supp-posepck01}. The results reveal that the keypoints on the face, such as the eyes and lips, exhibited better estimation compared to the arms and legs. This could be attributed to the fact that eyes and lips have more distinctive visual patterns than limbs, which are often surrounded by heavy fur. \cref{supp:tab:supp-posepck01-action} further reports the PCK@0.1 for each action category on the test set. We observe that different action types exhibit variations in pose accuracy, for example, with climbing generally achieving slightly higher accuracy compared to resting in most methods. This observation can be attributed to the higher potential for self-occlusion during resting, as chimpanzees tend to exhibit significant self-occlusion due to their flexible joints. This is evident in the visualized examples in \cref{supp:fig:supp_pose_action_type}, where (a) and (c) depict resting poses with pronounced self-occlusion. In contrast, during climbing, the body is mostly in an extended state, as shown in (b) and (d). Consequently, the PCK tends to be slightly higher for climbing compared to resting as shown in \cref{supp:tab:supp_pose_action_type}. To validate this assumption, we further evaluate the performance of all the methods for non-occluded poses in \cref{supp:tab:supp-posepck01-visible}. It is interesting to note that all the methods achieve high PCK accuracy when all the keypoints are visible. This demonstrates their effectiveness in accurately estimating poses when occlusions are minimal or absent.

\begin{figure*}[t!]
    \centering
    \includegraphics[width=0.8\linewidth]{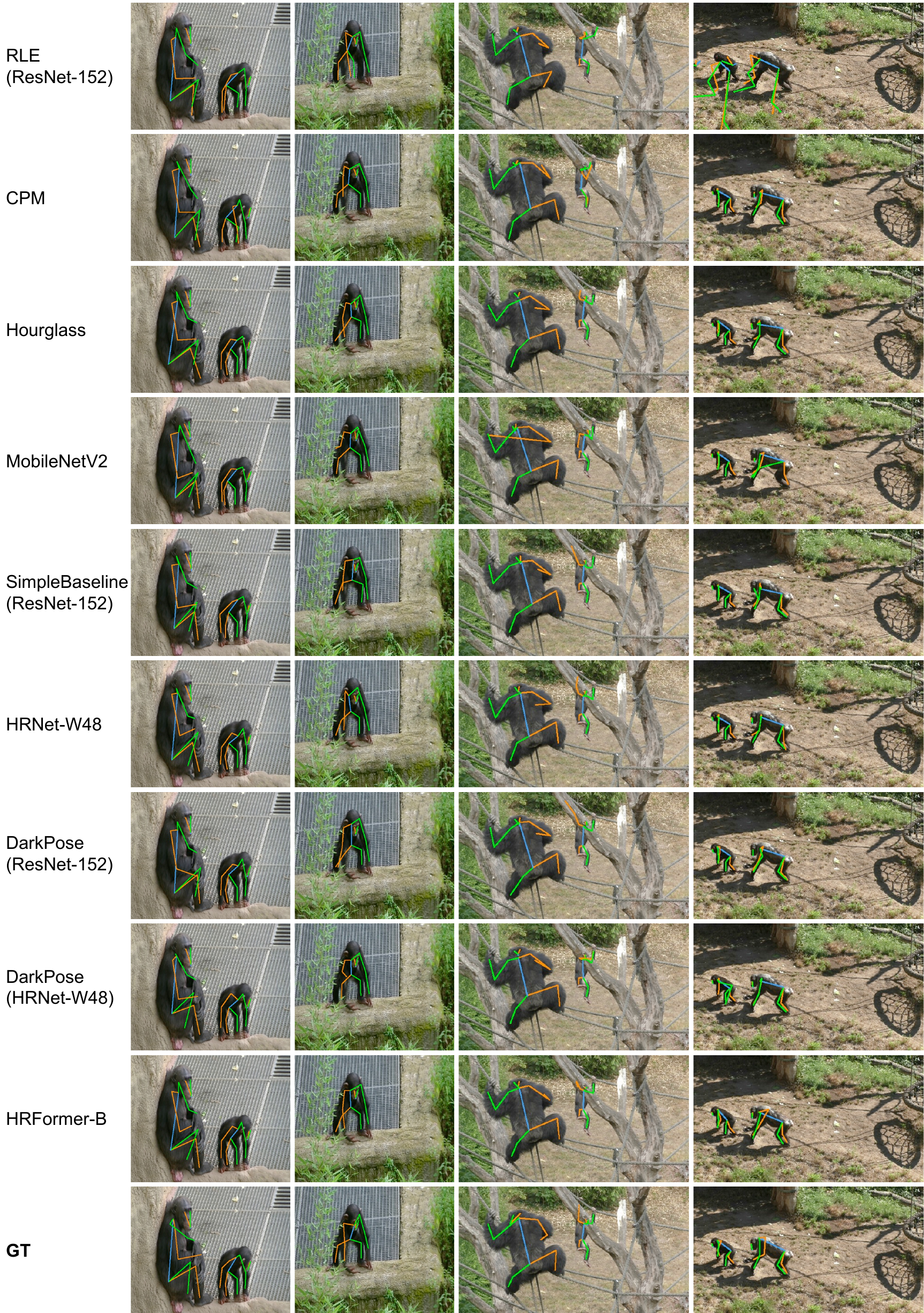} 
    \caption{\textbf{Qualitative results of representative methods on the \dataset test set on the pose estimation task.} The ground-truth poses are shown in the last row.}
    \label{supp:fig:supp_vis_pred_pose1}
\end{figure*}

\begin{table*}[t!]
    \center
    \small
    \caption{\textbf{Results of spatiotemporal action detection track on \dataset test set.} ``\textit{with \ac{gt} box}'' and ``\textit{with Det. box}'' mean using \ac{gt} bounding boxes or detected boxes, respectively.}
    \label{supp:tab:supp-action}
    \setlength{\tabcolsep}{2pt}
    \resizebox{\linewidth}{!}{%
        \begin{tabular}{l c c c c c c c c c c }
            \toprule
            Method & mAP & moving & climbing & sol. obj. playing & eating & grooming & playing & being begged from & aggressing & being nursed\\
            \toprule 
            \hline
            \rowcolor{ggrey} \textit{with \ac{gt} box} & & & & & & & & & & \\ 
            ACRN \cite{sun2018actor} & 24.4 & 60.2 & 23.2 & 38.2 & 54.3 & 7.7 & 42.9 & 0.0 & 0.0 & 4.4\\
            LFB \cite{wu2019long} & 22.4 & 45.3 & 10.0 & 34.4 & 56.3 & 8.7 & 51.0 & 0.4 & 0.0 & 32.1\\
            SlowOnly \cite{feichtenhofer2019slowfast} & 24.5 & 56.1 & 31.6 & 41.0 & 45.4 & 10.4 & 43.0 & 0.0 & 0.0 & 7.5\\
            SlowFast \cite{feichtenhofer2019slowfast} & 24.5 & 60.9 & 37.2 & 47.3 & 35.3 & 10.4 & 49.2 & 0.0 & 0.0 & 7.5\\
            \hline
            \rowcolor{ggrey} \textit{with Det. box} & & & & & & & & & & \\ 
            SlowOnly \cite{feichtenhofer2019slowfast} & 11.8 & 13.4 & 3.5 & 19.4 & 19.9 & 0.3 & 9.4 & 0.0 & 0.0 & 0.0\\
            SlowOnly \textit{w.} Ctx \cite{feichtenhofer2019slowfast} & 13.9 & 16.3 & 6.1 & 16.3 & 19.7 & 1.3 & 12.7 & 0.0 & 0.0 & 0.0\\
            SlowFast \cite{feichtenhofer2019slowfast} & 13.5 & 18.4 & 3.5 & 19.8 & 18.4 & 0.1 & 5.5 & 0.0 & 0.0 & 0.0\\
            SlowFast \textit{w.} Ctx \cite{feichtenhofer2019slowfast} & 16.2 & 16.8 & 6.4 & 20.4 & 19.4 & 0.2 & 1.8 & 0.0 & 0.0 & 0.0\\
            \textbf{\method} (Ours) & 34.3 & 33.3 & 33.4 & 37.1 & 55.0 & 1.9 & 48.9 & 0.3 & 0.0 & 74.0\\
            \bottomrule 
        \end{tabular}%
    }%
\end{table*}

\begin{figure*}[ht!]
    \centering
    \includegraphics[width=0.8\linewidth]{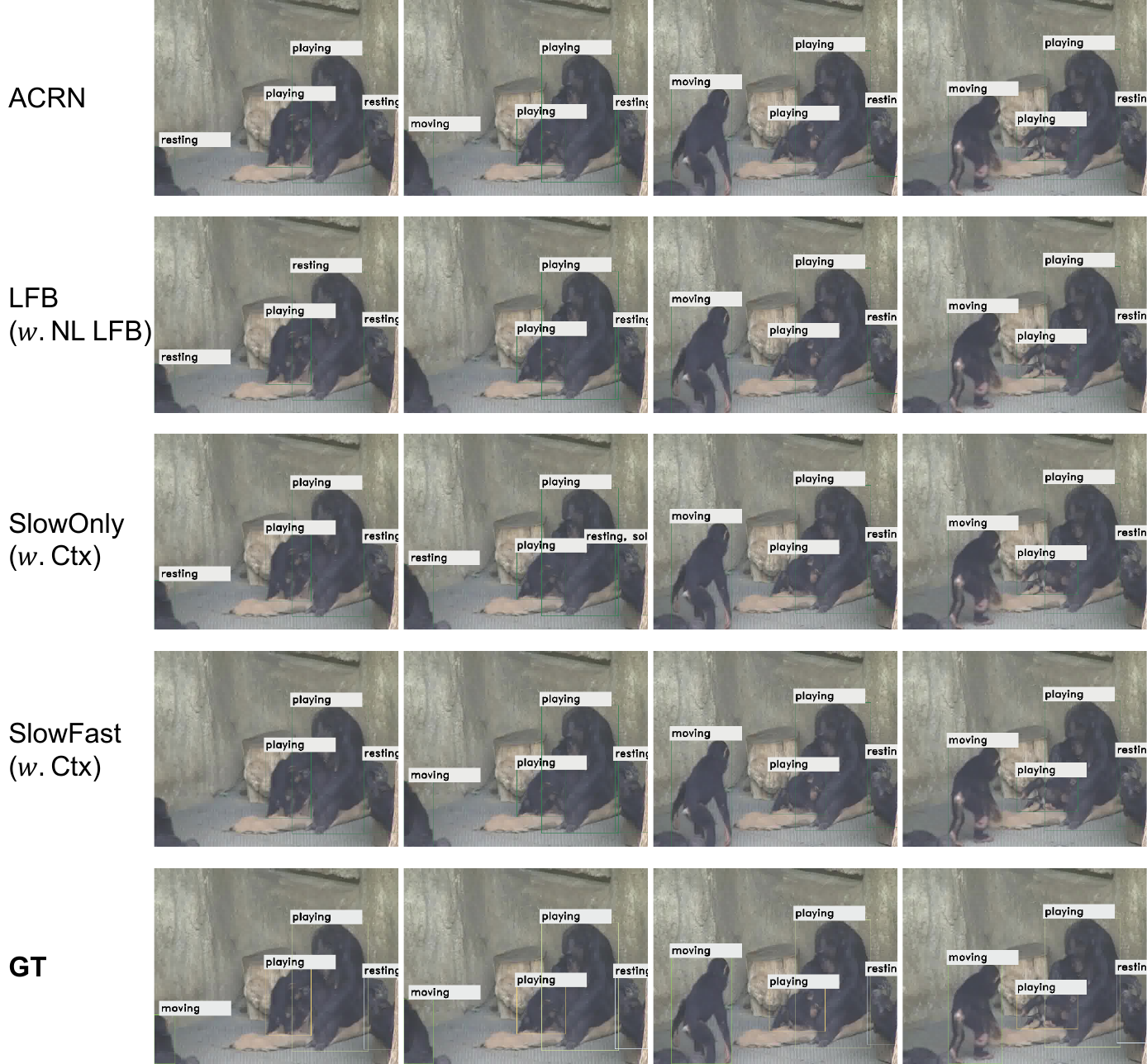} 
    \caption{\textbf{Qualitative results of representative methods on the \dataset test set on the spatiotemporal action detection task.} The ground-truth actions are shown in the last row.}
    \label{supp:fig:supp_vis_pred_action1}
\end{figure*}

These observations highlight the unique and intricate nature of chimpanzee pose estimation, which is complicated by their flexible joint articulations and extended range of motion, as well as the dissimilar physical appearances of their fur in comparison to that of humans. Consequently, developing accurate pose estimation algorithms for chimpanzees requires careful consideration and specialized techniques that account for their unique characteristics.

\begin{figure*}[t!]
    \centering
    \includegraphics[width=0.8\linewidth]{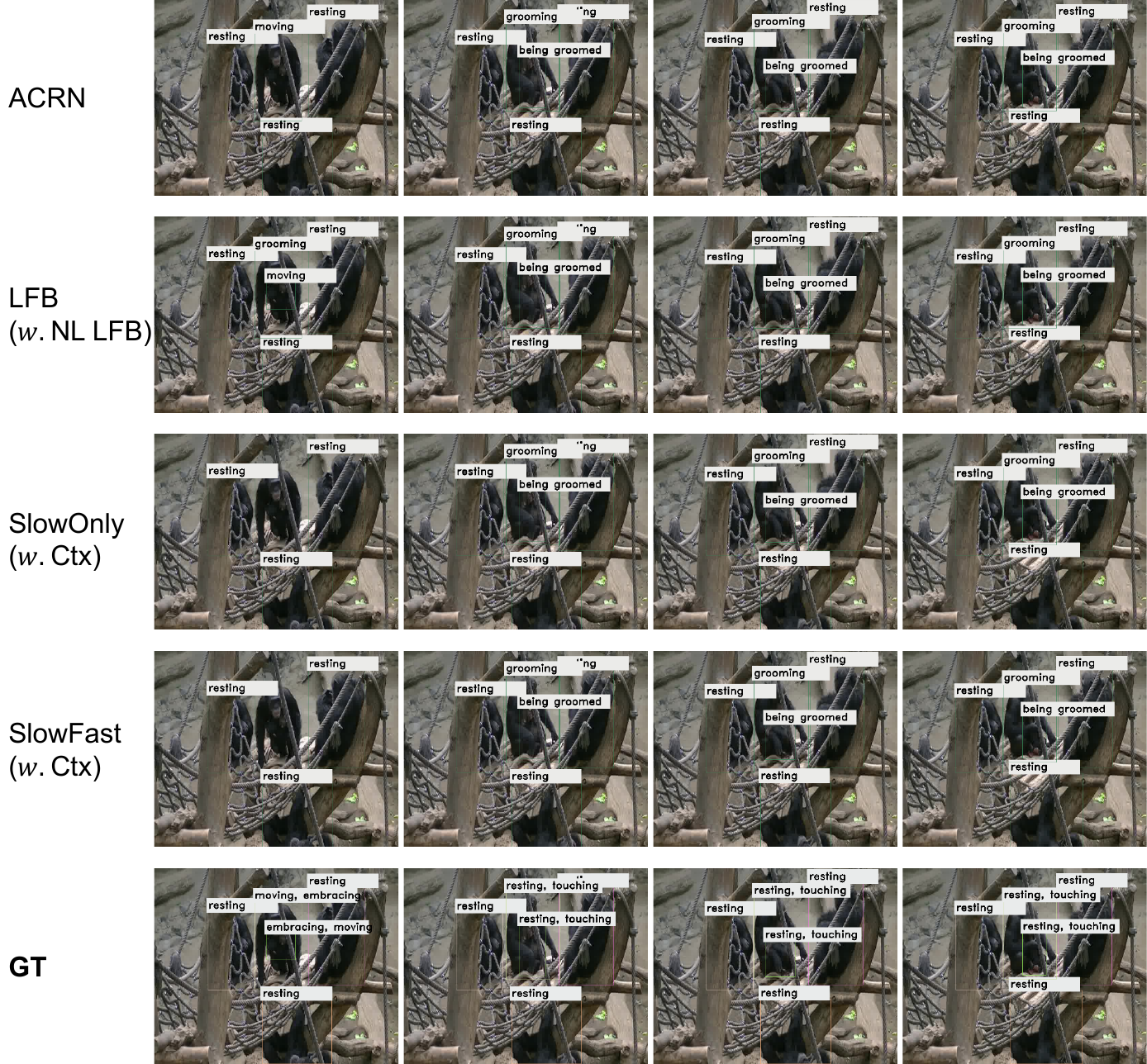} 
    \caption{\textbf{More qualitative results of representative methods on the \dataset test set on the spatiotemporal action detection task.} The ground-truth actions are shown in the last row.}
    \label{supp:fig:supp_vis_pred_action2}
\end{figure*}

\cref{supp:fig:supp_vis_pred_pose1} presents the qualitative results of several models on the \dataset test split, with the ground-truth poses displayed in the last row. It is promising to observe that directly transferring human pose estimation algorithms to chimpanzees yielded decent performance. However, due to self-occlusions and different physical appearance and joint articulations, these models are susceptible to errors in estimating the positions of limbs, as seen in the misaligned right arm and leg of the young chimpanzee in the first column of \cref{supp:fig:supp_vis_pred_pose1}.

\begin{figure*}[t!]
    \centering
    \includegraphics[width=\linewidth]{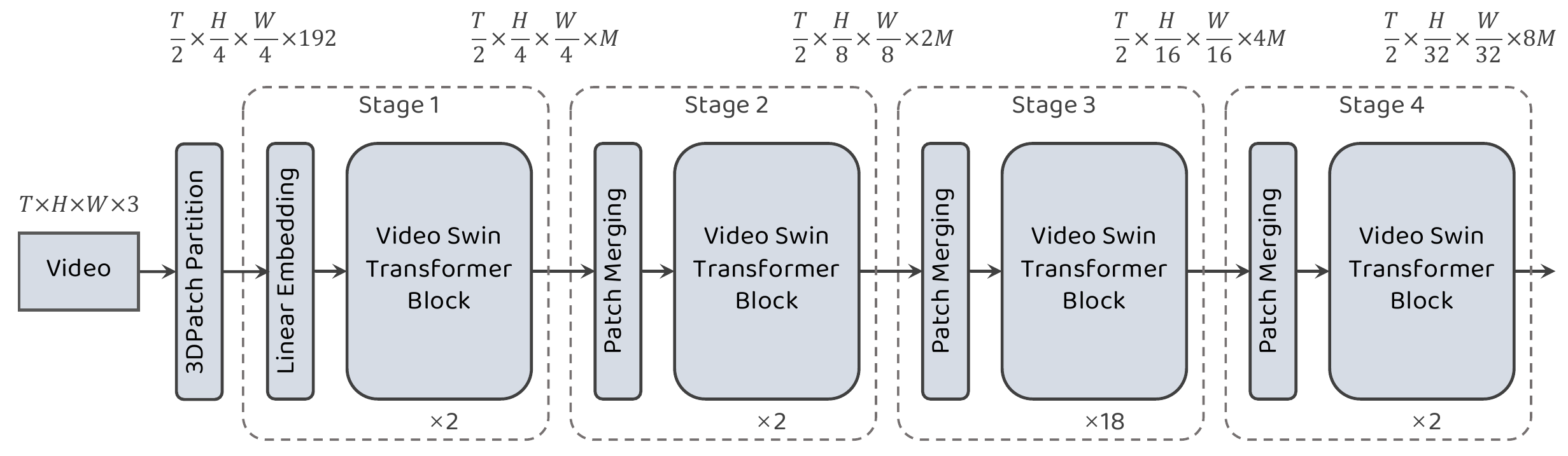} 
    \caption{\textbf{Detailed architecture of the video backbone in our \method}. We follow the Swin-L \cite{liu2022video} architecture design.}
    \label{supp:fig:suppvideobackbone}
\end{figure*}

\subsection{Spatiotemporal action detection}\label{supp:sec:action}

We adopted the same dataset partition as the first track. 
For the four representative methods, we ablated different modules. For LFB \cite{wu2019long}, we ablated different ways of the feature bank operator instantiations, by using non-local (NL) blocks \cite{wang2018non} or average (Avg) or max (Max) pooling. For SlowFast \cite{feichtenhofer2019slowfast} and the variant SlowOnly, we ablated the context module (Ctx), which indicates that using both the RoI feature and the global pooled feature for the action classification. \\

\noindent\textbf{Results.\quad}
We report the mAP for each model's best configuration on several subcategory behaviors in \cref{supp:tab:supp-action}. The models exhibit better performance in detecting locomotion and solitary object interactions, possibly because these actions are relatively simple and involve less interaction between individuals, making it easier for the model to distinguish between action patterns. However, there is still considerable room for improvement in existing models for action categories with higher levels of interaction, such as social interactions. 

We provide qualitative results in \cref{supp:fig:supp_vis_pred_action1,supp:fig:supp_vis_pred_action2}. All methods recognized the playing action of the two chimpanzees in \cref{supp:fig:supp_vis_pred_action1}, but incorrectly classified the touching actions as grooming in \cref{supp:fig:supp_vis_pred_action2}. These two action patterns exhibit subtle differences that significantly challenge the models to distinguish them accurately. We recommend referring to the supplementary video for the video results to observe the difference. The challenges of such distinctions highlight the need for stronger algorithms to address these issues effectively.

Overall, we hope that our work will inspire further research and development in the area of chimpanzee behavior recognition, with the ultimate goal of improving our understanding of chimpanzee and primate behaviors and ecology.

\section{Additional details on \texorpdfstring{\method}{}}\label{supp:sec:methoddetail}
\subsection{Architecture details} 
\cref{supp:fig:suppvideobackbone} illustrates the detailed architectural configuration of the video backbone. In line with \cite{liu2022video}, the 3D patch partition layer obtains $\frac{T}{2} \times \frac{H}{4} \times \frac{W}{4}$ 3D tokens, with each patch (\ie token) consisting of a $\chin$-dimensional feature. Subsequently, four successive stages transform these video tokens into multi-resolution features: specifically, $\videofeat_1 \in \R^{\frac{T}{2} \times \frac{H}{4} \times \frac{W}{4} \times \chmid}$, $\videofeat_2 \in \R^{\frac{T}{2} \times \frac{H}{8} \times \frac{W}{8} \times 2\chmid}$, $\videofeat_3 \in \R^{\frac{T}{2} \times \frac{H}{16} \times \frac{W}{16} \times 4\chmid}$, and $\videofeat_4 \in \R^{\frac{T}{2} \times \frac{H}{32} \times \frac{W}{32} \times 8\chmid}$.

The feature fusion module then transforms these multi-scale temporal features into $\feat_1 \in \R^{\frac{H}{4} \times \frac{W}{4} \times \chout}$, ..., $\feat_4 \in \R^{\frac{H}{32} \times \frac{W}{32} \times \chout}$. This module operates in two key steps: first, temporal merging compresses the features along the temporal dimension using 3D convolutional layers. Next, a channel mapping layer adjusts the feature channels to ensure uniformity, standardizing them to $\chout$ using 2D convolutional layers.

Before inputting the multi-scale features $\feat$ into the Transformer encoder-decoder, we flatten and concatenate them along the spatial dimension. This process results in a feature input of size $(\frac{H}{4}\times\frac{W}{4} + \frac{H}{8}\times\frac{W}{8} + \frac{H}{16}\times\frac{W}{16} + \frac{H}{32}\times\frac{W}{32}) \times \chout$.

\subsection{Implementation details}
In practice, we set $\chin=192$, $M=192$, and $C=512$.
Following the successful pertaining paradigm in foundation models, we first train our model on the Object365 dataset \cite{shao2019objects365} for 40$K$ iterations. We then fine-tune the model on our \dataset dataset for 20$K$ iterations. We set the thresholds for category and behavior classification at 0.3 and 0.3, respectively.

\subsection{Additional results}\label{supp:sec:methodresult}
We report the accuracy of several subcategory behaviors in \cref{supp:tab:supp-action}. It is evident that when using the same detected boxes as input, our method \method shows significant improvements over the baselines, especially in social behaviors. For example, previous methods almost completely failed in categories like playing or being nursed, whereas we achieved substantial improvements. Even when compared to baselines using \ac{gt} boxes as input, our method \method still maintains impressive accuracy.

\begin{figure*}[t!]
    \centering
    \includegraphics[width=.87\linewidth]{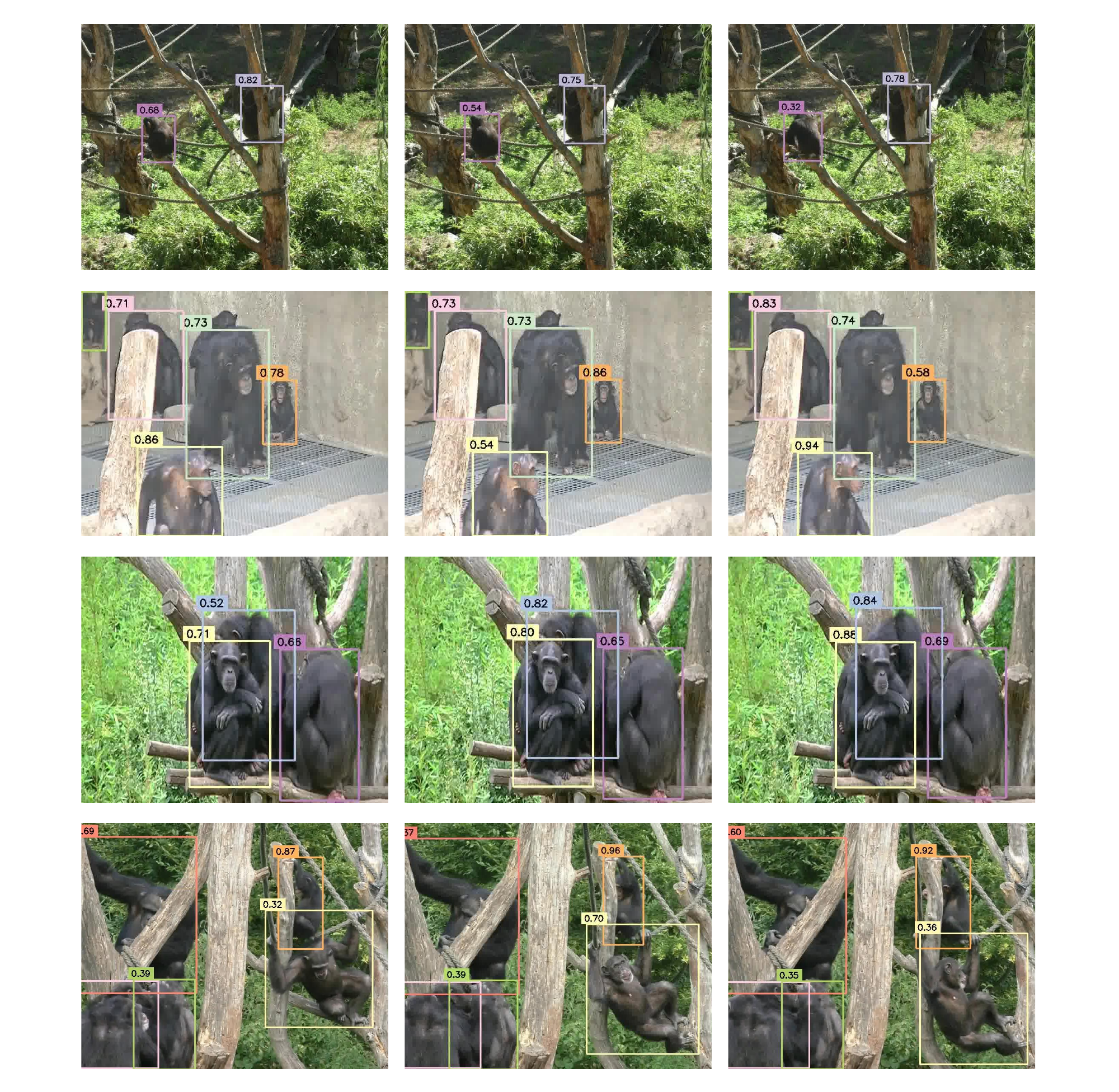}
    \caption{\textbf{Additional visualization of \method's tracking results in different scenarios from \dataset test set.} Consistent colored boxes indicate successful tracking of the same chimpanzee across frames. The numbers represent the confidence scores of chimpanzee classification.}%
    \label{supp:fig:supp_track_success}%
\end{figure*}

We present additional tracking results by our \method in \cref{supp:fig:supp_track_success}. These examples further demonstrate the robustness and effectiveness of our approach, particularly in challenging scenarios. Even when significant occlusion occurs, such as in the third row where a chimpanzee is partially hidden from view, our method successfully detects and tracks the occluded chimpanzee. 

\begin{figure}[t!]
    \centering
    \includegraphics[width=\linewidth]{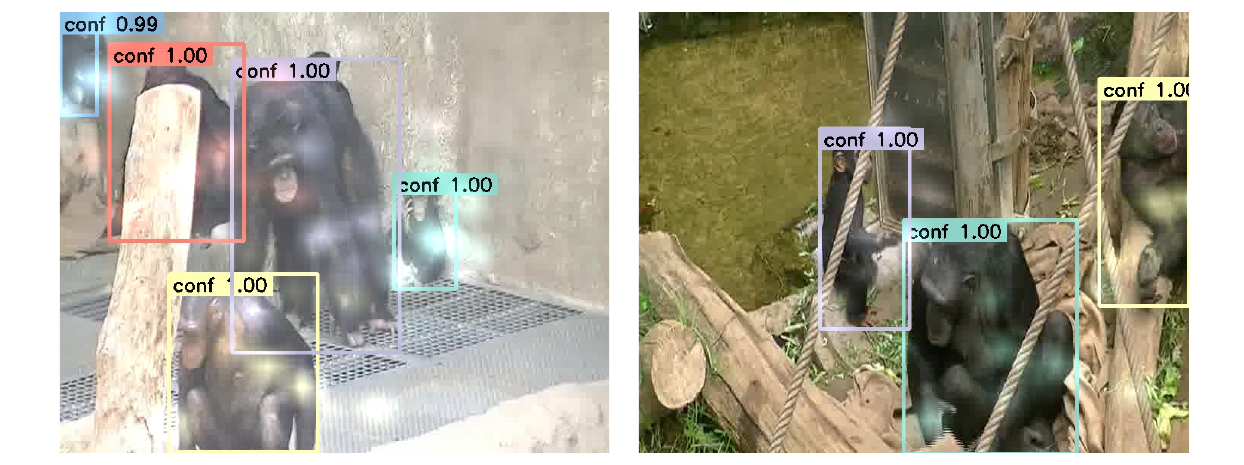}
    \caption{\textbf{Additional visualization of the reference points in the deformable attention module, with a blurring effect applied based on the attention weights.} We visualize the reference points for each query (represented by each box in the image) using the same color as the box. Each reference point is blurred proportionally to its attention weight, with brighter points indicating greater significance. }%
    \label{supp:fig:vis_attention}%
\end{figure}

\cref{supp:fig:vis_attention} visualizes more examples of the reference points within the deformable attention module, with each point blurred according to its attention weights for clarity. These reference points mainly target keypoint areas on chimpanzees, indicating that joint regions may have unique features essential for distinguishing chimpanzees from other objects.

\begin{figure*}[t!]
    \centering
    \includegraphics[width=.87\linewidth]{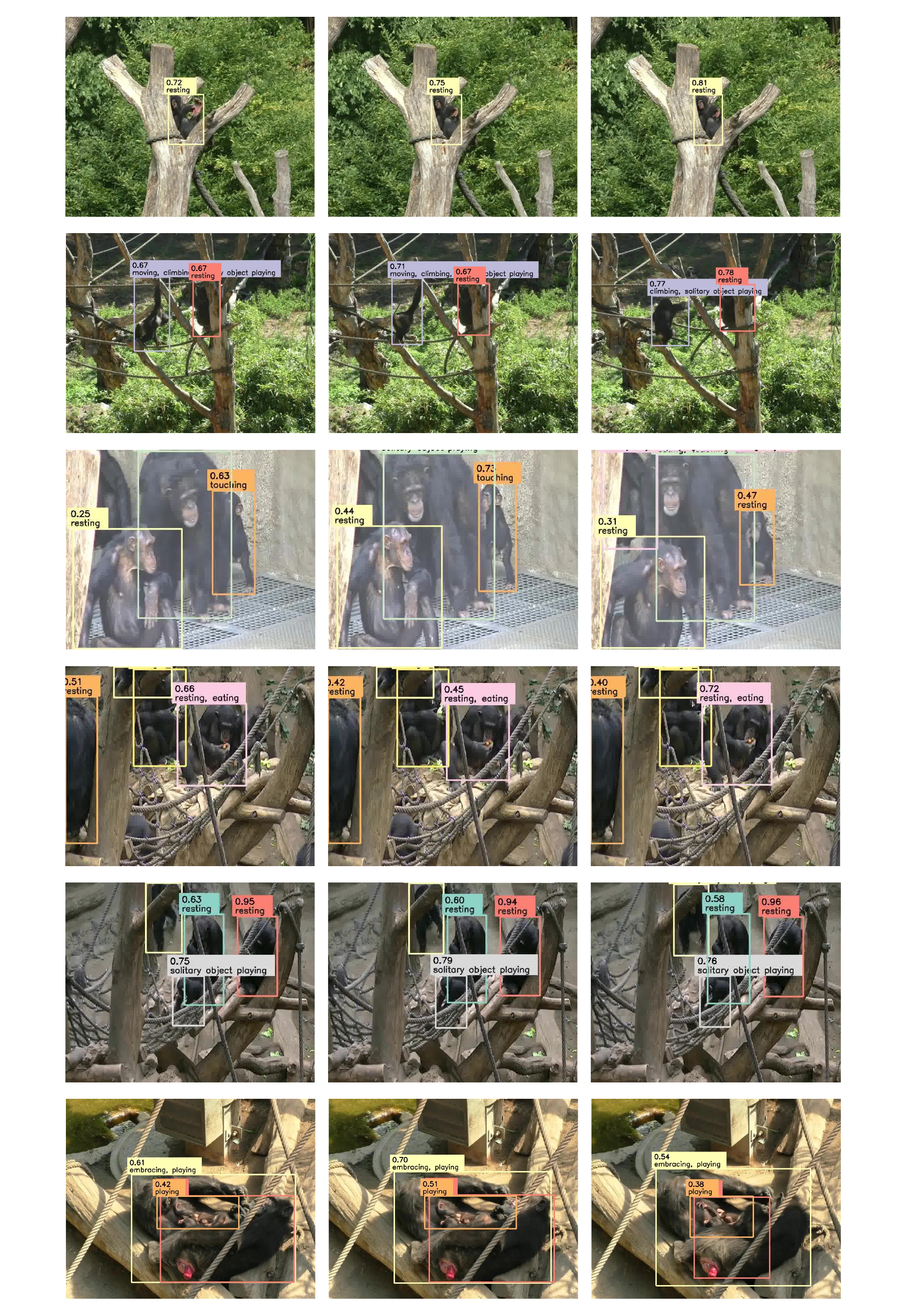} 
    \caption{\textbf{Additional visualization of \method's detection, tracking, and behavior detection results in different scenarios from \dataset test set.} Consistent colored boxes indicate successful tracking of the same chimpanzee across frames. The numbers represent the confidence scores of chimpanzee classification.}%
    \label{supp:fig:supp_quality_result}
\end{figure*}

We present additional qualitative results in \cref{supp:fig:supp_quality_result}, showcasing predictions made by our \method on the test set of the \dataset dataset. These results illustrate the comprehensive capabilities of our model, which not only predicts detection bounding boxes but also performs simultaneous classification of both the class and behaviors in an end-to-end manner. This integrated approach allows \method to efficiently process complex visual scenes, identifying individual chimpanzees and their corresponding actions. By effectively combining detection and classification tasks, \method provides a holistic view of the observed scenes, making it a powerful tool for analyzing and understanding primate behavior in natural settings. Please refer to our \projpage for more video results.

\begin{figure*}[t!]
    \centering
    \includegraphics[width=.85\linewidth]{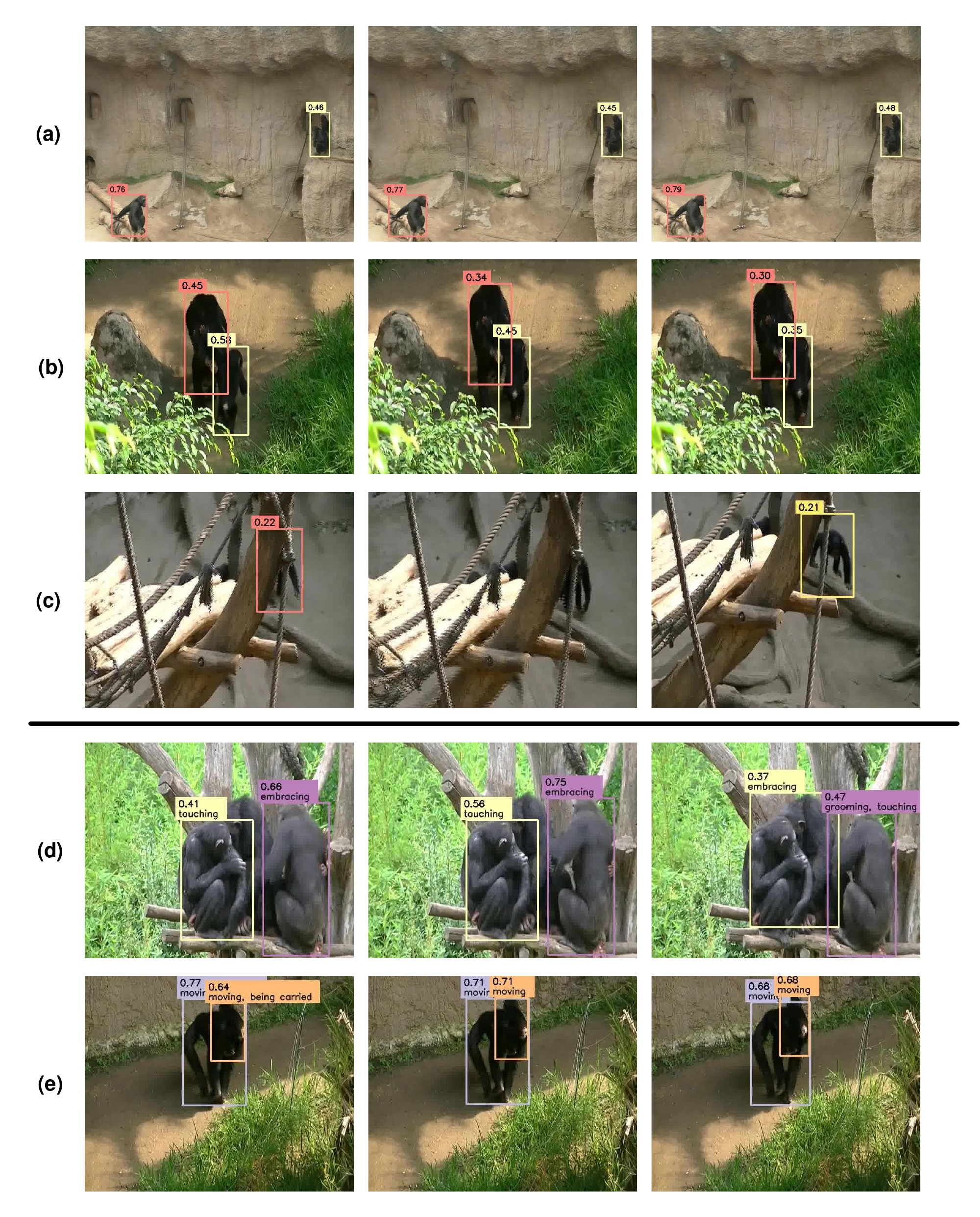}
    \caption{\textbf{Typical failure cases of \method's on \dataset test set.} (a-c) show the failure cases in chimpanzee detection and tracking. (d-e) show the failure cases in chimpanzee behavior recognition.}%
    \label{supp:fig:failure}%
\end{figure*}

\cref{supp:fig:failure} illustrates typical failure cases encountered by our model. Panels (a-c) demonstrate tracking failures, while (d-e) highlight behavior recognition issues. Tracking failures primarily involve detection errors and ID mismatches. In (a), two closely positioned chimpanzees are mistakenly identified as one due to their small appearance in the image. Panel (b) shows an inaccurate bounding box for a young chimpanzee, excluding its left hand---a challenging scenario even for human observers. In (c), an ID change occurs after occlusion, resulting in a bounding box color change upon the chimpanzee's reappearance.

Behavior recognition failures are exemplified in panels (d) and (e). Panel (d) shows a missed behavior estimation due to a detection error, while (e) demonstrates incorrect behavior recognition despite successful detection. For instance, an adult chimpanzee carrying a young one is not correctly identified, and the young chimpanzee's state of `being carried' is unrecognized in subsequent frames.

These failure cases highlight the inherent challenges in chimpanzee perception and behavior analysis. Factors such as the intrinsic appearance similarity among chimpanzees, frequent occlusions in their natural environment, and the complexity of social behaviors involving multiple individuals contribute to these difficulties. These examples underscore the need for continued refinement of our model to better handle such challenging scenarios in chimpanzee behavior analysis.

\section{Discussion}\label{supp:sec:disall}

We now discuss three topics related to the presented work in greater depth.

\subsection{Data collection}

Our proposed \dataset dataset is the first to achieve longitudinal observation of the same social group over a span of four years, which is invaluable for gaining meaningful insights into chimpanzee growth and development. Unlike existing primate datasets, which often consist of unrelated individuals \cite{labuguen2021macaquepose} or are gathered in laboratory settings \cite{marks2022deep}, our dataset offers a more holistic and natural perspective. However, this approach comes with significant costs, requiring primatologists to engage in long-term observation and filming. Additionally, the current focal sampling methods may introduce biases.

In considering future data collection for chimpanzee behavior analysis, non-intrusive methods utilizing automated technology for long-term capture should be prioritized. One promising approach involves the strategic use of camera traps. These devices can be discreetly placed within chimpanzee habitats and programmed to automatically trigger and capture footage when motion is detected. This method allows for continuous monitoring without human presence, thereby minimizing disturbances to the chimpanzees' natural behaviors. 

To ensure sustainability and practicality, these camera traps could be designed with low power consumption features, enabling them to operate efficiently over extended periods. Furthermore, integrating basic AI technologies could enable real-time preprocessing of captured footage, potentially filtering out irrelevant data and focusing on key behavioral events. This approach could significantly reduce logistical challenges while potentially increasing the volume and diversity of captured behaviors, leading to even more comprehensive datasets for future research.

\subsection{Perception of primates}

Our approach, \method, marks a significant advancement as the first to simultaneously perceive and recognize chimpanzee behaviors. This achievement builds upon leveraging strong backbone models from object detection. Prior to our work, there were no models specifically designed for chimpanzee detection, and directly transferring human-centric models \cite{pang2021quasi} yielded suboptimal results. This performance gap likely stems from the inherent differences in appearance and social dynamics between chimpanzees and humans.

Chimpanzees, as highly social animals, often stay in close proximity to each other and share similar appearances. These characteristics pose unique challenges for detection and tracking, as evidenced by the failure cases illustrated in \cref{supp:fig:failure}. Our method addresses these challenges by incorporating temporal context, harnessing additional information from motion to enhance detection capabilities.

While \method demonstrates significant improvements over existing algorithms, there remains room for further refinement. Future work could focus on integrating more chimpanzee-specific prior knowledge into the model. For instance, incorporating facial features with distinctive patterns could potentially enhance detection accuracy, particularly in scenarios involving partial occlusion or ambiguous poses. 

Moreover, exploring advanced techniques in multi-object tracking and behavior recognition could further improve the model's performance in complex social scenarios. By combining these enhancements with our current approach, we anticipate developing more robust and accurate systems for primate perception and behavior analysis, contributing to a deeper understanding of chimpanzee social dynamics and individual development.

\subsection{Understanding of primates}

Automatically understanding chimpanzee behavior presents significant challenges, requiring not only accurate detection and categorization of chimpanzees but also precise behavior estimation. Our approach, \method, represents an initial attempt in this complex domain. To enhance the estimation of chimpanzee social behavior, we designed feature fusion in both temporal and spatial domains and employed attention mechanisms to compute relationships between chimpanzees more effectively.

Drawing inspiration from human-centric understanding, where pose estimation has proven valuable for behavior analysis, we considered using chimpanzee pose as an intermediate representation. However, this approach introduces new complexities due to chimpanzees' high degree of limb movement freedom. Our attempts to incorporate pose estimation in behavior analysis have thus far yielded suboptimal results, highlighting the need for further exploration in this area.

The challenge of recognizing chimpanzee behavior is merely the first step toward a deeper understanding of these primates. Integrating insights from their social networks is crucial for a comprehensive comprehension of the chimpanzee world, which in turn can provide valuable insights into human social development. This underscores the importance of developing more sophisticated models that can capture not only individual behaviors but also complex social dynamics and interactions.

Future research directions might include refining pose estimation techniques specifically tailored to chimpanzee anatomy and movement patterns. Additionally, exploring alternative intermediate representations could bridge the gap between raw visual data and high-level behavior understanding. Developing methods to analyze long-term social patterns and group dynamics from sequences of detected behaviors also presents a promising avenue for advancement.

As we continue to progress in this field, the potential for cross-disciplinary insights between primatology and human social sciences grows. However, the path to fully understanding chimpanzee behavior through automated means remains a long and challenging journey, requiring ongoing collaboration between computer vision experts and primatologists.

\end{document}

%% file: config.tex
\usepackage{hyperref}
\usepackage{xspace}
\usepackage{graphicx} \graphicspath{{figures/}}
\usepackage{amsmath,amssymb,nicefrac,bbm,pifont}
\usepackage{cite}
\usepackage[linesnumbered,ruled,vlined]{algorithm2e}
\usepackage{acronym}
\usepackage[skip=3pt,font=small]{subcaption}
\usepackage[skip=3pt,font=small]{caption}
\usepackage[dvipsnames,svgnames,x11names,table]{xcolor}
\usepackage[capitalise,noabbrev,nameinlink]{cleveref}
\usepackage{booktabs,tabularx,colortbl,multirow,multicol,array,makecell,tabularray}
\usepackage{overpic,wrapfig}
\usepackage[misc]{ifsym}
\usepackage{enumitem}
\usepackage{orcidlink}
\usepackage{dblfloatfix}

\newcommand{\dataset}[0]{\texttt{ChimpACT}\xspace}
\newcommand{\method}[0]{AlphaChimp\xspace}
\newcommand{\dataprojpage}{\href{https://shirleymaxx.github.io/ChimpACT/}{project page}\xspace}
\newcommand{\projpage}{\href{https://sites.google.com/view/alphachimp/home}{project page}\xspace}

\definecolor{gblue}{HTML}{4285F4}
\definecolor{gred}{HTML}{DB4437}
\definecolor{ggreen}{HTML}{0F9D58}
\definecolor{ggrey}{HTML}{E9E9E9}
\definecolor{gbest}{HTML}{FFFFFF}
\definecolor{gsecond}{HTML}{FFFFFF}

\crefname{algorithm}{Alg.}{Algs.}
\Crefname{algocf}{Algorithm}{Algorithms}
\crefname{section}{Sec.}{Secs.}
\Crefname{section}{Section}{Sections}
\crefname{table}{Tab.}{Tabs.}
\Crefname{table}{Table}{Tables}
\crefname{figure}{Fig.}{Figs.}
\Crefname{figure}{Figure}{Figures}
\crefname{equation}{Eq.}{Eqs.}
\Crefname{equation}{Equation}{Equations}
\crefname{appendix}{Appx.}{Appxs.}
\Crefname{appendix}{Appendix}{Appendices}

\newcommand{\cmark}{\ding{51}}%
\newcommand{\xmark}{\ding{55}}%

\makeatletter
\DeclareRobustCommand\onedot{\futurelet\@let@token\@onedot}
\def\@onedot{\ifx\@let@token.\else.\null\fi\xspace}
\def\eg{\emph{e.g}\onedot} 

\def\ie{\emph{i.e}\onedot}

\makeatother

\newcommand{\image}{\mathbf{I}}
\newcommand{\R}{\mathbb{R}}
\newcommand{\videofeat}{\mathbf{V}}
\newcommand{\feat}{\mathbf{F}}
\newcommand{\bbox}{\mathbf{b}}
\newcommand{\allbbox}{\mathbf{B}}
\newcommand{\allcls}{\mathbf{C}}
\newcommand{\allbehave}{\mathbf{A}}

\newcommand{\action}{\mathbf{a}}
\newcommand{\cls}{c}

\newcommand{\actionnum}{K}
\newcommand{\chin}{C_{in}}
\newcommand{\chmid}{M}
\newcommand{\chout}{C}

\newcommand{\querynum}{Q}

\acrodef{sota}[SOTA]{State-of-the-Art}
\acrodef{mot}[MOT]{Multi-Object Tracking}
\acrodef{mlp}[MLP]{Multilayer Perceptron}
\acrodef{gt}[GT]{ground-truth}
\acrodef{giou}[GIOU]{Generalized Intersection over Union}
\acrodef{gpu}[GPU]{Graphics Processing Unit}
\acrodef{reid}[ReID]{Re-Identification}
\acrodef{map}[mAP]{mean Average Precision}
\acrodef{mota}[MOTA]{Multiple Object Tracking Accuracy}
\acrodef{motp}[MOTP]{Multiple Object Tracking Precision}
\acrodef{fp}[FP]{False Positives}
\acrodef{fn}[FN]{False Negatives}
\acrodef{id}[ID]{Identity Switch}
\acrodef{idf1}[IDF1]{Identity F1 Score}
\acrodef{hota}[HOTA]{Higher Order Tracking Accuracy}
\acrodef{qdtrack}[QDTrack]{Quasi-Dense Similarity Learning for Multiple Object Tracking}
\acrodef{cpm}[CPM]{Convolutional Pose Machines}
\acrodef{hrnet}[HRNet]{High-Resolution Network}
\acrodef{pck}[PCK]{Percentage of Correctly estimated Keypoints}